\documentclass[journal]{IEEEtran}
\usepackage{graphicx} % DO NOT CHANGE THIS
\usepackage{amsmath,amsfonts}
\usepackage{algorithmic}
\usepackage{algorithm}
\usepackage{array}
\usepackage{textcomp}
\usepackage{stfloats}
\usepackage{url}
\usepackage{verbatim}
\usepackage{natbib}
\setcitestyle{numbers,square}
\usepackage[colorlinks=true, linkcolor=red, citecolor=green, urlcolor=red]{hyperref}

\usepackage{helvet}  % DO NOT CHANGE THIS
\usepackage{courier}  % DO NOT CHANGE THIS
\usepackage{ragged2e} 
\usepackage{booktabs,makecell, multirow, tabularx}
\usepackage{colortbl}

\usepackage{amssymb}
\usepackage[capitalize]{cleveref}
\usepackage{natbib}  % DO NOT CHANGE THIS AND DO NOT ADD ANY OPTIONS TO IT
\usepackage{algorithm}
\usepackage{algorithmic}
\usepackage{graphicx}
\usepackage{subcaption}
\usepackage{caption}

\usepackage{newtxtext,newtxmath}
\usepackage[export]{adjustbox}
\usepackage{newfloat}
\usepackage{listings}
\usepackage[x11names]{xcolor}
\definecolor{first}{HTML}{F48288}  % 最深 - 第一名
\definecolor{second}{HTML}{FAC791} % 中等 - 第二名
\definecolor{third}{HTML}{FFFF8F}  % 最浅 - 第三名
\definecolor{gray}{HTML}{D3D3D3}  % 差
\DeclareCaptionStyle{ruled}{labelfont=normalfont,labelsep=colon,strut=off} % DO NOT CHANGE THIS
\floatstyle{ruled}
\newfloat{listing}{tb}{lst}{}
\floatname{listing}{Listing}
\begin{document}
\title{Fast and Robust Deformable 3D Gaussian Splatting}

\author{Han Jiao, Jiakai Sun, Lei Zhao*, 
Zhanjie Zhang, 
Wei Xing, and Huaizhong Lin*
\thanks{This work was supported in part by Zhejiang Province Program (22024C03263, 025C01068, LZ25F020006), Zhejiang Provincial Cultural Relics Protection Science and Technology Project (2024009), the National Program of China (62172365), Macau project: Key technology research and display system development for new personalized controllable dressing dynamic display, and Ningbo Science and Technology Plan Project (2025Z052, 2025Z062, 2022Z167, 2023Z137).}
\thanks{Han Jiao, Jiakai Sun, Lei Zhao, 
Zhanjie Zhang, 
Wei Xing, and Huaizhong Lin are with the College of Computer Science and Technology,  Zhejiang University, China (e-mail: csjh@zju.edu.cn; csjk@zju.edu.cn; cszhl@zju.edu.cn; 
cszzj@zju.edu.cn; 
wxing@zju.edu.cn; linhz@zju.edu.cn).}
\thanks{Lei Zhao* and Huaizhong Lin* are the corresponding authors.}
}

\markboth{IEEE TRANSACTIONS ON VISUALIZATION AND COMPUTER GRAPHICS}%
{How to Use the IEEEtran \LaTeX \ Templates}

\maketitle
\begin{abstract}
3D Gaussian Splatting has demonstrated remarkable real-time rendering capabilities and superior visual quality in novel view synthesis for static scenes. Building upon these advantages, researchers have progressively extended 3D Gaussians to dynamic scene reconstruction. Deformation field-based methods have emerged as a promising approach among various techniques. These methods maintain 3D Gaussian attributes in a canonical field and employ the deformation field to transform this field across temporal sequences. Nevertheless, these approaches frequently encounter challenges such as suboptimal rendering speeds, significant dependence on initial point clouds, and vulnerability to local optima in dim scenes. To overcome these limitations, we present FRoG, an efficient and robust framework for high-quality dynamic scene reconstruction. FRoG integrates per-Gaussian embedding with a coarse-to-fine temporal embedding strategy, accelerating rendering through the early fusion of temporal embeddings. Moreover, to enhance robustness against sparse initializations, we introduce a novel depth- and error-guided sampling strategy. This strategy populates the canonical field with new 3D Gaussians at low-deviation initial positions, significantly reducing the optimization burden on the deformation field and improving detail reconstruction in both static and dynamic regions. Furthermore, by modulating opacity variations, we mitigate the local optima problem in dim scenes, improving color fidelity. Comprehensive experimental results validate that our method achieves accelerated rendering speeds while maintaining state-of-the-art visual quality.
\end{abstract}
\begin{IEEEkeywords}
3D Gaussian Splatting, dynamic scene reconstruction
\end{IEEEkeywords}

\section{Introduction}
\IEEEPARstart{T}{he} exponential growth of video content and pervasive adoption of VR/AR technologies have positioned dynamic scene reconstruction as a pivotal research domain in computer vision and graphics, garnering substantial academic interest.  Early geometry-driven approaches \cite{HSFVV, Motion2fusion} employed dynamic primitive-based representations for scene modeling, yet were constrained by inadequate realism and insufficient detail preservation. 
\par
% Neural Radiance Fields (NeRF)~\cite{NeRF} set a new benchmark for high-fidelity static view synthesis, inspiring many follow-up works to apply this paradigm to dynamic scenes~\cite{D-NeRF, NeuralSceneFlow, SpacetimeNerf, Hypernerf, Nerfies, StreamRF, DyNeRF, TiNeuVox, Dynibar, TemporalInterpolation, K-Planes, Hexplane, 9845414,10612248,10599827,10980040}. 
Recently, Neural Radiance Fields (NeRF)\cite{NeRF} have set a new benchmark for high-fidelity static view synthesis, stimulating extensive research in diverse areas\cite{wang2023neuslearningneuralimplicit,vgos,M_ller_2022,verbin2021refnerfstructuredviewdependentappearance,bian2023nopenerfoptimisingneuralradiance}. Notably, a significant line of these follow-up works focuses on adapting this powerful paradigm to dynamic scenes~\cite{D-NeRF, NeuralSceneFlow, SpacetimeNerf, Hypernerf, Nerfies, StreamRF, DyNeRF, TiNeuVox, Dynibar, TemporalInterpolation, K-Planes, Hexplane, 9845414,10612248,10599827,10980040}.
While these extensions are capable of representing dynamic content, they also inherit NeRF's major limitation: the rendering pipeline requires costly per-pixel ray sampling and MLP evaluations, resulting in prohibitive latency for real-time use. 
\par
To break this real-time barrier, 3D Gaussian Splatting (3DG-S)~\cite{3DGS} introduced an explicit 3D Gaussian (3DG) representation and a rasterization-based pipeline, achieving high-quality real-time rendering. This success has spurred numerous extensions to dynamic scenes, with deformation field-based methods being a prominent direction. 
% Among these, E-D3DGS~\cite{ed3dgs} is a state-of-the-art (SOTA) approach by introducing per-Gaussian and temporal embeddings.
Among these, E-D3DGS~\cite{ed3dgs} introduces per-Gaussian and temporal embeddings.
It employs a dual-deformation mechanism, where coarse and fine temporal embeddings are separately fed into the deformation network—a form of late-fusion architecture. However, this method faces several critical challenges. First, applying the deformation field twice per forward pass incurs significant rendering overhead. Second, and more fundamentally, it inherits a strong sensitivity to the initial point cloud. Most deformation-based methods rely on gradient-based densification, which performs poorly in sparsely initialized regions. This forces the deformation field to navigate a highly complex and ill-conditioned optimization landscape in order to learn long-range transformations and ``pull'' distant 3DGs into empty areas, a process that often leads to motion artifacts and detail loss. Third, these methods are often prone to local optima in dim scenes, causing color shifts. 
\par
To overcome these limitations, we propose FRoG, a fast and robust framework for dynamic scene reconstruction. We utilize per-Gaussian embedding and temporal embedding as input to the deformation field. Unlike E-D3DGS, which employs coarse and fine deformations, we significantly improve the rendering speed through Hadamard product-based early temporal embedding fusion, while preserving high-quality rendering results.
Moreover, we address the challenge of sparse initialization with a novel canonical field sampling strategy. 
While depth-guided initialization or error-based sampling has been explored in static reconstruction~\cite{fang2024minisplattingrepresentingscenesconstrained} and curve-based methods~\cite {SpacetimeGaussians}, these approaches are often limited to static or explicitly structured scenes. In contrast, our work explores the far more challenging problem of sampling within an unstructured canonical field. By systematically observing the canonical field's stability and experimentally investigating the trade-offs of inverse deformation mappings, we propose to solve this by injecting low-deviation anchors. This approach significantly simplifies the deformation field's learning task without the instability of solving ill-posed inverse problems.
% While sampling based on rendering errors was explored in the curve-based ST-GS~\cite{SpacetimeGaussians}, its explicit trajectory model allows for straightforward analytical sampling. In contrast, our work tackles the far more complex inverse problem of inferring canonical properties from a deformed space, which we solve by injecting low-deviation anchors to simplify the deformation field's learning task.
Additionally, we implement an aggressive opacity reduction strategy to address the issue of local optima, which occurs in dim scenes when 3DGs near the camera introduce color compensation to training views. These improvements make our method more robust when handling both sparse initial point regions and dim scenarios. In conclusion, our main contributions are outlined as follows:
\begin{itemize}
    \item We propose FRoG, a fast and robust method for dynamic scene reconstruction using deformable 3D Gaussians.
    \item We systematically explore temporal embedding fusion strategies and introduce an early-stage fusion based on the Hadamard product, which significantly improves rendering speed by avoiding the costly dual deformation of late-fusion schemes, while maintaining comparable rendering quality.
    \item We design a canonical field sampling strategy that injects low-deviation anchors to ease the deformation field's optimization challenge, enhancing reconstruction quality in regions with sparse point cloud initialization.
    \item We propose an aggressive opacity reduction strategy that mitigates local optima in dim scenes with challenging shadows, improving color fidelity.
\end{itemize}

\section{Related work}
\subsection{NeRF-Based Dynamic Scene Reconstruction}
Dynamic scene reconstruction is a challenging task in computer vision. Given the remarkable capability of NeRF to model static scenes, many researchers have extended its application to dynamic scenes. We categorize NeRF-based dynamic scene reconstruction methods as follows.

\textbf{Flow-based} methods \cite{NeuralSceneFlow, Dynibar} introduce optical flow priors to model and supervise scene flow, achieving notable results in dynamic scene reconstruction. However, these methods heavily depend on 2D priors. \textbf{Spatiotemporal coupling} methods \cite{TiNeuVox, fourier, K-Planes, TemporalInterpolation, mixedvoxel, Hexplane, DyNeRF} incorporate a time input $t$; they directly query the radiance field using spatiotemporal coordinates, significantly reducing storage requirements and achieving high rendering quality. \textbf{Per-frame training} methods \cite{ReRF, StreamRF} learn an initial neural radiance field from the first frame and update it in subsequent frames, enabling fast and online reconstruction of dynamic scenes. However, the constraints of online reconstruction prevent direct rendering of any frame and require more storage. \textbf{Deformation field-based} methods~\cite{D-NeRF, SpacetimeNerf, nonrigid, Nerfies, Hypernerf,nerfplayer,forwardflow,10666828,10599827} assume that scene motion stems from the deformation of the static structure. They employ the deformation field to transform the canonical radiance field to any time step, yielding promising results.
\par
Despite their advancements, most of these methods struggle to achieve real-time rendering due to the high computational demands inherent to neural radiance fields.
\subsection{3DG-S-Based Dynamic Scene Reconstruction}
The field of 3DG-S-based dynamic reconstruction can also be categorized along similar lines to the NeRF-based methods.
\par
\textbf{Flow-based} methods\cite{gaussianflowoptic, motiongs, MoDGS} leverage optical flow priors to assist in dynamic scene reconstruction, achieving impressive reconstruction results. However, these methods heavily rely on priors such as depth or optical flow, which limits their performance in complex scenarios where 2D priors are inaccurate. 
\textbf{4DGS-based} methods \cite{realtime4dgs, 4dgsrotor,zhang2025megamemoryefficient4dgaussian} extend 3DGs into 4D XYZT Gaussians to model dynamic scenes.  These methods obtain 3D Gaussian representations by slicing the 4D Gaussians at different time steps, thereby inheriting the fast rendering speed of 3DG-S. Among them, MEGA~\cite{zhang2025megamemoryefficient4dgaussian} introduces a compact DAC color representation and entropy-constrained deformation to significantly compress 4DGS storage. \textbf{Per-frame training} methods \cite{3dgstream, hicom, queen, hu20254dgcrateaware4dgaussian} achieve rapid training and real-time rendering of dynamic scenes. However, similar to per-frame training methods in NeRF, they lack flexibility in rendering specific frames and suffer from high storage overhead. \textbf{Curve-based} methods~\cite{SpacetimeGaussians, gaussianflow, ex4dgs} fit 3DG attributes into time-dependent curves, enabling fast rendering. However, they tend to exhibit significant degradation as the video sequence becomes longer.
\par
\textbf{Deformation field-based} methods~\cite{deformable3dgs, xu2024grid4d4ddecomposedhash, SC-GS, zhao2024gaussianprediction, mihajlovic2024splatfieldsneuralgaussiansplats, 4dgs_kplanes, dn4dgs, ed3dgs,wu2025swift4dadaptivedivideandconquergaussiansplatting,10932755,10969103,11198844,lu20243dgeometryawaredeformablegaussian,jiao2025mapomotionawarepartitioning} employ a deformation field to transform the canonical field to each time step. Specifically, D3DGS~\cite{deformable3dgs} models the deformation field using an MLP and takes xyzt as input, while 4DGaussians~\cite{4dgs_kplanes} first extracts spatiotemporal features from Hexplane~\cite{Hexplane} and then derives the deformation results. However, both methods struggle with scenes featuring rapid or complex motions. Subsequent works have explored various directions to enhance these models. For example, GaGS~\cite{lu20243dgeometryawaredeformablegaussian} explicitly extracts 3D geometry features and integrates them in learning the 3D deformation.  Some focus on improving motion modeling or control by introducing auxiliary structures, such as control points in SC-GS~\cite{SC-GS} and GaussianPrediction~\cite{zhao2024gaussianprediction}. Others, like Swift4D~\cite{wu2025swift4dadaptivedivideandconquergaussiansplatting}, prioritize efficiency by employing 4D hash grids for temporal modeling, though often at the cost of reconstruction quality. A powerful trend is to enhance the deformation mechanism itself using learnable embeddings. Both DN-4DGS~\cite{dn4dgs} and E-D3DGS~\cite{ed3dgs} explore a dual-deformation paradigm. DN-4DGS aggregates spatiotemporal information from adjacent points and frames, while E-D3DGS replaces the xyzt input with per-Gaussian and temporal embeddings, connecting coarse and fine temporal embeddings to per-Gaussian embeddings separately and feeding them into the deformation field to achieve dual deformation. Although this embedding-based approach significantly improves the modeling quality of dynamic regions, the dual-deformation design common to these methods severely compromises rendering speed. Furthermore, these deformation field-based methods primarily rely on gradient-based densification. In regions with sparse or missing initial 3DGs, this densification mechanism is largely ineffective. Consequently, the deformation field is forced to compensate by learning highly complex, long-range transformations. Such an immense optimization challenge limits their ability to reconstruct fine details and introduces artifacts. Additionally, many of these methods are susceptible to local optima when modeling dim scenes, leading to noticeable color shifts. In contrast, our work presents a holistic solution that addresses these key limitations in efficiency and robustness. First, we replace the dual-pass, late-fusion paradigm with a single-pass model enabled by an early fusion of temporal embeddings, significantly boosting rendering efficiency. Second, to address the shortcomings of gradient-based densification, our novel canonical field sampling strategy directly injects low-deviation anchors, a more robust approach for completing geometry in sparse regions by directly reducing the optimization challenge. Finally, our aggressive opacity reduction strategy tackles another key challenge to robustness: it targets and mitigates local optima issues that arise in dim scenes with complex shadows.

\subsection{Opacity and Attribute Regularization}
Recent advancements have explored various strategies for opacity manipulation to enhance the training stability and efficiency of 3DG-S. Revising Densification~\cite{bulò2024revisingdensificationgaussiansplatting} addresses the cumulative opacity bias during the cloning process by introducing a mathematical correction, ensuring the rendering remains consistent before and after densification. It also replaces the standard periodic hard reset with a gradual linear decay of opacity values to achieve smoother training dynamics. 3DGS-MCMC~\cite{kheradmand20253dgaussiansplattingmarkov} reinterprets the optimization as a sampling process, employing an opacity redistribution scheme during 3DG relocalization to maintain rendering consistency. Furthermore, it incorporates L1-regularization on opacity into the loss function to encourage global parsimony and eliminate redundant 3DGs. Perceptual-GS~\cite{zhou2025perceptualgssceneadaptiveperceptualdensification} incorporates human visual sensitivity into the densification process to prioritize visually critical regions. Specifically, it introduces an opacity decline mechanism during the cloning operation, employing a power function to suppress the proliferation of redundant 3DGs and facilitate pruning.
\par
Our work targets the specific challenges inherent in dynamic deformation, distinguishing itself from the aforementioned methods in several key aspects. Our strategy is not intended to maintain rendering consistency during densification, nor is it designed for model compression or simple redundancy removal by limiting the total count of 3DGs. Instead, our aggressive opacity reduction strategy serves as a dynamic error-correction mechanism. By monitoring the temporal evolution of attributes ($\Delta \alpha$) during deformation, our strategy actively suppresses ``shadow floaters'' that introduce color bias when the deformation field converges to local optima. This approach ensures high color fidelity in complex dynamic scenes

\section{Preliminary}
\subsection{3D Gaussian Splatting}
3D Gaussian Splatting~\cite{3DGS} is an explicit, point-based representation. Each 3D Gaussian $G$ is defined by a mean $\mu$, a covariance matrix $\Sigma$, an opacity $\alpha$, and spherical harmonic (SH) coefficients $sh$. The covariance matrix $\Sigma$ is constructed from a scaling matrix $S$ and a rotation matrix $R$ via $\Sigma = RSS^{T}R^{T}$. The matrix $S$ is formed from a 3D scaling vector $s$, while $R$ is derived from a rotation quaternion $q$. For rendering, each 3DG is projected onto the 2D image plane, yielding a 2D Gaussian. These 2D Gaussians are then sorted by depth and alpha-blended to compute the pixel color $C$:
\begin{align}
    C = \sum_{i=1}^{N} c_{i} \alpha_{i}^{'} \prod_{j=1}^{i-1} \left(1 - \alpha_{j}^{'}\right), \label{eq:rendering}
\end{align}
where $c_i$ is the color derived from SH coefficients and viewing direction, and $\alpha_i^{'}$ is the 2D evaluation of the 3D Gaussian's opacity. This rasterization-based pipeline enables real-time novel view synthesis.

\subsection{Deformable 3D Gaussian Splatting}

To extend 3DG-S to dynamic scenarios, a deformation field $\Phi$, typically parameterized by a Multi-Layer Perceptron (MLP), is employed to transform the canonical 3D Gaussians to their states at a specific time $t$. 
For the $i$-th 3D Gaussian $G_i$, characterized by its canonical attributes $\{\mu_i, s_i, q_i, \alpha_i, sh_i\}$, the deformation field predicts the attribute variations $\Delta G_{i,t}$ as follows:
\begin{equation}
    \Delta {G}_{i,t} = \{\Delta \mu, \Delta s, \Delta q, \Delta \alpha, \Delta sh\}_{i,t} = \Phi(f_{\text{spatial}}, f_{\text{temporal}}),
\end{equation}
where $f_{\text{spatial}}$ represents the spatial information, which can be the positionally encoded 3D coordinates or a learnable per-Gaussian embedding. $f_{\text{temporal}}$ denotes the temporal information, typically provided as a positionally encoded timestamp or a learnable temporal embedding. The deformed attributes at time $t$ are subsequently obtained by adding these predicted variations to the corresponding canonical attributes:
\begin{equation}
    {G}_{i,t} = {G}_{i} + \Delta {G}_{i,t}.
\end{equation}
During training, the attributes of the canonical field and the parameters of the deformation field are jointly optimized to minimize reconstruction errors. 
\par
In particular, while the above formulation represents a general additive transformation, recent state-of-the-art methods like E-D3DGS \cite{ed3dgs} employ a dual-deformation paradigm. In such late-fusion architectures, the variation is predicted through multiple forward passes of the deformation network to handle coarse and fine temporal embeddings separately. Although effective for modeling complex motions, this design significantly increases rendering overhead. Our work builds upon the general formulation but optimizes the integration of temporal information to achieve a more efficient single-pass deformation.

\section{Methods}
\begin{figure*}
    \centering
    \includegraphics[width=\linewidth]{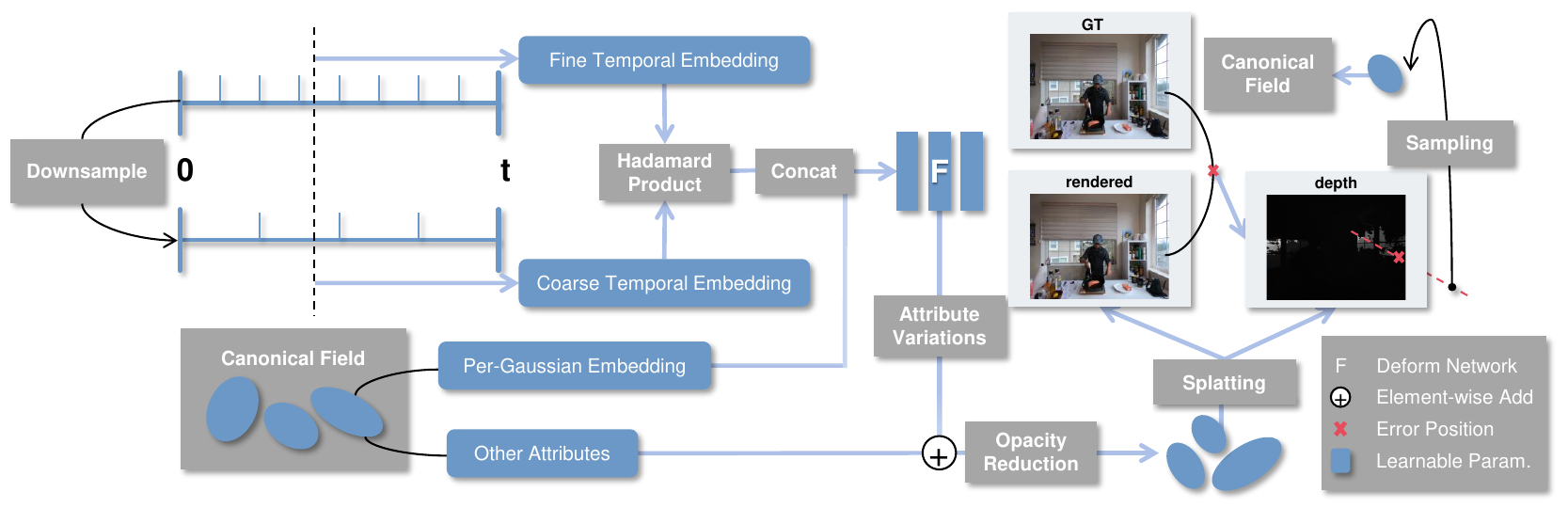}
    \caption{\textbf{Overview of FRoG.} We begin by performing early fusion on the coarse and fine temporal embeddings through the Hadamard product, aiming to enhance rendering speed while preserving high rendering quality. During rendering, we apply an aggressive opacity reduction strategy to prevent the model from converging to local optima. To further enhance the rendering quality, we compute the error between the rendered image and the ground truth, and sample in the canonical field based on the depth map, thereby complementing the details in regions where the initial point cloud is sparse.
}
\label{fig: overview}
\vspace{-2mm}
\end{figure*}

This section begins with a detailed exploration of early fusion for temporal embeddings in Sec \ref{sec:early_fusion}, where we propose an early fusion method based on the Hadamard product. Subsequently, in Sec \ref{sec:canonical_sample}, we introduce a sampling strategy for the canonical field, which samples new 3DGs using the depth map and error map. Furthermore, to address the issue of falling into local optima when modeling dim scenes, we present an aggressive opacity reduction strategy in Sec \ref{sec:opacity_reduction}. Finally, we describe our optimization strategy in Sec \ref{Optimization}. We depict an overview of our approach in Fig.\ref{fig: overview}.

\subsection{Early Fusion for Temporal Embedding}\label{sec:early_fusion}
Traditional deformation field-based methods\cite{deformable3dgs,4dgs_kplanes} use 4D coordinates to compute attribute variations of 3DGs. In contrast, E-D3DGS~\cite{ed3dgs} learns per-Gaussian and temporal embeddings, removing the need for coordinate inputs. 
However, E-D3DGS adopts a dual deformation based on coarse and fine temporal embeddings, where fine temporal embeddings are obtained via interpolation of the temporal embedding sequence, while coarse temporal embeddings are derived by downsampling the sequence, followed by interpolation. This late fusion method for temporal embeddings notably reduces rendering speed. 
\par
The inefficiency of the late-fusion paradigm in E-D3DGS naturally motivated us to explore early-fusion strategies to replace its time-consuming dual deformation. The concept of fusing multi-resolution features has been successfully validated for efficient spatio-temporal modeling in methods like HexPlane~\cite{Hexplane}, which provided a key inspiration for our exploration. Although our scenario focuses solely on one-dimensional temporal embeddings, we inferred that this multi-scale fusion philosophy could be effectively adapted. To this end, we systematically investigated several potential fusion mechanisms, including addition, attention, concatenation, and the Hadamard product, aiming to identify an optimal solution that could significantly improve rendering speed while ensuring effective temporal information integration.
\par
Addition simply aggregates information across all dimensions, making it difficult to highlight local variations, which reduces modeling accuracy in dynamic regions. The attention mechanism can learn weighted distributions between embeddings, but it relies on additional network layers, leading to high computational overhead. Concatenation directly stacks embeddings, requiring extra network parameters to learn the fusion relationships while being less sensitive to local signals in the embeddings. In contrast, the Hadamard product models nonlinear interactions between embeddings through element-wise multiplication. This operation effectively captures temporal-local signals without introducing additional computational overhead, offering high computational efficiency while maintaining reconstruction quality. Specifically, we derive the fused embedding $t_{fused}$ based on the coarse temporal embedding $t_c$ and the fine temporal embedding $t_f$ as follows:
\begin{equation}
    t_{fused} =  t_c \odot t_f.
\end{equation}
Next, we concatenate the fused temporal embedding with the per-Gaussian embedding $e$, which is unique to each 3DG, and feed it into the MLP $F$ to compute the attribute variations of 3DGs in the canonical field:
\begin{equation}
    \Delta \mu, \Delta s, \Delta q, \Delta \alpha,\Delta {sh} = F(concat(t_{fused},e)).
\end{equation}
By adding variations to the attributes of 3DGs in the canonical field, we can obtain the 3DGs at the target time step.

% ========== START OF REVISED SECTION 4.2 ==========
\subsection{Canonical Field Sampling Strategy}\label{sec:canonical_sample}
Standard densification techniques, which rely on splitting existing 3DGs, are often ineffective in regions with sparse or missing initial points. This limitation presents an immense optimization challenge for the deformation field, forcing it to learn complex, long-range transformations to compensate for the missing geometry. Such a process often struggles to converge well, resulting in blurred details and degraded reconstruction quality. To address this fundamental challenge, we propose a canonical field sampling strategy, which directly populates these sparse regions with new 3DGs. Our strategy first locates high-error 3D coordinates, then injects new 3DGs as low-cost anchors to simplify the subsequent optimization.

\begin{figure}[!htb]
    \centering
    % --- Row 1: flame_salmon ---
    \begin{subfigure}{0.48\linewidth}
        \centering
        \includegraphics[width=\linewidth]{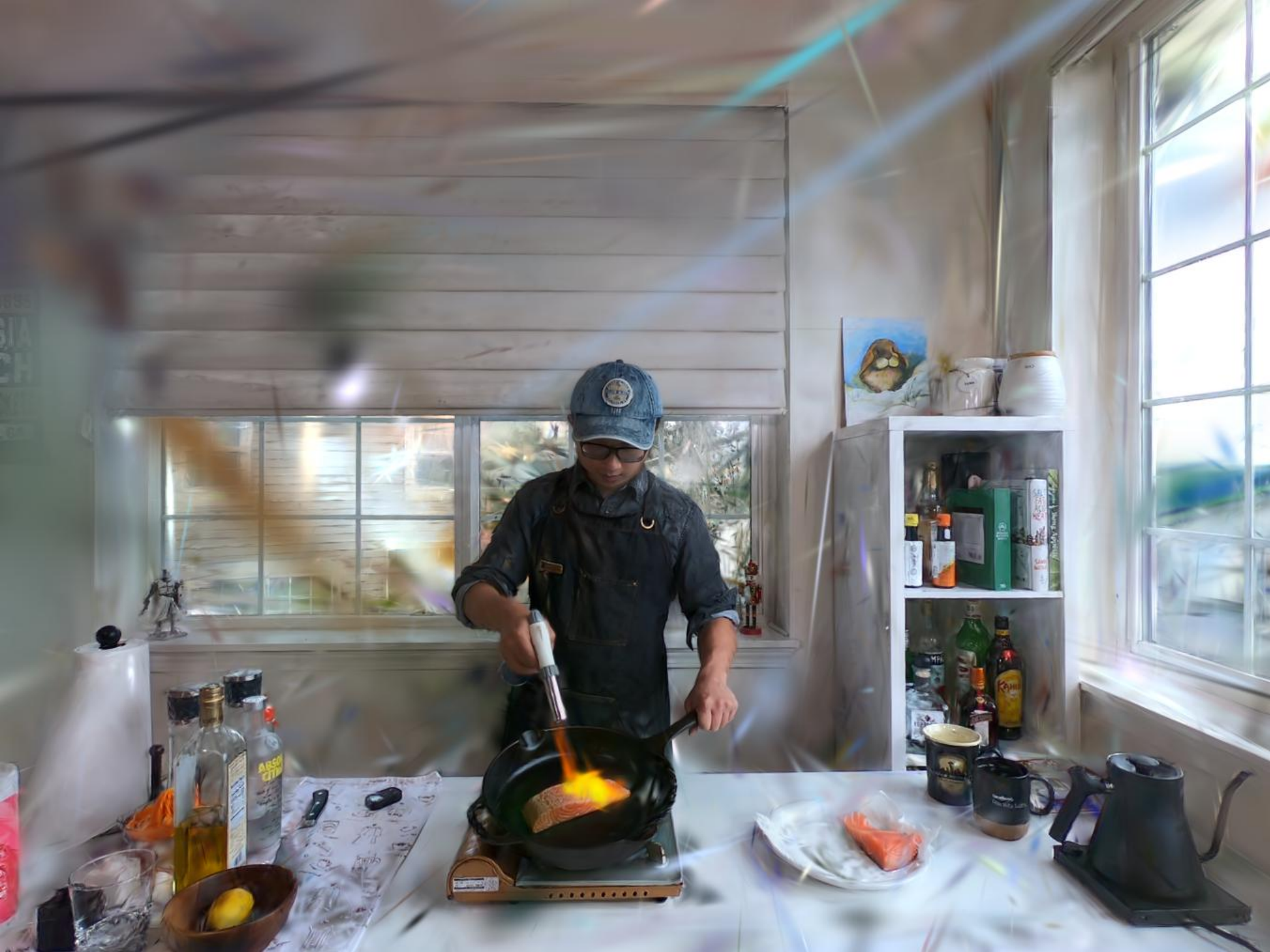}
        \caption*{(a) Canonical}
        \label{fig:canonical_salmon}
    \end{subfigure}
    \hfill
    \begin{subfigure}{0.48\linewidth}
        \centering
        \includegraphics[width=\linewidth]{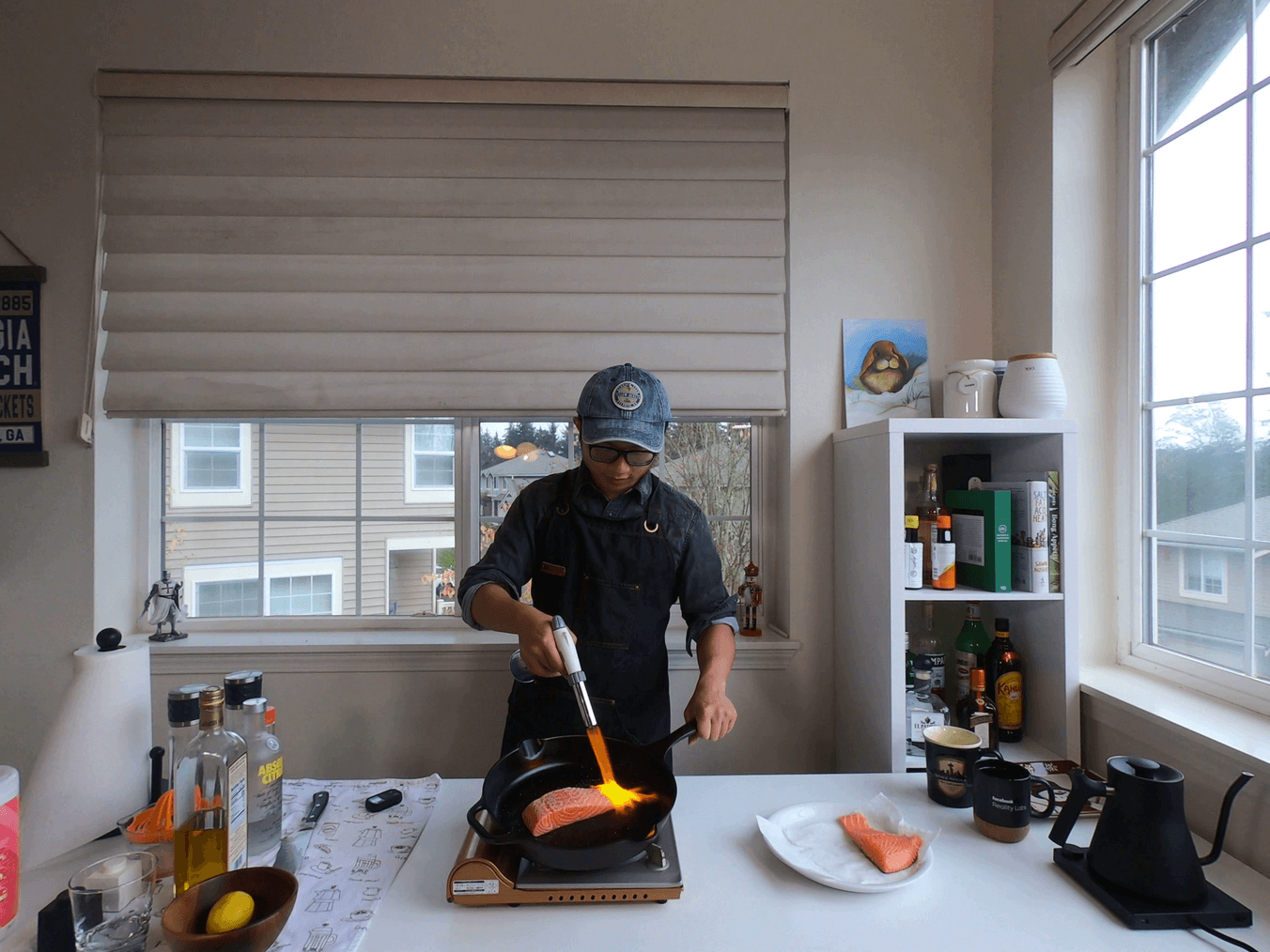}
        \caption*{(b) GT}
        \label{fig:gt_salmon}
    \end{subfigure}

    % --- VERTICAL SPACE CONTROL BETWEEN ROW 1 AND 2 ---
    % Adjust the value '5mm' to increase or decrease the space.
    % Use a negative value like '-2mm' to pull the rows closer.
    \vspace{2mm} 

    % --- Row 2: cook_spinach ---
    \begin{subfigure}{0.48\linewidth}
        \centering
        \includegraphics[width=\linewidth]{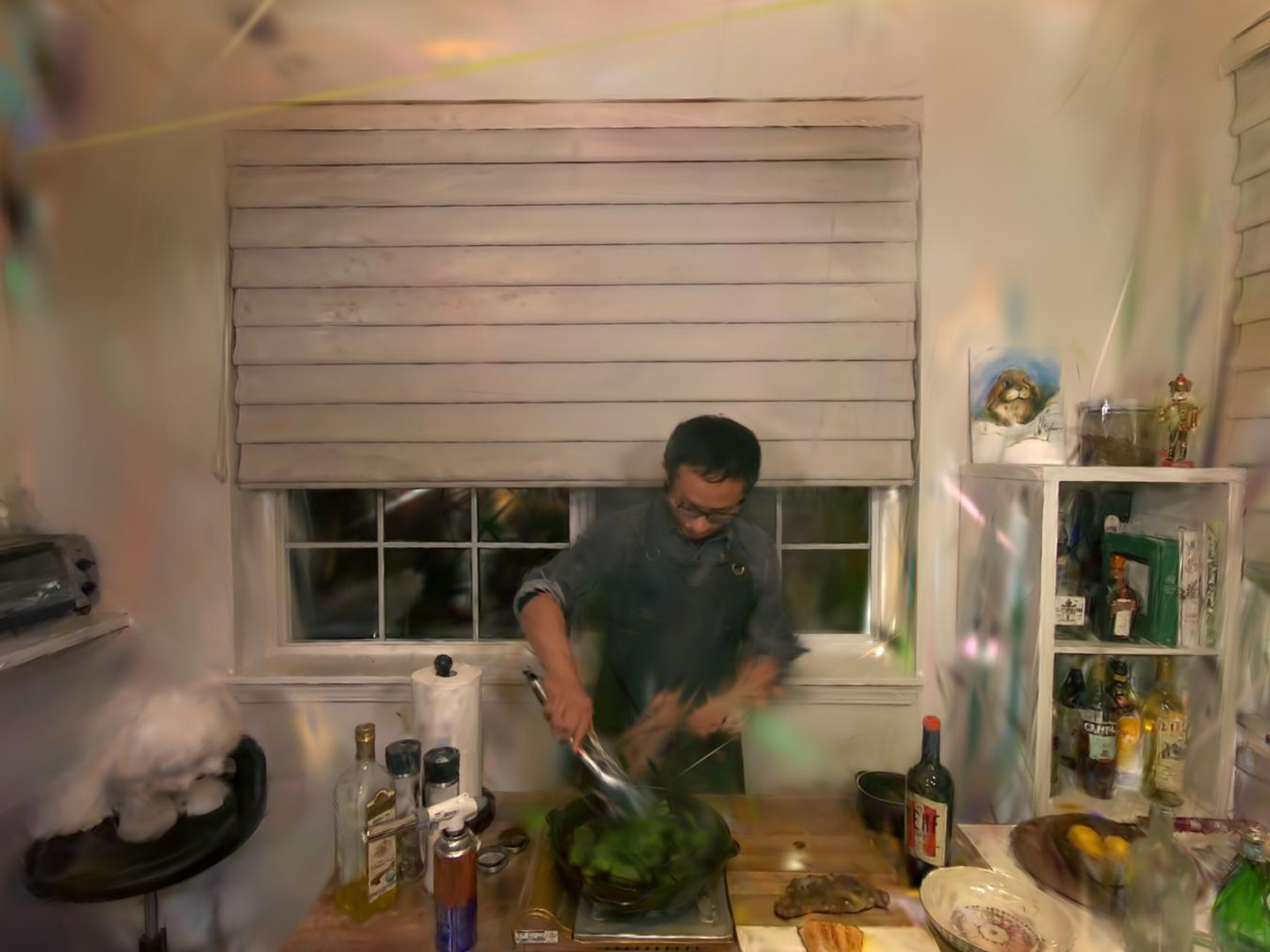}
        \caption*{(c) Canonical}
        \label{fig:canonical_spinach}
    \end{subfigure}
    \hfill
    \begin{subfigure}{0.48\linewidth}
        \centering
        \includegraphics[width=\linewidth]{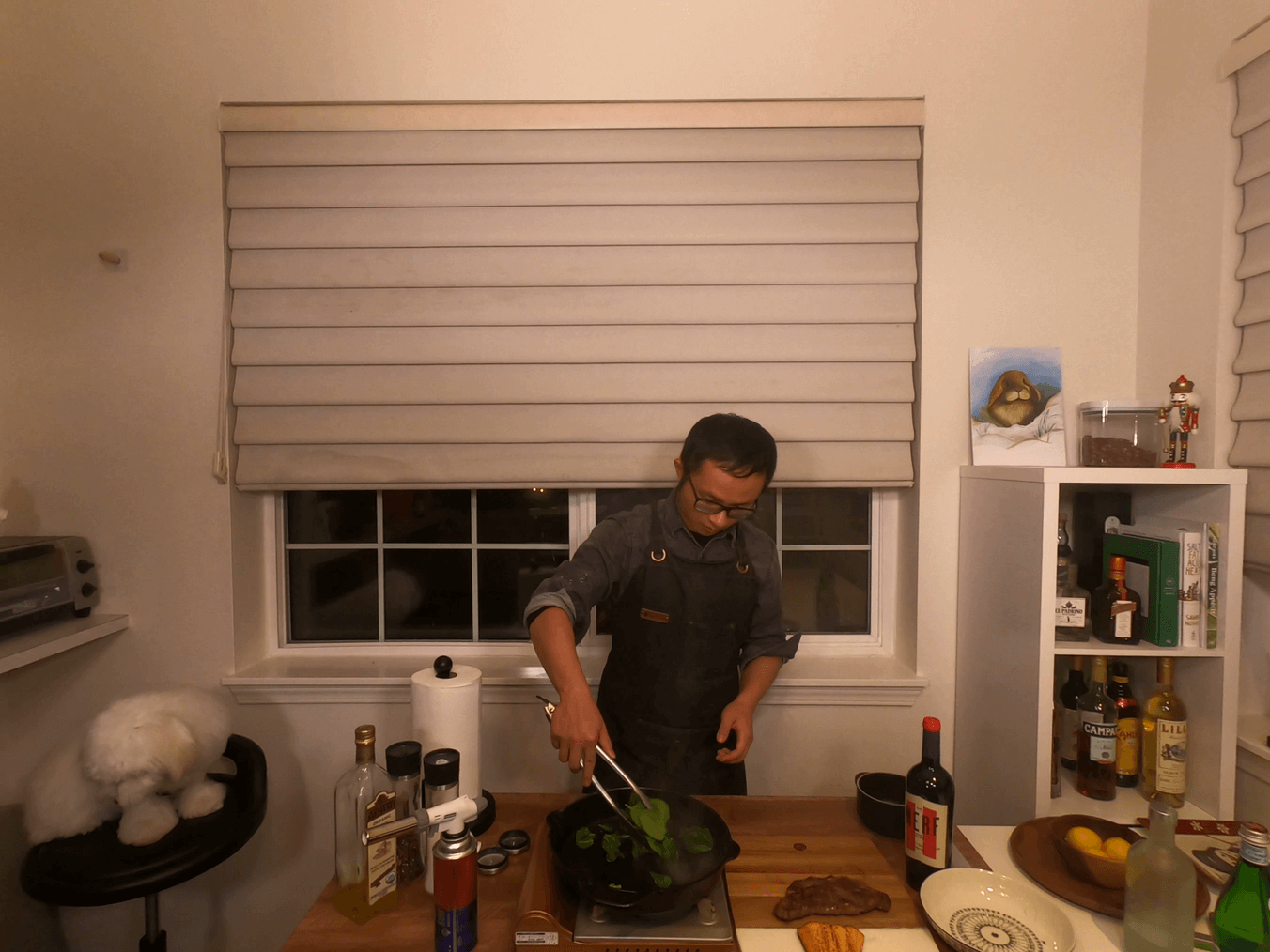}
        \caption*{(d) GT}
        \label{fig:gt_spinach}
    \end{subfigure}

    % --- VERTICAL SPACE CONTROL BETWEEN ROW 2 AND 3 ---
    % Adjust this value as needed.
    \vspace{2mm}

    % --- Row 3: chicken & peel-banana ---
    \begin{subfigure}{0.24\linewidth}
        \centering
        \includegraphics[width=\linewidth]{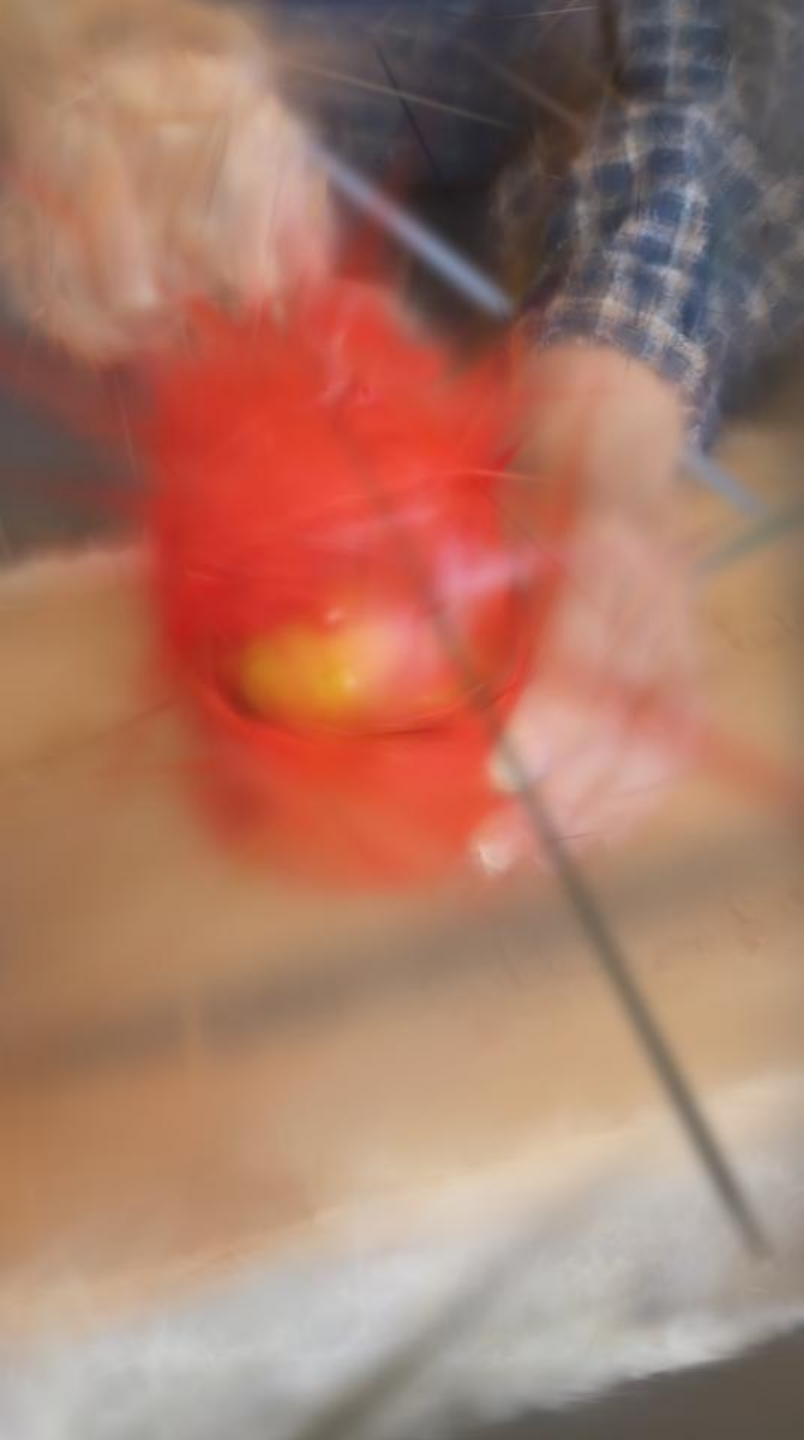}
        \caption*{(e) Canonical}
        \label{fig:canonical_chicken} % Corrected label
    \end{subfigure}
    \hfill
    \begin{subfigure}{0.24\linewidth}
        \centering
        \includegraphics[width=\linewidth]{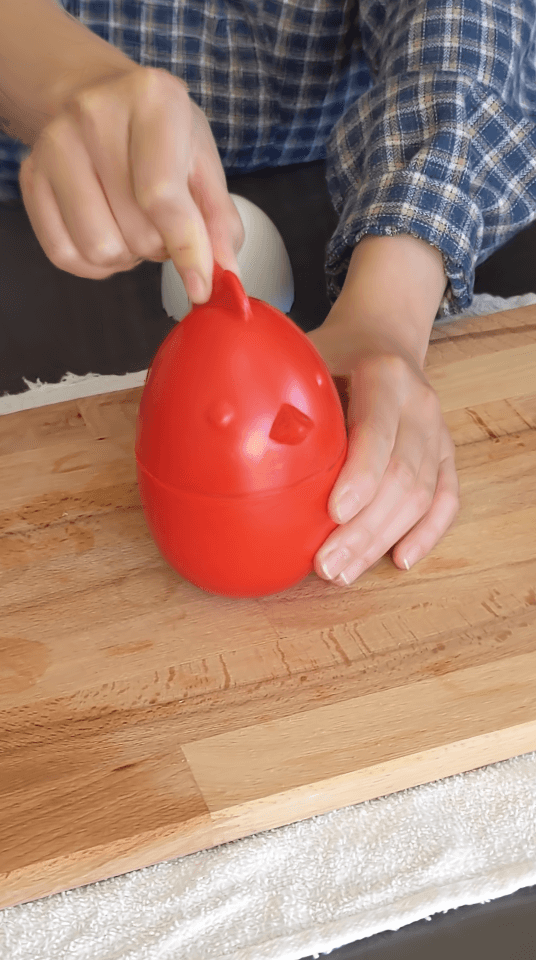}
        \caption*{(f) GT}
        \label{fig:gt_chicken} % Corrected label
    \end{subfigure}
    \hfill
    \begin{subfigure}{0.24\linewidth}
        \centering
        \includegraphics[width=\linewidth]{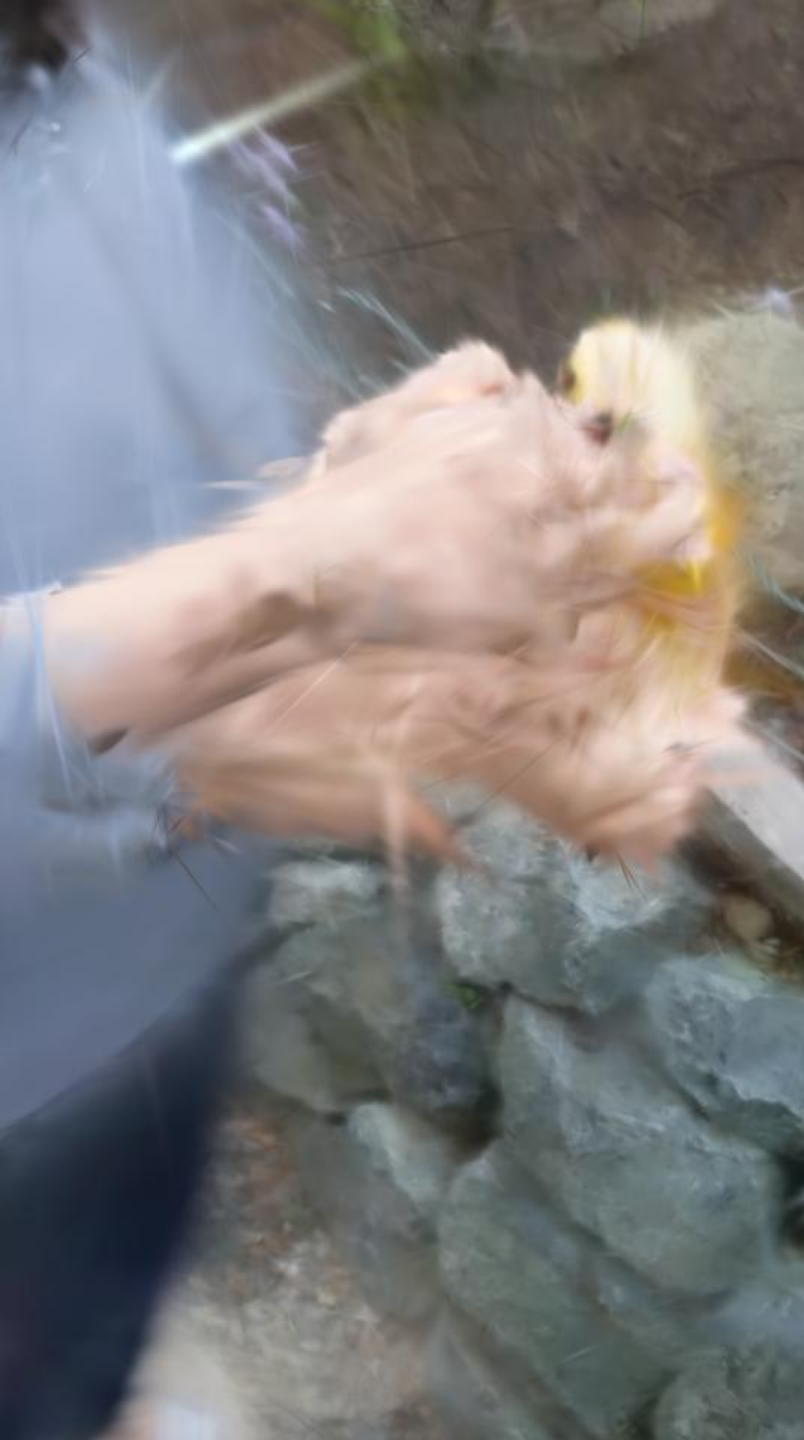}
        \caption*{(g) Canonical}
        \label{fig:canonical_banana} % Corrected label
    \end{subfigure}
    \hfill
    \begin{subfigure}{0.24\linewidth}
        \centering
        \includegraphics[width=\linewidth]{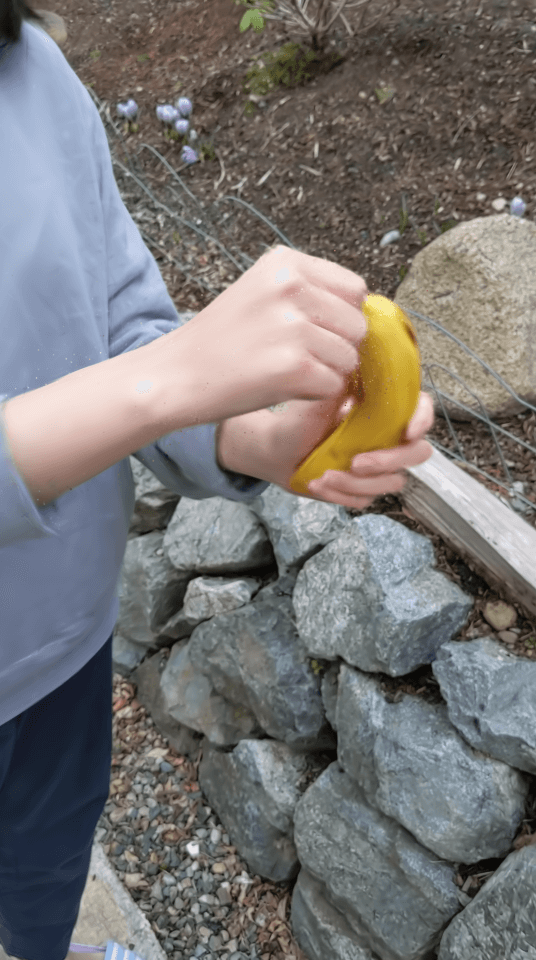}
        \caption*{(h) GT}
        \label{fig:gt_banana} % Corrected label
    \end{subfigure}
    
    \caption{
        \textbf{Visualization of the Learned Canonical Fields.}
        Direct renderings of the undeformed canonical fields from scenes in the N3DV (\textit{flame\_salmon}, \textit{cook\_spinach}) and HyperNeRF (\textit{chicken}, \textit{peel-banana}) datasets. 
        Each canonical rendering is paired with a corresponding ground truth frame. 
        The high visual similarity across both multiview and monocular datasets validates our assumption that the learned canonical field provides a stable and coherent representation of the scene, forming a solid basis for our sampling strategy.
    }
    \label{fig:canonical_ob_appendix_all}
\end{figure}

\paragraph{Observations on the Canonical Field} The viability of our sampling strategy hinges on the assumption that the learned canonical field provides a stable and coherent representation of the scene. To validate this, we present direct renderings of the undeformed canonical fields from various scenes in Fig.~\ref{fig:canonical_ob_appendix_all}. The high visual similarity between the canonical renderings and the ground truth frames, across both multiview (N3DV) and monocular (HyperNeRF) datasets, strongly supports our assumption and provides a solid basis for our sampling strategy.

% ========== START OF THE UPDATED FIGURE WITH SPACING CONTROL ==========
% ========== END OF THE UPDATED FIGURE ==========

\paragraph{Locating Problematic Coordinates}
To robustly identify regions requiring densification, our first step is to compute a reliable depth map for backprojection. We compared two common strategies: mean depth and median depth. Mean depth is a weighted average of all semi-transparent 3DGs along a ray, formulated as:
\begin{equation}
    D_{\text{mean}} = \sum_{i=1}^{N} \|\mu_i - o\|_2 \cdot \alpha'_{i} T_{i-1}.
\end{equation}
In contrast, median depth is determined by a single surface-like 3DG, identifying the first one to cross an accumulated transmittance threshold. Its formulation is:
\begin{equation}
D_{\text{median}} = \left\{
    \begin{array}{ll}
        \|\mu_i - o\|_2, & \text{if } \exists G_i \text{ s.t. } T_{i-1} > 0.5 \text{ and } T_i \le 0.5 \\
        \text{nan},     & \text{otherwise}
    \end{array}
\right.
\end{equation}
where $o$ is the camera center, and $T_i = \prod_{j=1}^{i} (1 - \alpha'_{j})$ is the accumulated transmittance.
\par
\begin{figure}[h]
    \centering
    
    % --- ROW 1: (a) GT and (b) Pseudo-GT Depth ---
    \begin{subfigure}{0.48\linewidth}
        \centering
        \includegraphics[width=\linewidth,
        trim=0pt 0pt 000pt 200pt,  % 左、下、右、上
        clip
        ]{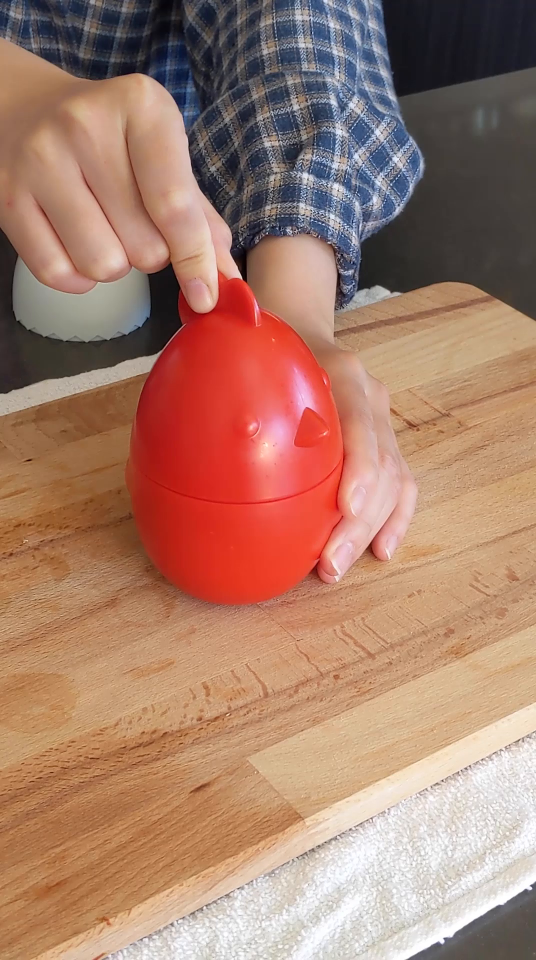}
        \caption*{(a) GT}
    \end{subfigure}
    \hfill
    \begin{subfigure}{0.48\linewidth}
        \centering
        \includegraphics[width=\linewidth,
        trim=0pt 0pt 000pt 200pt,  % 左、下、右、上
        clip]{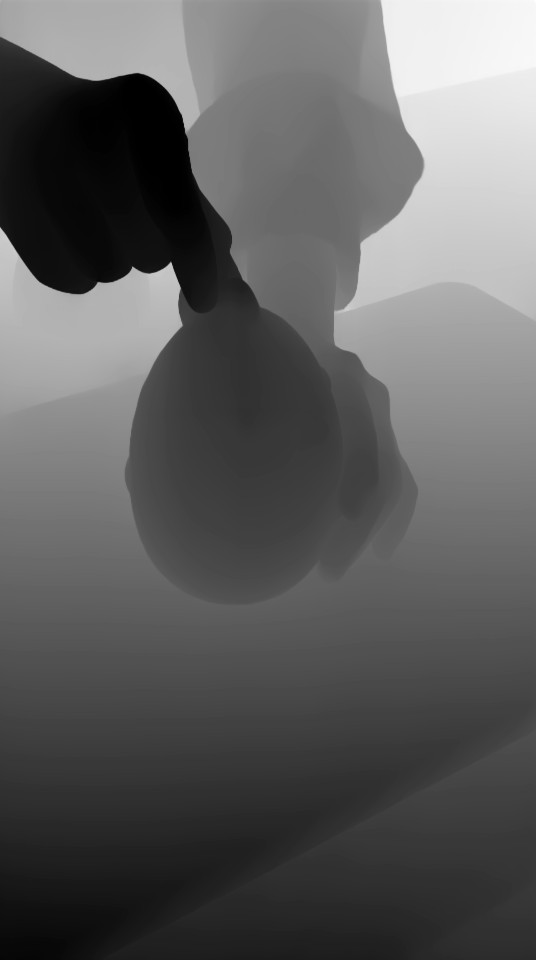}
        \caption*{(b) Pseudo-GT Depth}
    \end{subfigure}
    
    % --- VERTICAL SPACE CONTROL ---
    % Adjust '3mm' to your desired spacing.
    \vspace{2mm} 

    % --- ROW 2: (c) Median Depth and (d) Mean Depth ---
    \begin{subfigure}{0.48\linewidth}
        \centering
        \includegraphics[width=\linewidth,
        trim=0pt 0pt 000pt 200pt,  % 左、下、右、上
        clip]{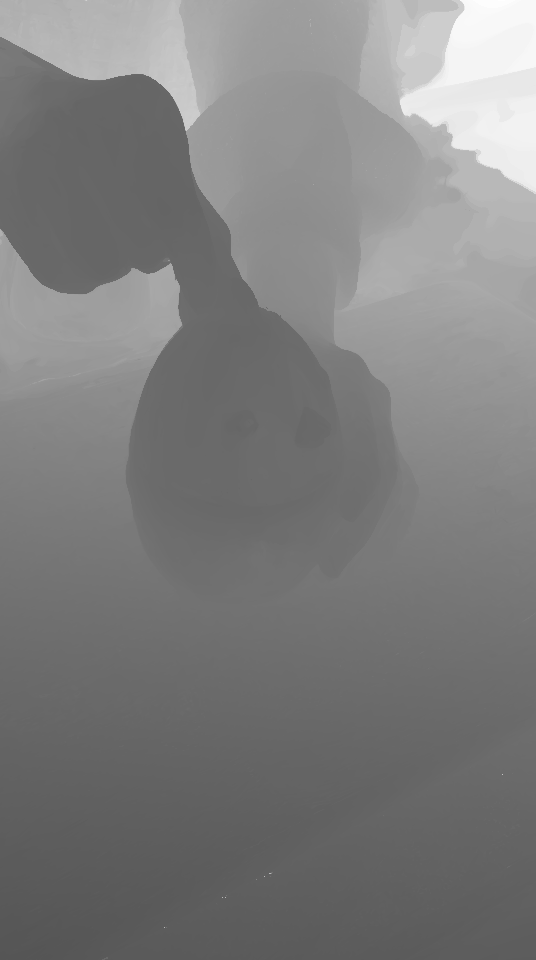}
        \caption*{(c) Median Depth}
    \end{subfigure}
    \hfill
    \begin{subfigure}{0.48\linewidth}
        \centering
        \includegraphics[width=\linewidth,
        trim=0pt 0pt 000pt 200pt,  % 左、下、右、上
        clip]{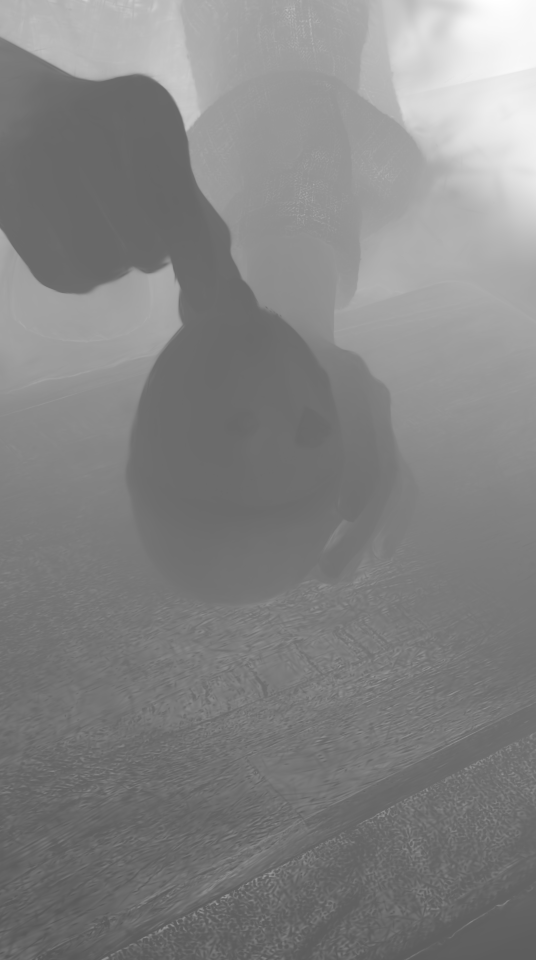}
        \caption*{(d) Mean Depth}
    \end{subfigure}
    
    \caption{
        \textbf{Comparison of depth computation methods.} 
        The visual results on the \textit{chicken} scene clearly show that our chosen median depth (c) produces a cleaner depth map with fewer floaters compared to mean depth (d). Its structure aligns better with the pseudo-GT depth (b), which was generated by the SOTA model Depth Anything~\cite{yang2024depthanythingunleashingpower}.
    }
    \label{fig:median_mean_new}
\end{figure}

The fundamental difference in their definitions leads to a significant disparity in robustness. As visually demonstrated in Fig.~\ref{fig:median_mean_new}, the averaging nature of mean depth makes it heavily corrupted by ``floater'' artifacts caused by semi-transparent structures. Median depth, however, correctly identifies the primary surface and produces a clean, accurate representation. Based on this clear advantage in robustness, we select median depth as the foundation for our sampling strategy.
\par
After establishing a reliable depth map, we compute the L1 error map $M_{\text{error}} = |M_{\text{render}} - M_{\text{gt}}|$ between the rendered image and the ground truth. By selecting pixels with high error values and backprojecting them, we obtain a set of 3D coordinates in the deformed space that correspond to rendering failures.

% ========== START OF THE UPDATED FIGURE WITH VERTICAL SPACING ==========
% ========== END OF THE UPDATED FIGURE ==========

\paragraph{Low-Cost Anchor Injection in Canonical Field}
Having identified the problematic 3D coordinates in the deformed space, the next question is how to use them to add new 3DGs to the canonical field. A seemingly intuitive approach would be to learn an inverse deformation to map these coordinates back. Such a process would typically involve: (1) finding the nearest 3DG in the deformed space to the problematic coordinate, (2) copying its embedding $e_c$, and (3) using this copied embedding to compute an inverse deformation back to a ``theoretically correct'' position $p_c$ in the canonical field. 
However, this strategy is fundamentally flawed. In sparse regions, it is forced to copy an embedding $e_c$ from a potentially distant and unrelated 3DG. This makes the subsequent step—solving an inverse problem ($p_c + \Delta \mu = p_d$)—highly unstable, as the process is ill-posed and extremely sensitive to the quality of this likely inaccurate embedding. Consequently, the resulting canonical position is unreliable, making the pursuit of ``theoretical precision'' misguided.
\par
Therefore, we adopt a more direct and robust strategy: using the problematic coordinate as a \textbf{low-cost initial anchor} for a new 3DG directly in the canonical field. This drastically simplifies the learning task. The optimization can more easily learn a local refinement for a nearby anchor and its embedding, rather than a distant, mismatched one. Specifically, these anchors are subsequently refined through joint optimization with the deformation network to ensure sequence-wide geometric consistency. As demonstrated in Fig.~\ref{fig:canonical_ob_appendix_all}, the inherent stability of the learned canonical field provides a solid and reliable foundation for such local refinements. 
\par
Finally, we initialize the attributes for this new 3DG. We compute the scaling factor $s$ through a logarithmic transformation of the nearest neighbor distance, set the rotation $q$ to the identity quaternion, initialize the opacity $\sigma$ to 0.1, and derive the spherical harmonic coefficients $sh$ from the pixel's RGB value. The per-Gaussian embedding is initialized by copying from the existing 3DG with the minimum total attribute deformation, providing a stable starting point for optimization.
\subsection{Aggressive Opacity Reduction Strategy}\label{sec:opacity_reduction}
% ========== START OF THE ALIGNED FIGURE ==========
\begin{figure}[t!] % Changed to [t!] for better placement
    \centering
    
    % --- ROW 1: (a) and (b) ---
    \begin{subfigure}{0.48\linewidth} % Adjusted width for better spacing
        \includegraphics[width=\linewidth]{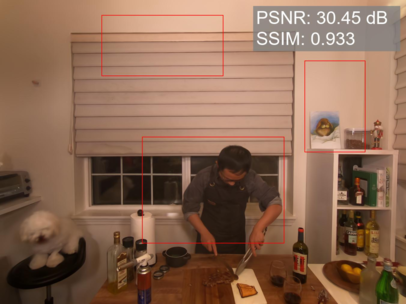}
        \caption*{(a) Original Model}
    \end{subfigure}
    \hfill % This is the key: creates flexible space to push (b) to the right
    \begin{subfigure}{0.48\linewidth}
        \includegraphics[width=\linewidth]{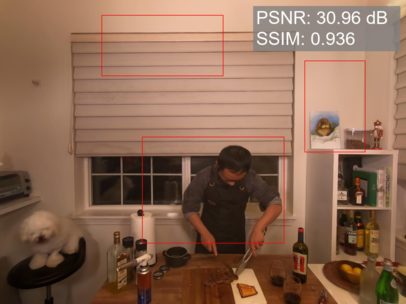}
        \caption*{(b) Opacity Filter}
    \end{subfigure}
    
    % Use \vspace{} for consistent spacing between rows
    \vspace{2mm} 
    
    % --- ROW 2: (c) and (d) ---
    \begin{subfigure}{0.48\linewidth}
        \includegraphics[width=\linewidth]{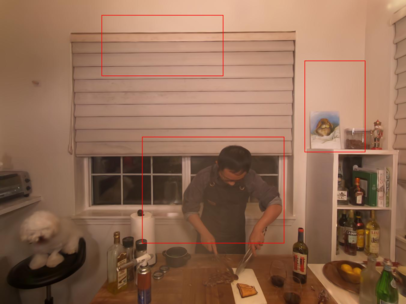}
        \caption*{(c) No Opacity Deform}
    \end{subfigure}
    \hfill % This pushes (d) to the right, aligning it with (b)
    \begin{subfigure}{0.48\linewidth}
        \includegraphics[width=\linewidth]{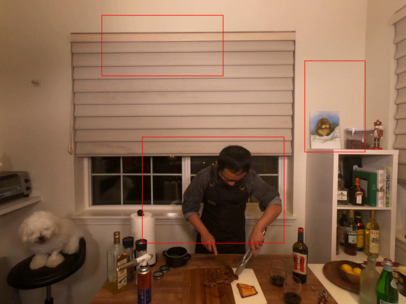}
        \caption*{(d) GT}
    \end{subfigure}
    
    \vspace{1mm} 
    
    % --- ROW 3: Heatmap ---
    \begin{subfigure}{0.97\linewidth} % Slightly less than 1.0 to avoid overfull box warnings
        \centering
        \includegraphics[width=1.0\linewidth]{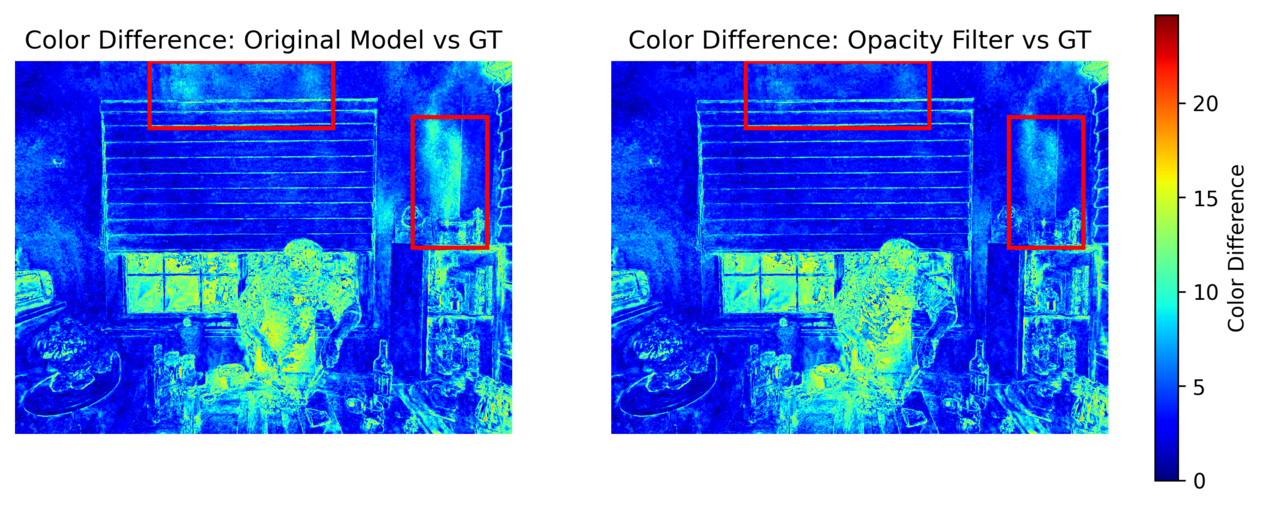}
        \vspace{-6mm}
        \caption*{(e) Color Difference Heatmap}
    \end{subfigure}
    \caption{\textbf{Observation of 3DGs in dim scenes}.}
    \label{fig:opacity_reduction_teaser}
\end{figure}
% ========== END OF THE ALIGNED FIGURE ==========
When conducting experiments on dim scenes, such as \textit{cut roasted beef} in the N3DV\cite{DyNeRF} dataset,  we observed that E-D3DGS tends to utilize 3DGs near the camera to model shadows. These foreground shadow 3DGs often introduce significant color bias in test views, manifesting as a substantial drop in Peak Signal-to-Noise Ratio (PSNR). To investigate this issue, we disabled opacity deformation during rendering for analysis. As shown in Fig. \ref{fig:opacity_reduction_teaser}, the No Opacity Deform variant shows that shadow 3DGs in the foreground blur the rendering view. In contrast, while the Original Model (with standard deformation) still exhibits color discrepancy compared to Ground Truth, this discrepancy is significantly reduced relative to the No Opacity Deform variant. This observation confirms that partial foreground shadow 3DGs undergo opacity reduction when processed by the deformation field. To verify whether these opacity-reduced shadow 3DGs indeed harm rendering quality, we implemented the Opacity Filter variant, which removes 3DGs with opacity below 0.05 during rendering. This filtering operation markedly reduces color discrepancy and improves visual quality.
\par
Through these experiments, we have established that:
(1) Partial foreground shadow 3DGs experience opacity reduction during deformation;
(2) These shadow 3DGs adversely affect rendering quality.
To minimize their impact, we take advantage of their natural opacity reduction during deformation and introduce an enhanced opacity attenuation function to further decrease their visibility:
\begin{equation}     \alpha_{final} = \phi(\phi^{-1}(\alpha) + \Delta\alpha) \cdot \phi(k\Delta\alpha),     \label{eq:alpha_final} \end{equation}

Here, $\alpha_{final}$ is the final opacity for rendering, $\Delta\alpha$ is the opacity variation, $\phi$ is the Sigmoid function, and $k$ is a hyperparameter. Intuitively, the multiplicative term $\phi(k\Delta\alpha)$ acts as an additional penalty factor. In dynamic scene reconstruction, foreground shadow artifacts typically manifest as shadow 3DGs whose canonical opacity $\alpha$ is significantly larger than their deformed opacity. When the deformation field predicts a substantial negative variation, the Sigmoid function $\phi(k\Delta\alpha)$ operates in its rapid-decay region, imposing a decisive multiplicative penalty on the final opacity. This formulation ensures that these shadow 3DGs are assigned lower rendering weights much more aggressively than through standard updates, effectively mitigating shadow floaters while preserving the reconstruction of stable, high-fidelity structures.
\subsection{Optimization}\label{Optimization}
We employed a loss function similar to E-D3DGS, which consists of a combination of the L1 loss, an intermittent D-SSIM loss, and a local smoothness regularization. The term ``intermittent'' means that after 10K training steps, we apply the D-SSIM loss for five iterations every 50 iterations. The local smoothness regularization encourages adjacent Gaussians in the canonical space to have similar per-Gaussian embeddings, and is formulated as:
\begin{equation}
    L_{\text{emb-reg}} = \frac{1}{k|S|}\sum_{i \in S}\sum_{j \in \text{KNN}_{k}(i)}(w_{i,j} \| e_{i} - e_{j} \|_2),
\end{equation}
where $S$ is the set of 3DGs, $\text{KNN}_{k}(i)$ denotes the $k$ nearest neighbors of the $i$-th Gaussian, and $w_{i,j} = \exp(-\lambda_{w} \| \mu_j - \mu_i \|_2^2)$. To reduce computational overhead, we compute the KNN only when densification occurs.

\begin{table*}[t]
  \centering
  \caption{
    \textbf{Quantitative comparison on the N3DV dataset.}
    Storage, training time, and FPS are calculated on \textit{flame\_salmon\_frag1}.
    $^\dagger$For a fairer comparison under our setting, ST-GS was evaluated using a single first-frame COLMAP initialization. 
  }
  \label{tab:N3DV_Comparisons_Avg}
  
  % --- KEY CHANGE: Use tabular* to span the full text width ---
  % We remove \setlength{\tabcolsep} and let @{\extracolsep{\fill}} handle the spacing automatically.
  \begin{tabular*}{\textwidth}{@{\extracolsep{\fill}}lcccccc} 
    \toprule
    Method & PSNR$\uparrow$ & SSIM$\uparrow$ & LPIPS$\downarrow$ & Storage$\downarrow$ & Training Time$\downarrow$ & FPS$\uparrow$ \\
    \midrule
    Mix Voxels~\cite{mixedvoxel}              & 30.30           & 0.918 & 0.127 & 512 MB                         & 1h 28m                    & 1.01                  \\
    K-Planes~\cite{K-Planes}                &  30.86           & 0.939 & 0.096 & 309 MB                         & 1h 33m                    & 0.15                  \\
    D3DGS~\cite{deformable3dgs}                   & 28.27           & 0.917 & 0.156 & 75MB                        & 2h 17m                    & 20.29                   \\
    ST-GS$^\dagger$~\cite{SpacetimeGaussians}         & 28.94           & 0.920 & 0.078 & 49 MB                         & 4h 25m                    & 124                     \\
    Swift4D~\cite{wu2025swift4dadaptivedivideandconquergaussiansplatting}                 & 30.05           & 0.931 & 0.055 & 116MB                       & 48m      & 138.00 \\
    4DGS~\cite{4dgs_towards}                    & 30.30           & 0.933 & 0.069 & 3.6GB                       & 7h 43m                    & 54.36                   \\
    4DGaussians~\cite{4dgs_kplanes}             & 30.19           & 0.917 & 0.061 & 53 MB    & 1h 13m                    & 78.28                   \\
    E-D3DGS~\cite{ed3dgs}                 & 30.79           & 0.937 & 0.051 & 73 MB             & 2h 41m                    & 40.52                   \\
    Ours                    & 31.01           & 0.941 & 0.050 & 69 MB             & 1h 56m                    & 91.13                   \\
    \bottomrule
  \end{tabular*}
\end{table*}
% ========== END OF FINAL TABLE ==========
\section{Experiments}
We first elaborate on the selection of the baselines and evaluation metrics. Then, we compare our method with baselines on different datasets in Sec.~\ref{dataset and comparison}. Finally, we conduct ablation studies and analysis on our method in Sec.~\ref{sec: evaluation}. Furthermore, we provide more results in the appendix.
\par
% \textbf{Baselines}. In NeRF-based methods, we compare with Mix Voxels \cite{mixedvoxel}, K-Planes \cite{K-Planes}, HyperNeRF \cite{Hypernerf}, Nerfies \cite{Nerfies}, TiNeuVox \cite{TiNeuVox}. For 3DG-S-based methods, we select deformation field-based methods such as D3DGS \cite{deformable3dgs}, Swift4D~\cite{wu2025swift4dadaptivedivideandconquergaussiansplatting}, 4DGaussians \cite{4dgs_kplanes}, DN-4DGS \cite{dn4dgs}, and E-D3DGS \cite{ed3dgs}. We also include 4DGS-based method 4DGS \cite{realtime4dgs}. 
% We reference the quality metrics of NeRF-based methods directly from the E-D3DGS paper~\cite{ed3dgs}. For 3DG-S-based methods, we conducted systematic evaluations under uniform point cloud initialization conditions to ensure fairness. We are uncertain whether the discrepancy stems from variations in initialization or parameter configurations, as we were unable to reproduce most of the results reported in E-D3DGS. While ST-GS~\cite{SpacetimeGaussians} is omitted from our main tables to avoid an unfair comparison stemming from its per-frame SfM dependency, we present a comparative analysis under aligned settings in the appendix. 
\textbf{Baselines}. In NeRF-based methods, we compare with Mix Voxels~\cite{mixedvoxel}, K-Planes~\cite{K-Planes}, HyperNeRF~\cite{Hypernerf}, Nerfies~\cite{Nerfies}, and TiNeuVox~\cite{TiNeuVox}. For 3DG-S-based methods, we select deformation field-based methods such as D3DGS~\cite{deformable3dgs}, Swift4D~\cite{wu2025swift4dadaptivedivideandconquergaussiansplatting}, 4DGaussians~\cite{4dgs_kplanes}, DN-4DGS~\cite{dn4dgs}, and E-D3DGS~\cite{ed3dgs}. We also include the 4DGS-based method 4DGS~\cite{realtime4dgs} and curve-based method ST-GS~\cite{SpacetimeGaussians}.
\par
For NeRF-based methods, we reference the quality metrics of NeRF-based methods directly from the E-D3DGS paper~\cite{ed3dgs}. For 3DG-S-based methods, our primary strategy is a strictly aligned comparison: acknowledging the high sensitivity of these methods to the initial point cloud, we evaluate them under a unified initialization condition. Specifically, all methods use the initial point cloud generated by the script provided by 4DGaussians~\cite{4dgs_kplanes}. For certain baselines that are difficult to reproduce due to unavailable code or prohibitive computational costs, we replicate the experimental setup reported in their papers and run our method under their specific settings for a direct comparison. All results obtained under these special settings are presented separately in our main comparison tables to ensure full transparency.
% we conducted systematic evaluations under uniform point cloud initialization conditions to ensure fairness. A direct comparison with ST-GS is challenging due to its reliance on per-frame SfM initializations, which provides a significant advantage. Therefore, to ensure a more equitable evaluation, we re-ran both ST-GS and our method under an aligned setting using a single, first-frame initialization. These specific results are presented separately in our main comparison table to ensure transparency.
\par
\textbf{Metrics}. We use PSNR, SSIM, and LPIPS (based on AlexNet) to evaluate the quality of rendered images: PSNR measures pixel color error, SSIM assesses perceptual similarity, and LPIPS evaluates high-level perceptual similarity. Higher PSNR and SSIM values, along with lower LPIPS values, indicate better visual quality. Additionally, we focus on rendering speed, training time, and storage.
\begin{figure*}[t!]
    \centering

    % --- ROW 1: flame_salmon scene ---
    % Widths are adjusted to slightly less than 1/5 of the total width.
    % \hfill automatically fills the remaining space, stretching the row.
    \begin{subfigure}[b]{0.19\linewidth}
      \includegraphics[width=\linewidth]{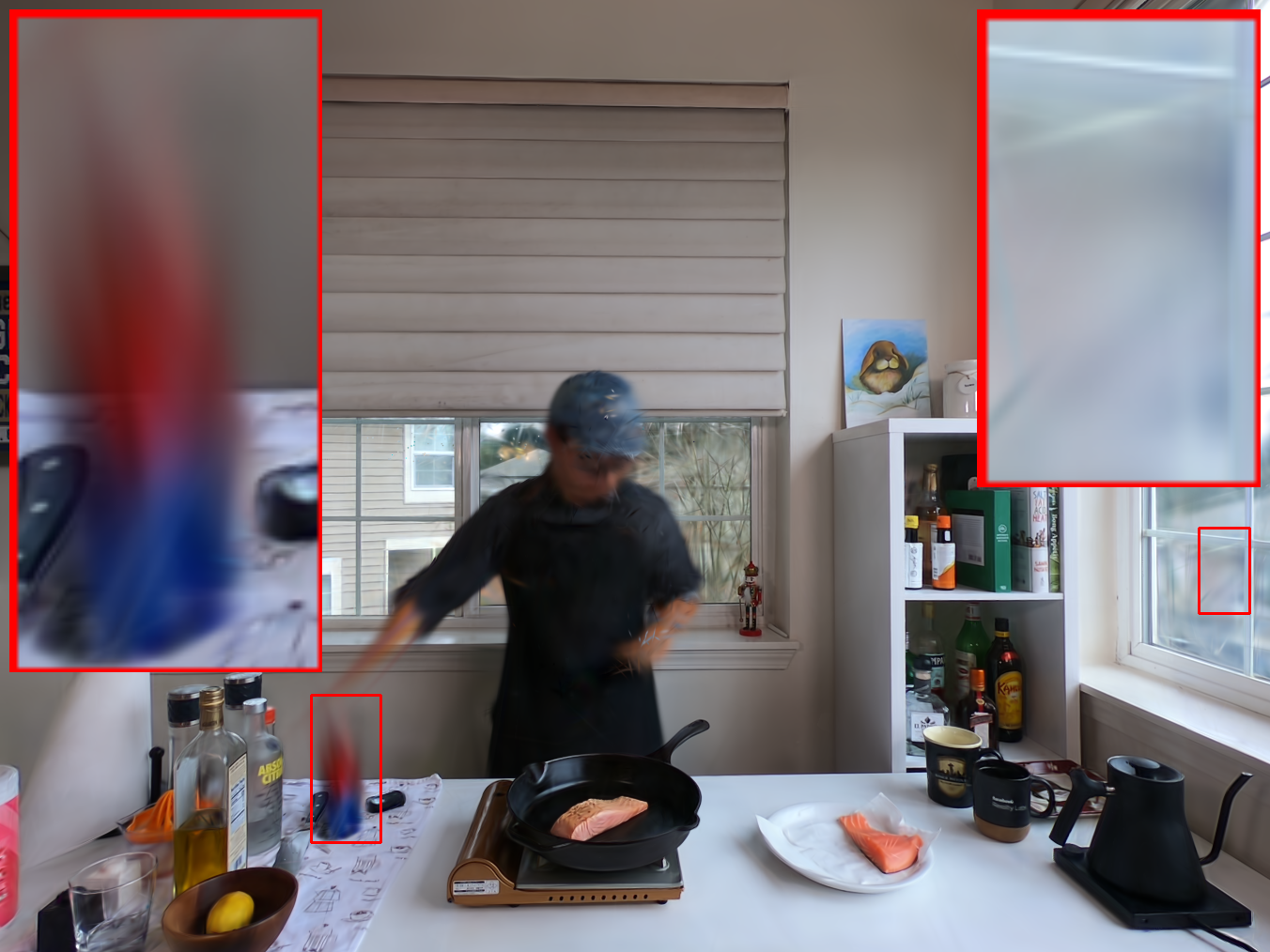}
    \end{subfigure}
    \hfill
    \begin{subfigure}[b]{0.19\linewidth}
        \includegraphics[width=\linewidth]{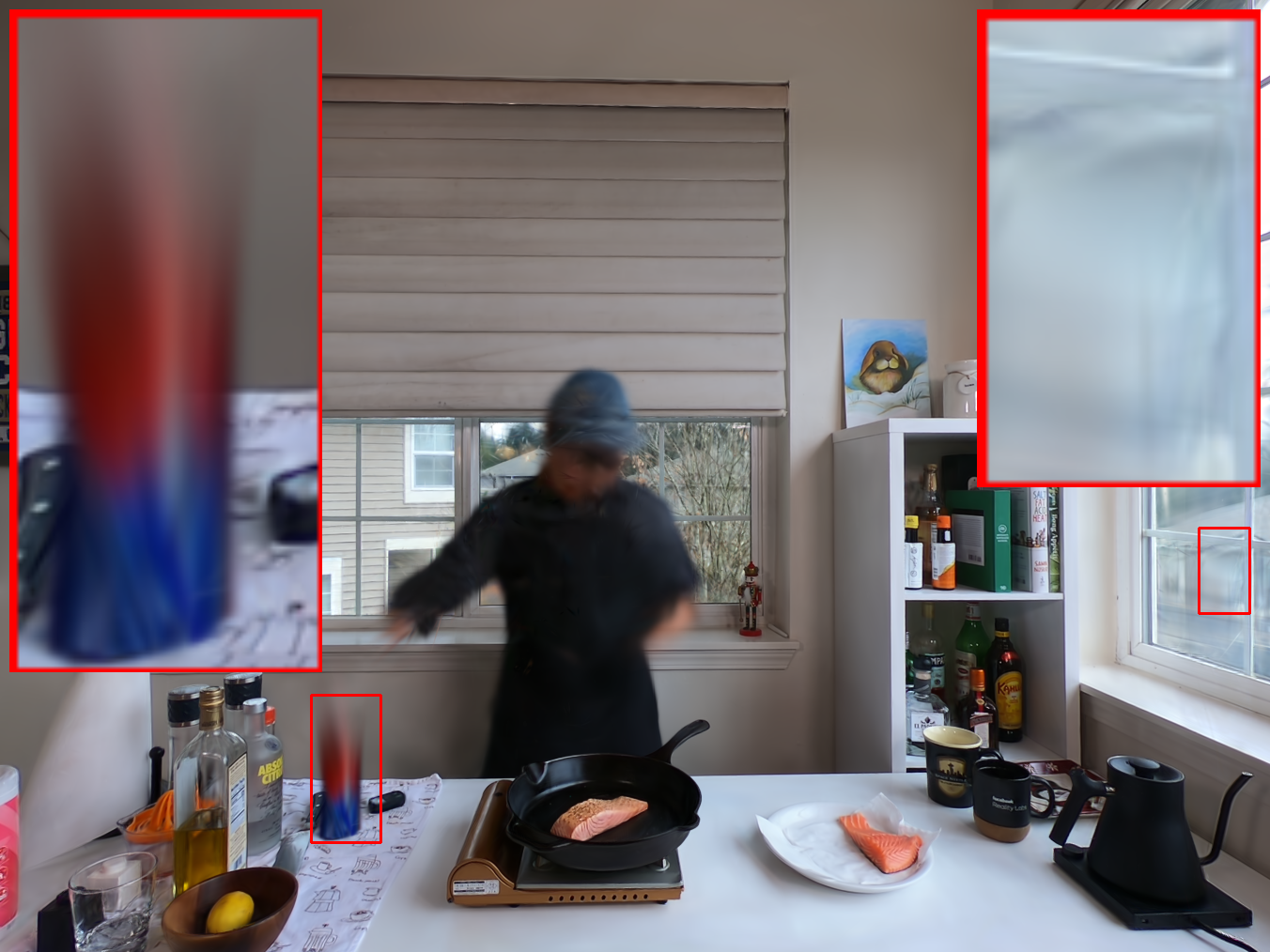}
    \end{subfigure}
    \hfill
    \begin{subfigure}[b]{0.19\linewidth}
        \includegraphics[width=\linewidth]{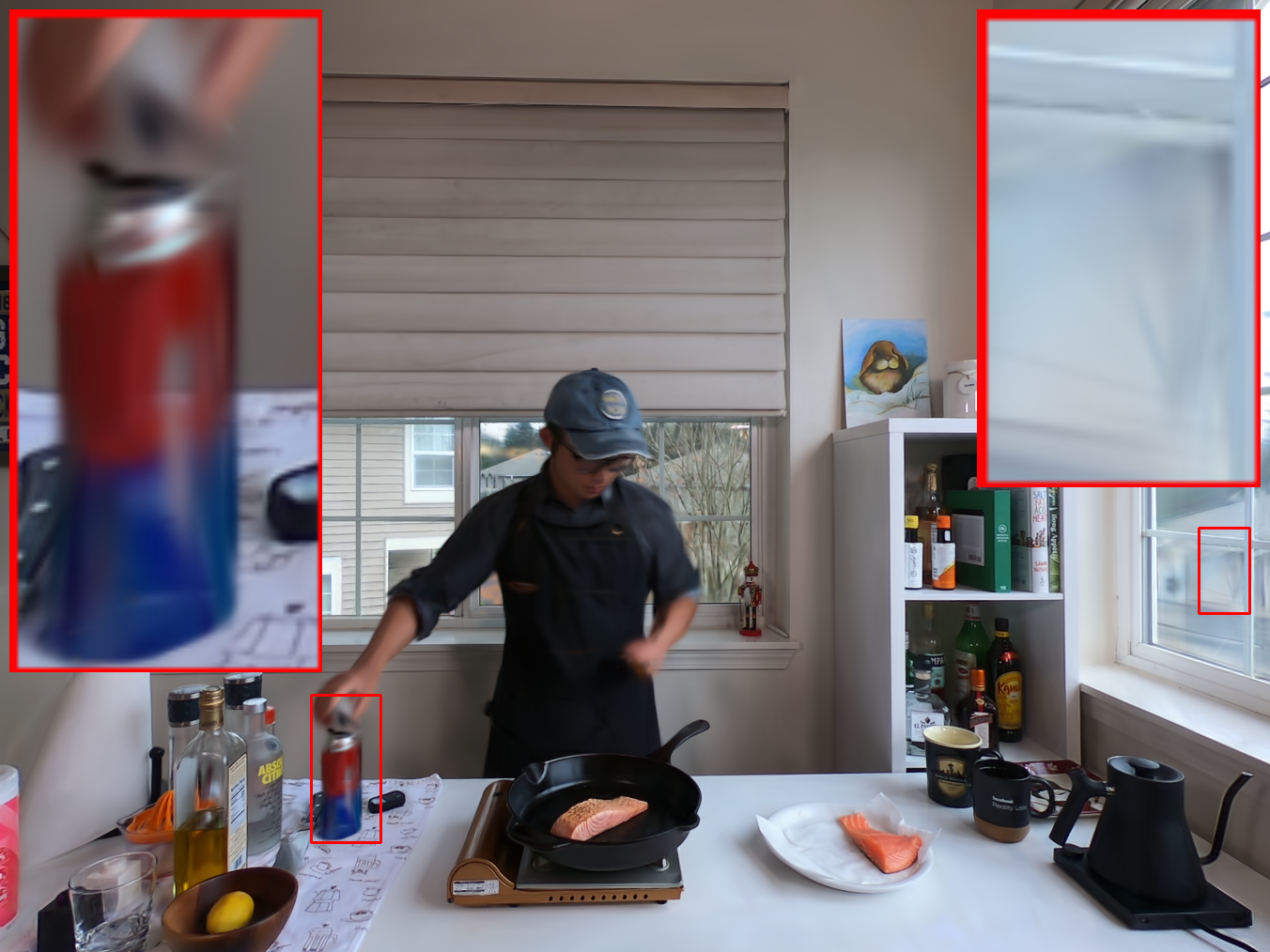}
    \end{subfigure}
    \hfill
    \begin{subfigure}[b]{0.19\linewidth}
        \includegraphics[width=\linewidth]{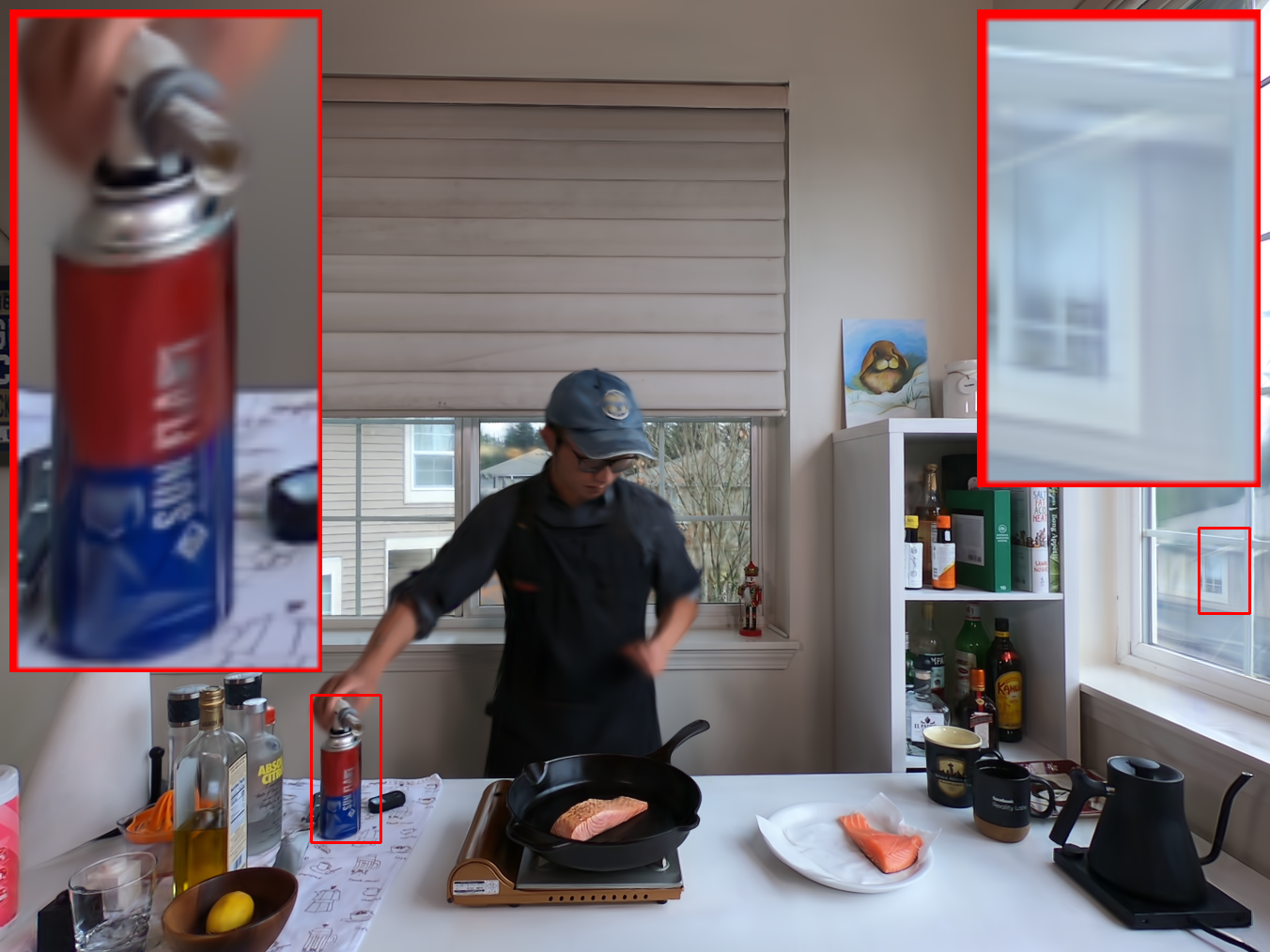}
    \end{subfigure}
    \hfill
    \begin{subfigure}[b]{0.19\linewidth}
        \includegraphics[width=\linewidth]{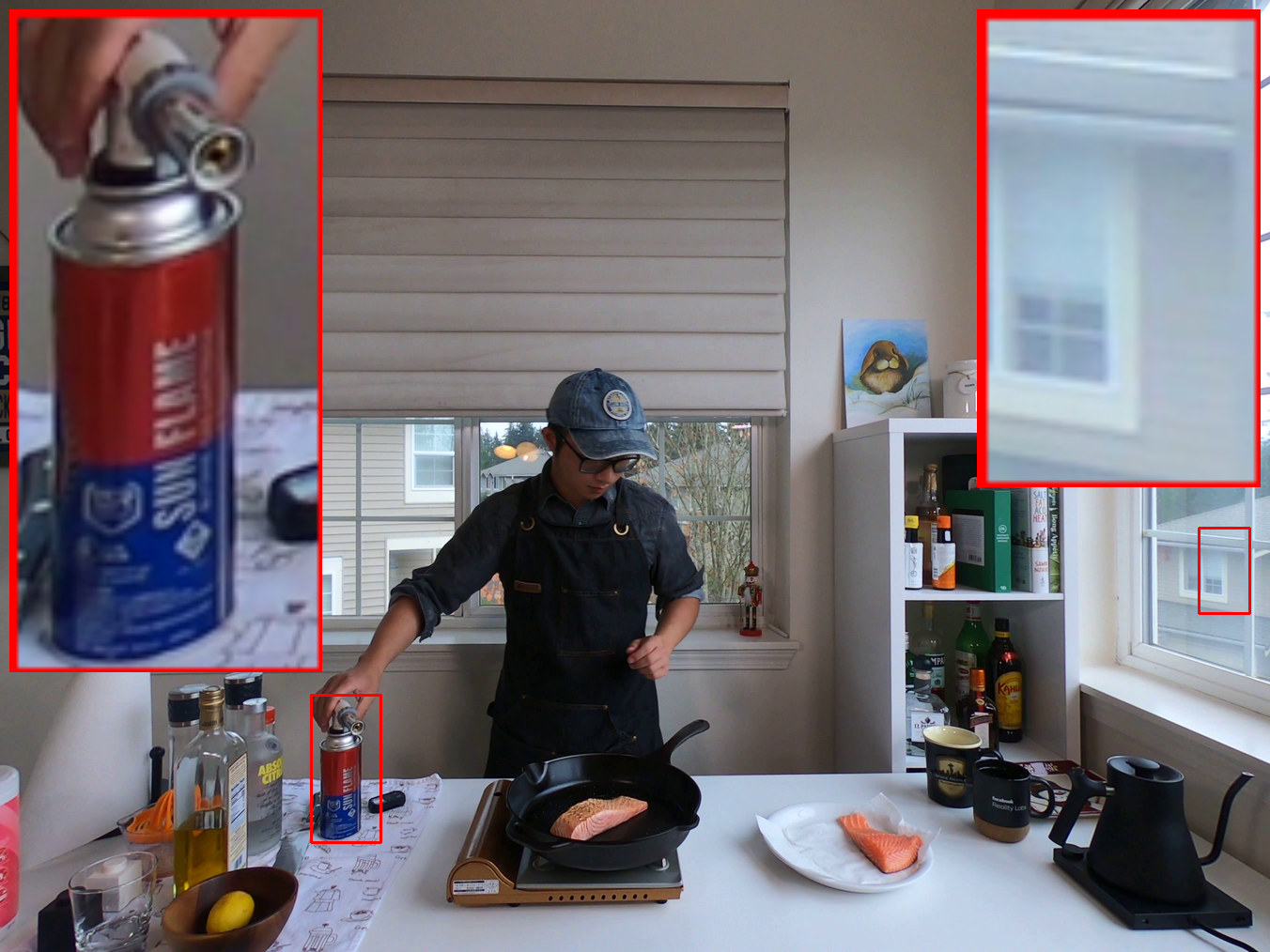}
    \end{subfigure}

    \vspace{2mm} % Vertical space between rows

    % --- ROW 2: cut_roasted_beef scene ---
    \begin{subfigure}[b]{0.19\linewidth}
      \includegraphics[width=\linewidth]{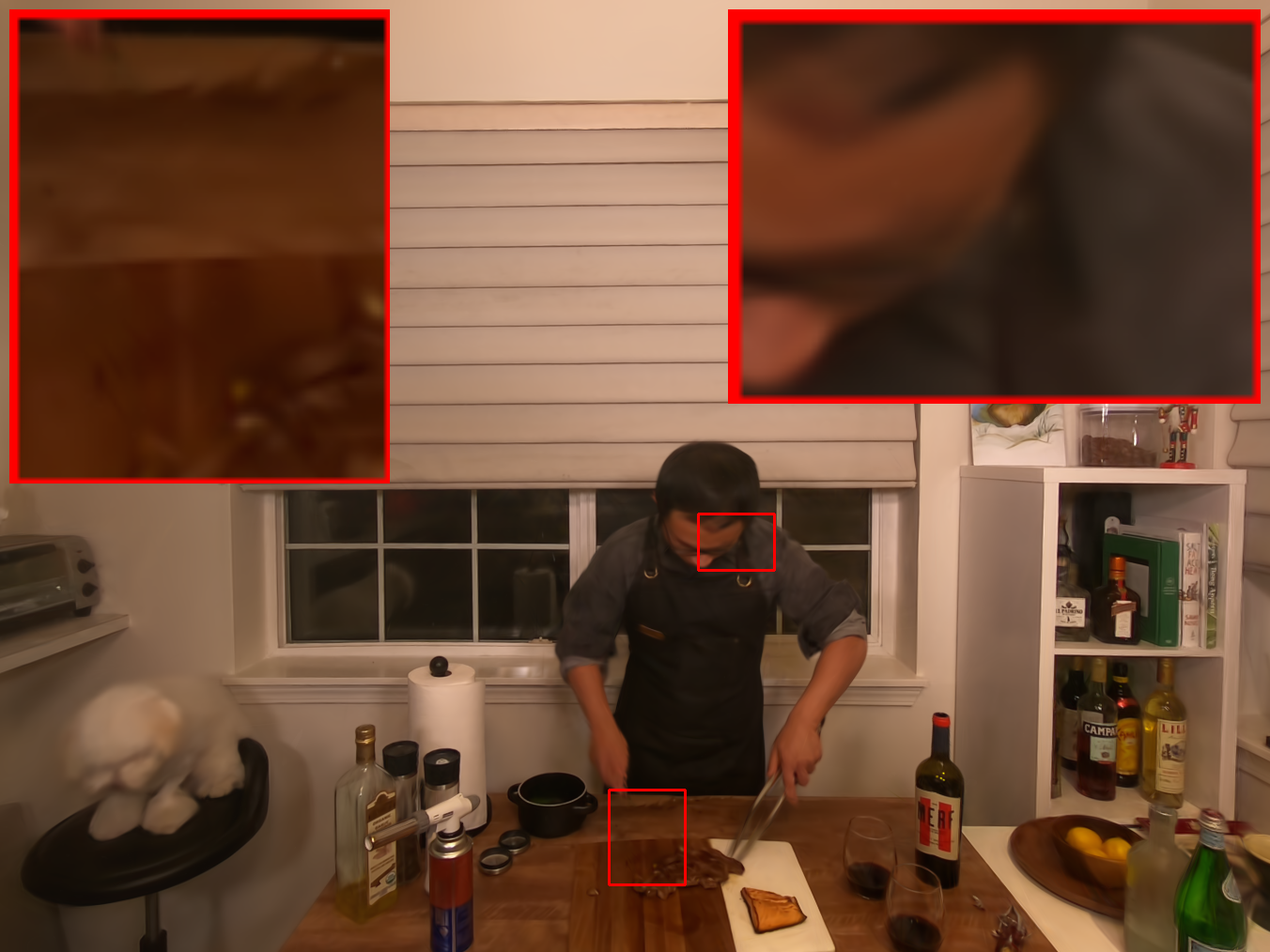}
        \caption*{(a) 4DGaussians~\cite{4dgs_kplanes}}
    \end{subfigure}
    \hfill
    \begin{subfigure}[b]{0.19\linewidth}
        \includegraphics[width=\linewidth]{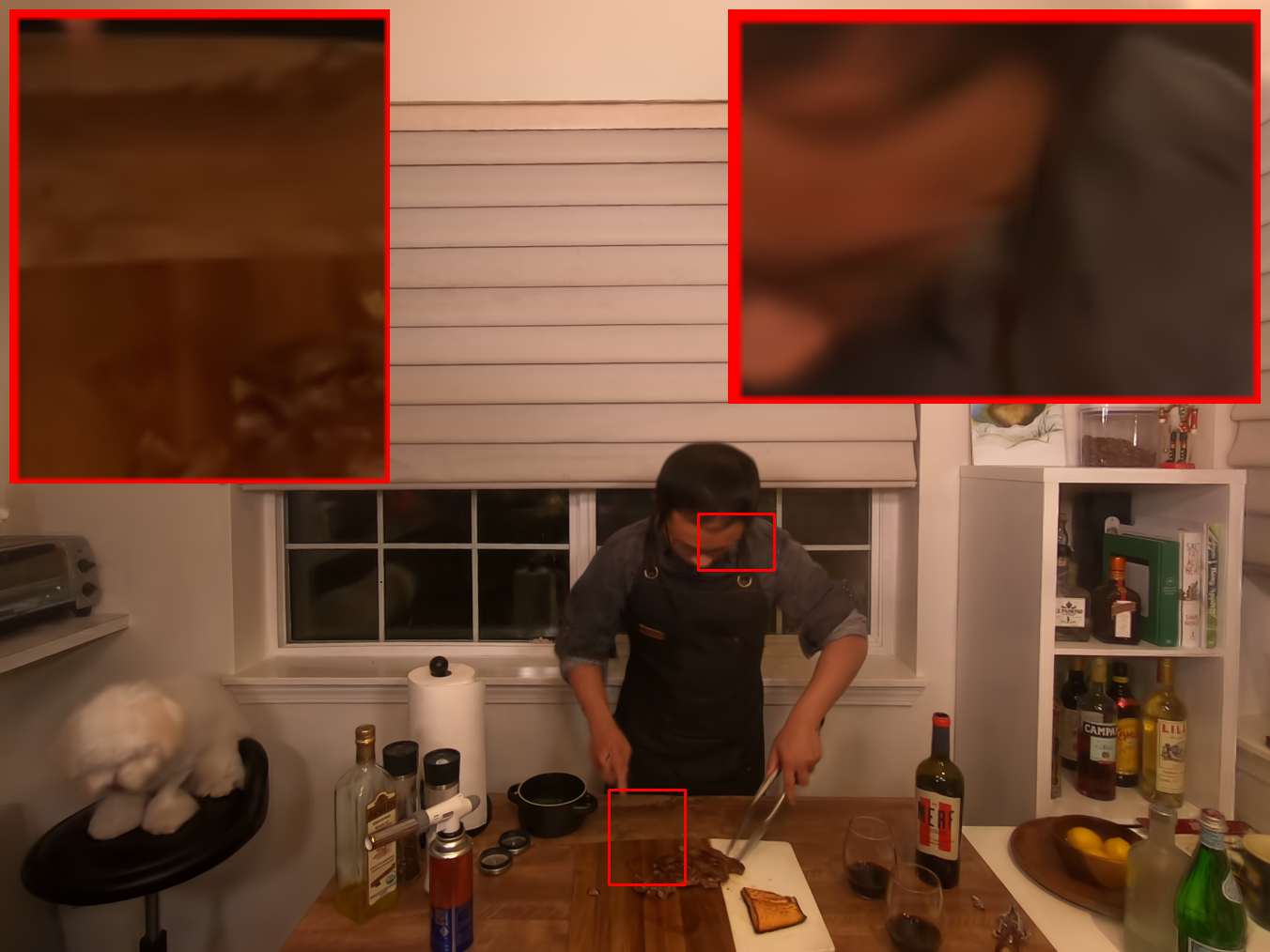}
        \caption*{(b) Swift4D~\cite{wu2025swift4dadaptivedivideandconquergaussiansplatting}}
    \end{subfigure}
    \hfill
    \begin{subfigure}[b]{0.19\linewidth}
        \includegraphics[width=\linewidth]{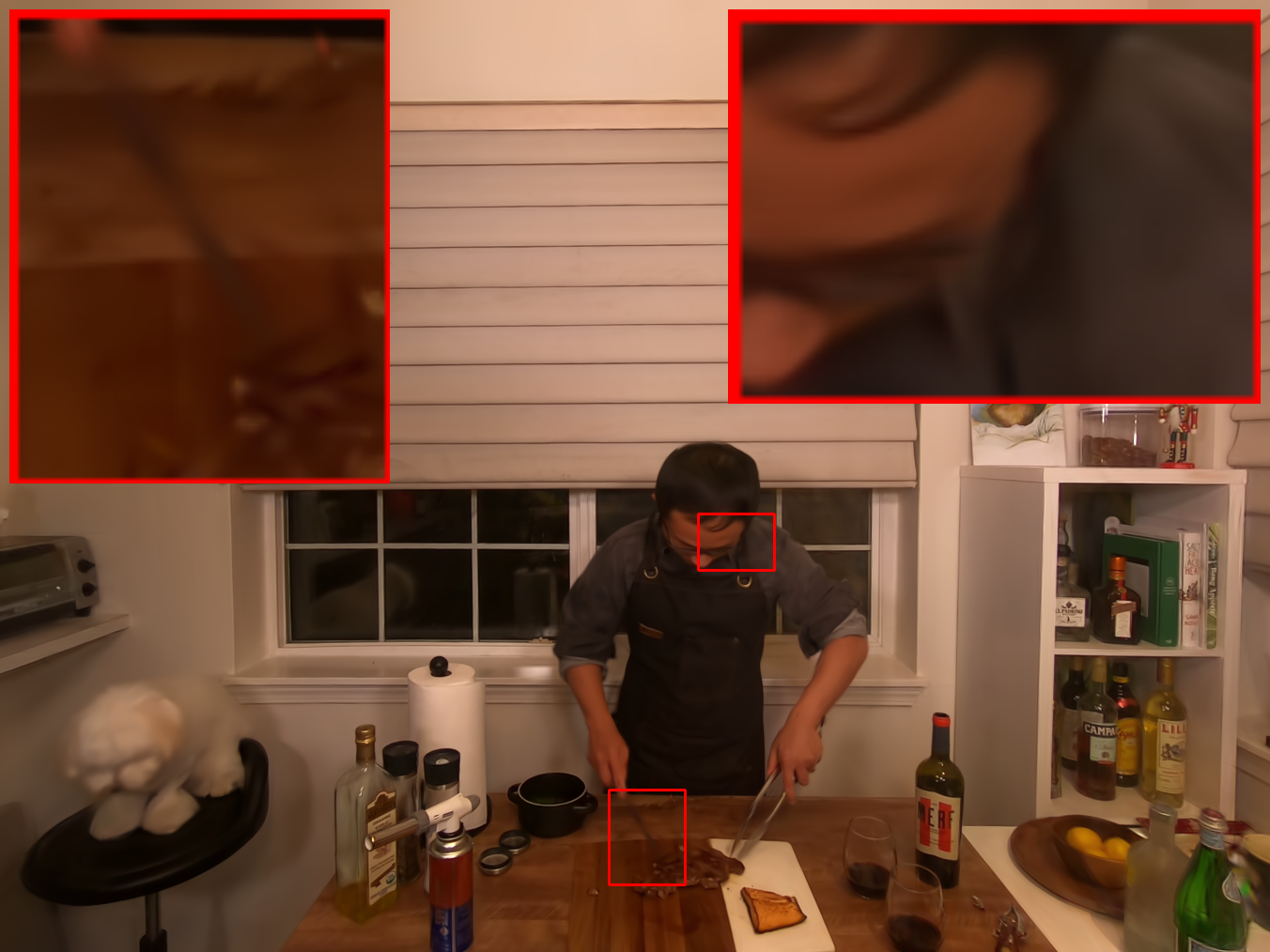}
        \caption*{(c) E-D3DGS~\cite{ed3dgs}}
    \end{subfigure}
    \hfill
    \begin{subfigure}[b]{0.19\linewidth}
        \includegraphics[width=\linewidth]{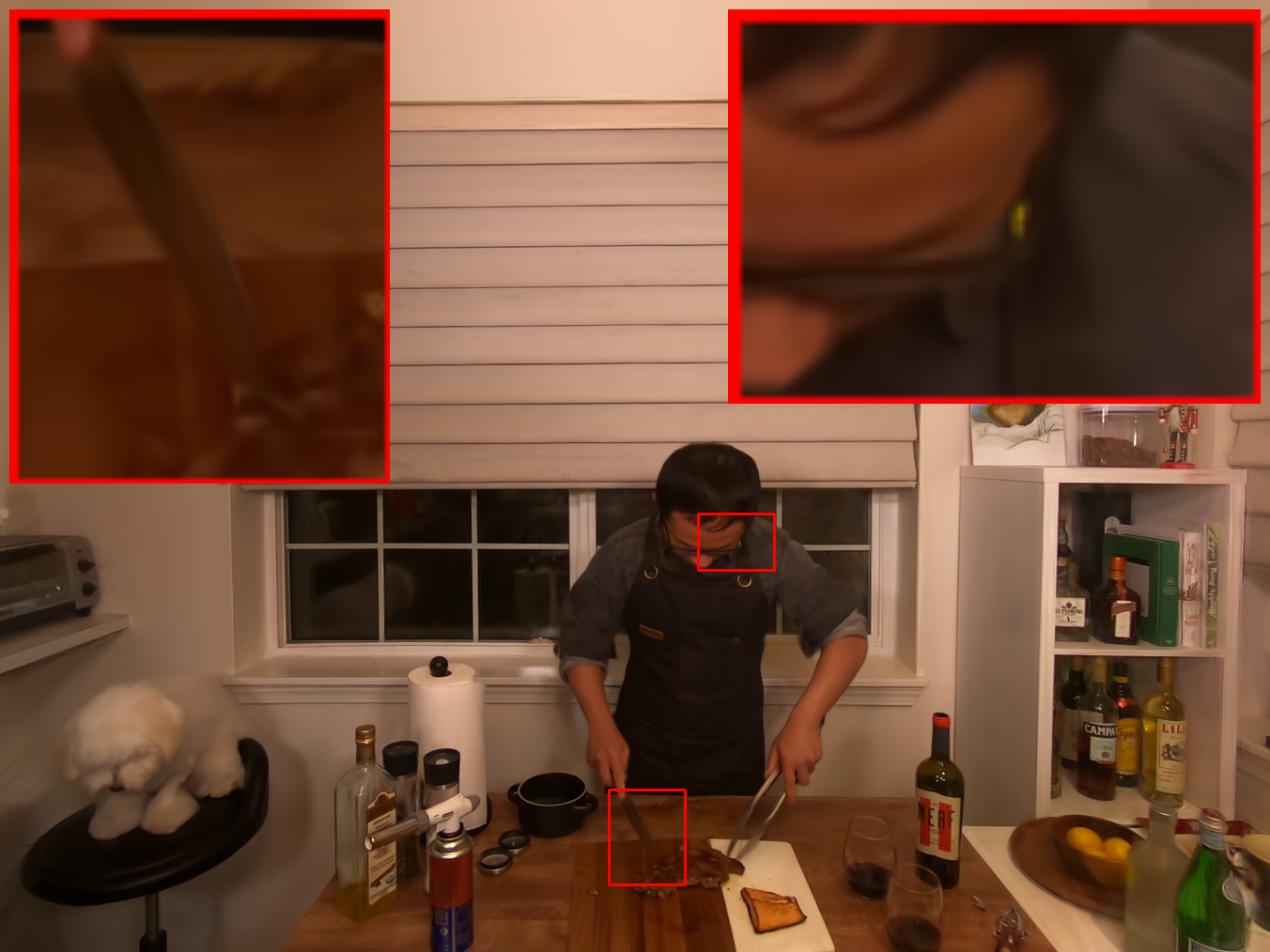}
        \caption*{(d) Ours}
    \end{subfigure}
    \hfill
    \begin{subfigure}[b]{0.19\linewidth}
        \includegraphics[width=\linewidth]{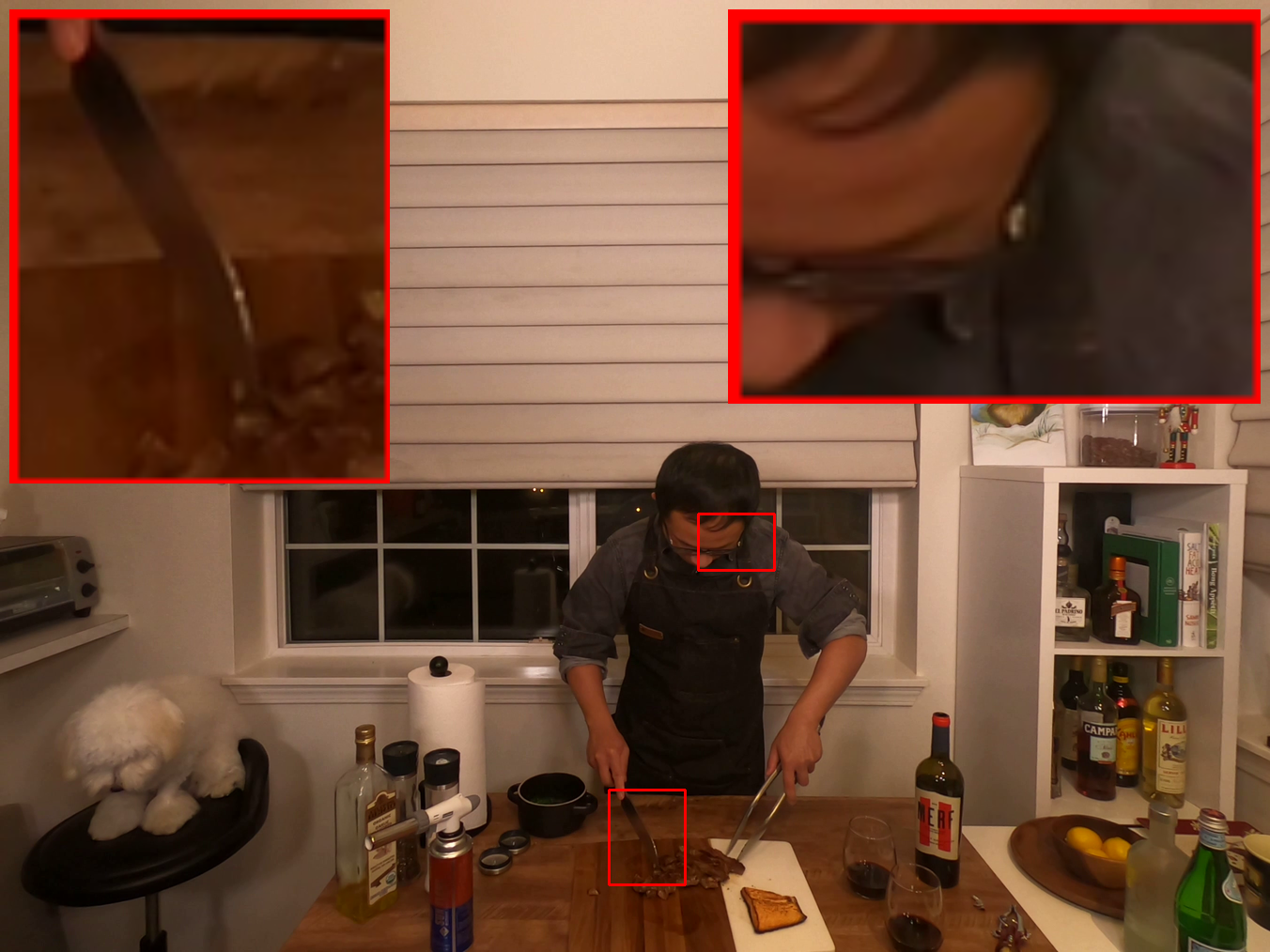}
        \caption*{(e) Ground Truth} % Corrected label from (d) to (e)
    \end{subfigure}
    
    \caption{\textbf{Qualitative comparisons} on the N3DV dataset.}
    \label{fig:n3dv_visual_Comparisons}
\end{figure*}
% ========== END OF FINAL VERSION ==========
\par
\subsection{Datasets and Comparisons} \label{dataset and comparison}
We selected three datasets to validate the effectiveness of our method: the N3DV~\cite{DyNeRF}, HyperNeRF~\cite{Hypernerf}, and Technicolor~\cite{technicolar} datasets.

\par
\textbf{N3DV} consists of scenes captured by 20 fixed cameras, with each video containing 300 frames (except for \textit{flame\_salmon}, which has 1200 frames). This dataset features various motions of multiple objects and long video sequences, and we conducted experimental comparisons on all its scenes. We divide the \textit{flame\_salmon} into 4 equal segments for training. 
In Tab.~\ref{tab:N3DV_Comparisons_Avg}, we present the quantitative results on the N3DV dataset. Specifically, NeRF-based methods generally suffer from long training times and slow rendering speeds. The 4DGS-based method, 4DGS, suffers from substantial training and storage overhead. Among deformation field-based methods, 4DGaussians and Swift4D offer fast training and rendering, but with subpar rendering quality. E-D3DGS delivers excellent rendering quality but suffers from slower rendering speeds and sensitivity to scene lighting, resulting in lower PSNR values. In contrast, our method achieves an exceptional balance between rendering quality, storage, and speed, delivering fast rendering with high quality.
\par
In Fig.~\ref{fig:n3dv_visual_Comparisons}, we present a comparison of rendering quality. Our method demonstrates excellent performance in both dynamic and static regions. Taking the \textit{flame\_salmon} as an example, compared to other methods, our approach effectively reconstructs the outdoor scene, including the background behind the person and the window on the right, while achieving high-quality modeling of the intense motion of the human body. All baselines perform poorly in modeling the outdoor scene, with sparse trees and blurred details of the house. In dynamic regions, 4DGaussians and Swift4D result in completely blurred human figures, while E-D3DGS exhibits significant blurring in the hand and the bottle area. Additionally, in the \textit{cut\_roasted\_beef} scene, our method achieves a better reconstruction of the glasses and the knife compared to all other methods.

% --- HyperNeRF Section ---
\begin{figure*}[!htb]
    \centering

    % --- ROW 1: vrig-peel-banana scene ---
    % Using \hfill to automatically distribute horizontal space
    \begin{subfigure}[b]{0.19\linewidth}
      \includegraphics[width=\linewidth]{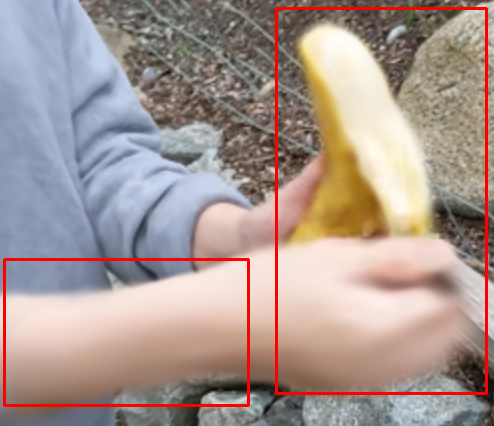}
    \end{subfigure}
    \hfill 
    \begin{subfigure}[b]{0.19\linewidth}
        \includegraphics[width=\linewidth]{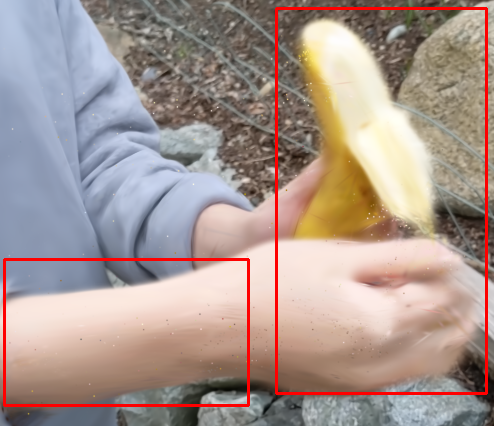}
    \end{subfigure}
    \hfill
    \begin{subfigure}[b]{0.19\linewidth}
        \includegraphics[width=\linewidth]{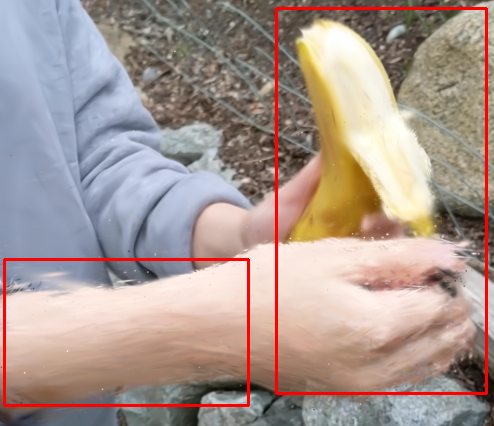}
    \end{subfigure}
    \hfill
    \begin{subfigure}[b]{0.19\linewidth}
        \includegraphics[width=\linewidth]{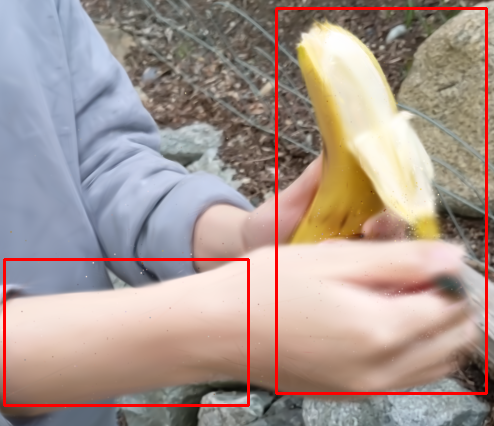}
    \end{subfigure}    
    \hfill
    \begin{subfigure}[b]{0.19\linewidth}
        \includegraphics[width=\linewidth]{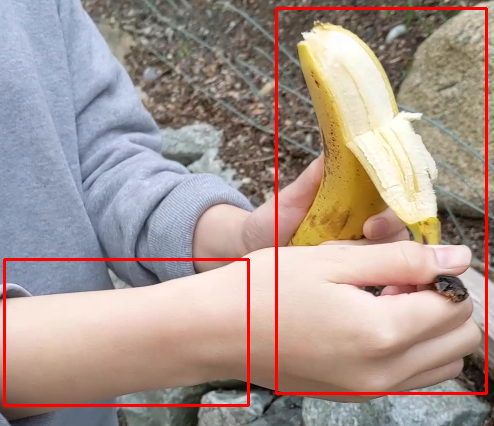} % Corrected from Ours.png to gt.png
    \end{subfigure}

    \vspace{2mm} % Vertical space between rows

    % --- ROW 2: vrig-broom scene ---
    \begin{subfigure}[b]{0.19\linewidth}
      \includegraphics[width=\linewidth]{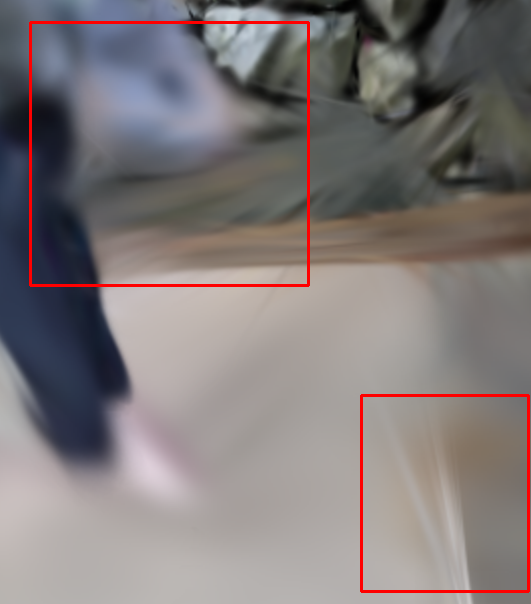}
      \caption*{(a) D3DGS~\cite{deformable3dgs}}
    \end{subfigure}
    \hfill
    \begin{subfigure}[b]{0.19\linewidth}
        \includegraphics[width=\linewidth]{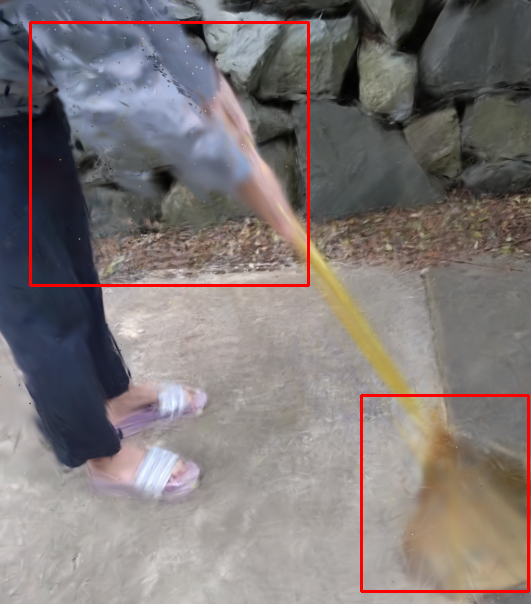}
        \caption*{(b) 4DGaussians~\cite{4dgs_kplanes}}
    \end{subfigure}
    \hfill
    \begin{subfigure}[b]{0.19\linewidth}
        \includegraphics[width=\linewidth]{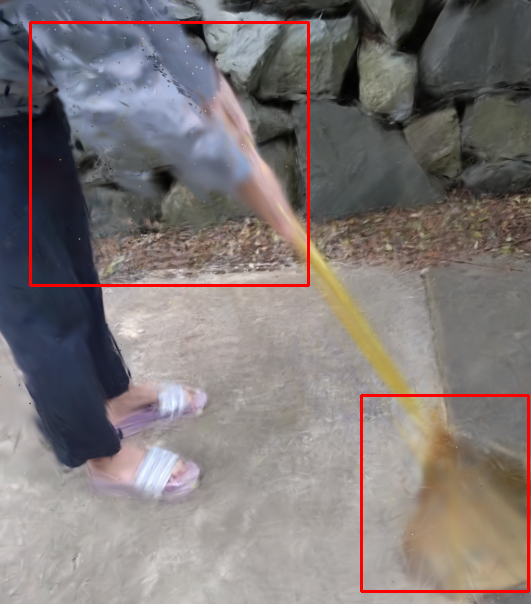}
        \caption*{(c) E-D3DGS~\cite{ed3dgs}}
    \end{subfigure}
    \hfill
    \begin{subfigure}[b]{0.19\linewidth}
        \includegraphics[width=\linewidth]{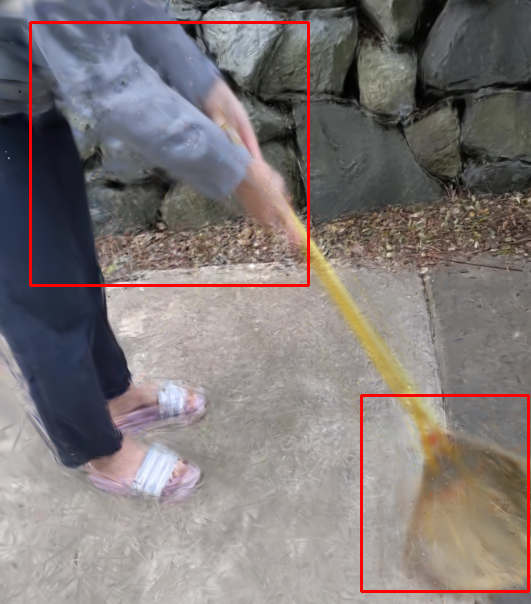}
        \caption*{(d) Ours}
    \end{subfigure}    
    \hfill
    \begin{subfigure}[b]{0.19\linewidth}
        \includegraphics[width=\linewidth]{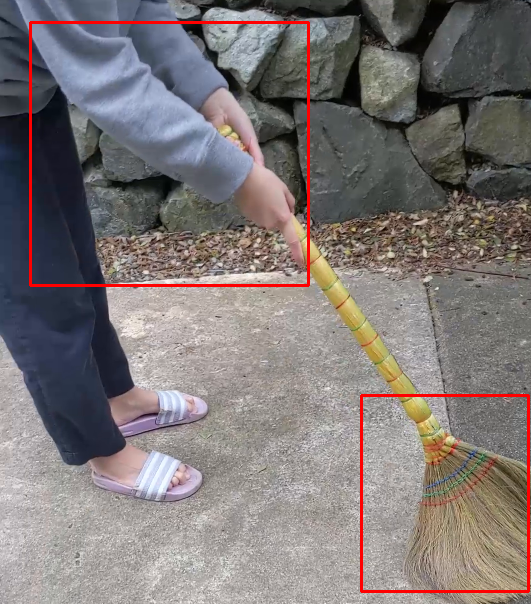}
        \caption*{(e) GT}
    \end{subfigure}
    
    \caption{\textbf{Qualitative comparisons} on the HyperNeRF dataset.}
    \label{fig:hypernerf_visual_Comparisons}
\end{figure*}
\begin{table*}[t]
  \centering
  \caption{
    \textbf{Quantitative comparison on the HyperNeRF dataset.}
    Storage, training time, and FPS are tested across all four scenes, based on the NVIDIA RTX A6000. 
    $^1$Storage, training time, and FPS are tested only on \textit{Broom}.
    $^2$Storage, training time, and FPS are tested using an NVIDIA RTX 3090. For a fair comparison, we also report results using the specific settings of other methods.
  }
  \label{tab:HyperNeRF_Comparisons_Avg}

  % --- KEY CHANGE: Use tabular* to span the full text width ---
  % We remove \setlength{\tabcolsep} and let @{\extracolsep{\fill}} handle the spacing automatically.
  \begin{tabular*}{\textwidth}{@{\extracolsep{\fill}}lcccccc} 
    \toprule
    Method & PSNR$\uparrow$ & SSIM$\uparrow$ & LPIPS$\downarrow$ & Storage$\downarrow$ & Training Time$\downarrow$ & FPS$\uparrow$ \\
    \midrule
    Nerfies$^{1,2}$~\cite{Nerfies}             & 22.23           & -       & 0.170 & -          & $\sim$ hours       & $<$ 1                     \\
    HyperNeRF$^{1,2}$~\cite{Hypernerf}           & 22.29           & 0.598   & 0.153    & 15 MB & 32h 45m            & $<$ 1                     \\
    TiNeuVox$^{1,2}$~\cite{TiNeuVox}            & 24.20           & 0.616   & 0.393    & 48 MB      & 30m                & 1                       \\
    Ours$^{1,2}$                & 26.31           & 0.721   & 0.199    & 37 MB      & 57m                & 146.50                  \\
    \midrule
    DN-4DGS$^{2}$~\cite{dn4dgs}               & 25.59 & 0.691 & -          & 68 MB      & 1h 12m             & 20                      \\
    Ours$^{2}$                  & 26.31           & 0.721   & 0.199    & 51 MB      & 1h 28m             & 110.45                  \\
    \midrule
    D3DGS~\cite{deformable3dgs}                       & 22.40           & 0.598   & 0.275    & 195 MB     & 1h 54m             & 11.54                   \\
    4DGaussians~\cite{4dgs_kplanes}                 & 25.10           & 0.688   & 0.271    & 71 MB      & 20m & 100.41                  \\
    E-D3DGS~\cite{ed3dgs}                     & 25.92 & 0.716 & 0.197 & 53 MB      & 1h 50m             & 78.45                   \\
    Ours                        & 26.31    & 0.721 & 0.187 & 51 MB & 1h 21m & 125.91 \\
    \bottomrule
  \end{tabular*}
\end{table*}
% ========== END OF FINAL TABLE ==========
\par \textbf{HyperNeRF} includes videos captured by two fixed smartphones, and we compared various methods on all frames of four scenes (3D Printer, Banana, Broom, Chicken). In Tab.~\ref{tab:HyperNeRF_Comparisons_Avg}, our method demonstrates exceptional performance on the HyperNeRF dataset, with both rendering quality and speed outperforming the baselines. From the qualitative results in Fig.~\ref{fig:hypernerf_visual_Comparisons}, it can be observed that D3DGS fails to achieve effective reconstruction in \textit{broom}. In \textit{peel-banana}, 4DGaussians exhibits numerous noisy 3DGs in the hand region, while E-D3DGS shows severe blurring and fragmentation. Additionally, in \textit{broom}, both 4DGaussians and E-D3DGS exhibit noticeable fragmentation in the reconstruction of clothing, and the details of the broom are relatively blurred. In contrast, our method demonstrates higher robustness and detail reconstruction capabilities in these challenging scenarios.

\begin{table*}[t]
  \centering
  \caption{
    \textbf{Quantitative comparison on the Technicolor dataset.}
    Storage, training time, and FPS are tested on all five scenes and are calculated on an A6000. For storage, training time, and FPS, $^1$ are tested only on \textit{Painter}, and $^2$ are measured using a 3090.
  }
  \label{tab:Technicolar_Comparisons_Avg}

  % --- KEY CHANGE: Use tabular* to span the full text width ---
  % We remove \setlength{\tabcolsep} and let @{\extracolsep{\fill}} handle the spacing automatically.
  \begin{tabular*}{\textwidth}{@{\extracolsep{\fill}}lcccccc} 
    \toprule
    Method & PSNR$\uparrow$ & SSIM$\uparrow$ & LPIPS$\downarrow$ & Storage$\downarrow$ & Training Time$\downarrow$ & FPS$\uparrow$ \\
    \midrule
    DyNeRF$^{1,2}$~\cite{DyNeRF}              & 31.80           & -       & 0.140    & 0.6MB   & -                  & 0.02                    \\
    HyperReel$^{1,2}$~\cite{Hyperreel}           & 32.32 & 0.899 & 0.118 & 289 MB     & 2h 45m             & 0.91                    \\
    Ours$^{1,2}$                & 33.28           & 0.907   & 0.105    & 43 MB      & 1h 51m             & 109.10                  \\
    \midrule
    4DGS~\cite{4dgs_towards}                        & 29.54           & 0.873   & 0.109 & -          & -                  & -                       \\
    4DGaussians~\cite{4dgs_kplanes}                 & 29.62           & 0.844   & 0.176    & 85 MB      & 27m & 49.81                   \\
    E-D3DGS~\cite{ed3dgs}                     & 33.08 & 0.902 & 0.110 & 55 MB      & 2h 35m             & 71.72                   \\
    Ours                        & 33.28    & 0.907 & 0.105 & 58 MB      & 2h 17m             & 110.15 \\
    \bottomrule
  \end{tabular*}
\end{table*}
% ========== END OF FINAL TABLE ========== 
\begin{figure*}[!htb]
    \centering

    % 第二行：Painter
    \begin{subfigure}{0.24\textwidth}
      \includegraphics[width=\linewidth]{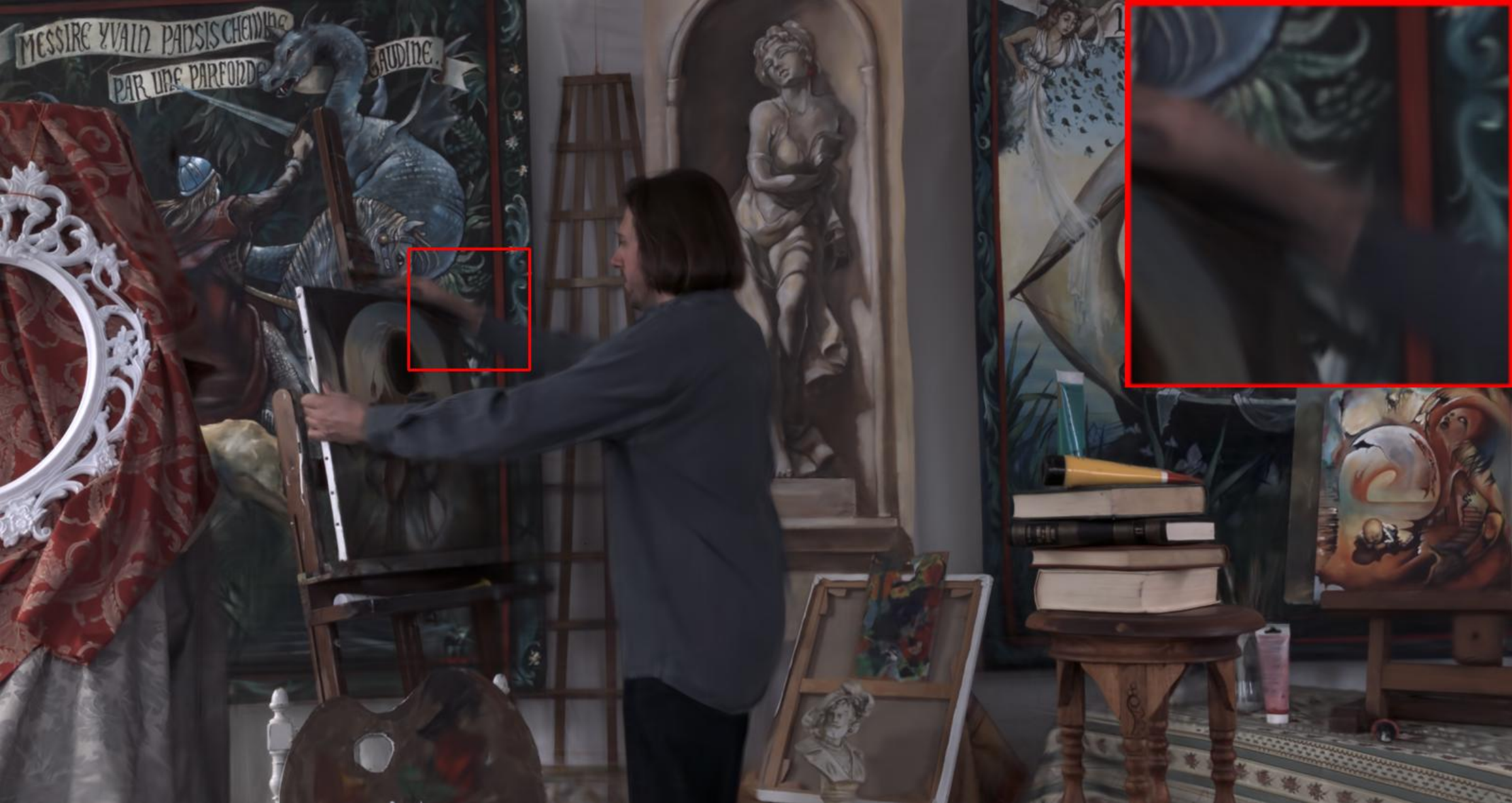}
    \end{subfigure}
    \begin{subfigure}{0.24\textwidth}
        \includegraphics[width=\linewidth]{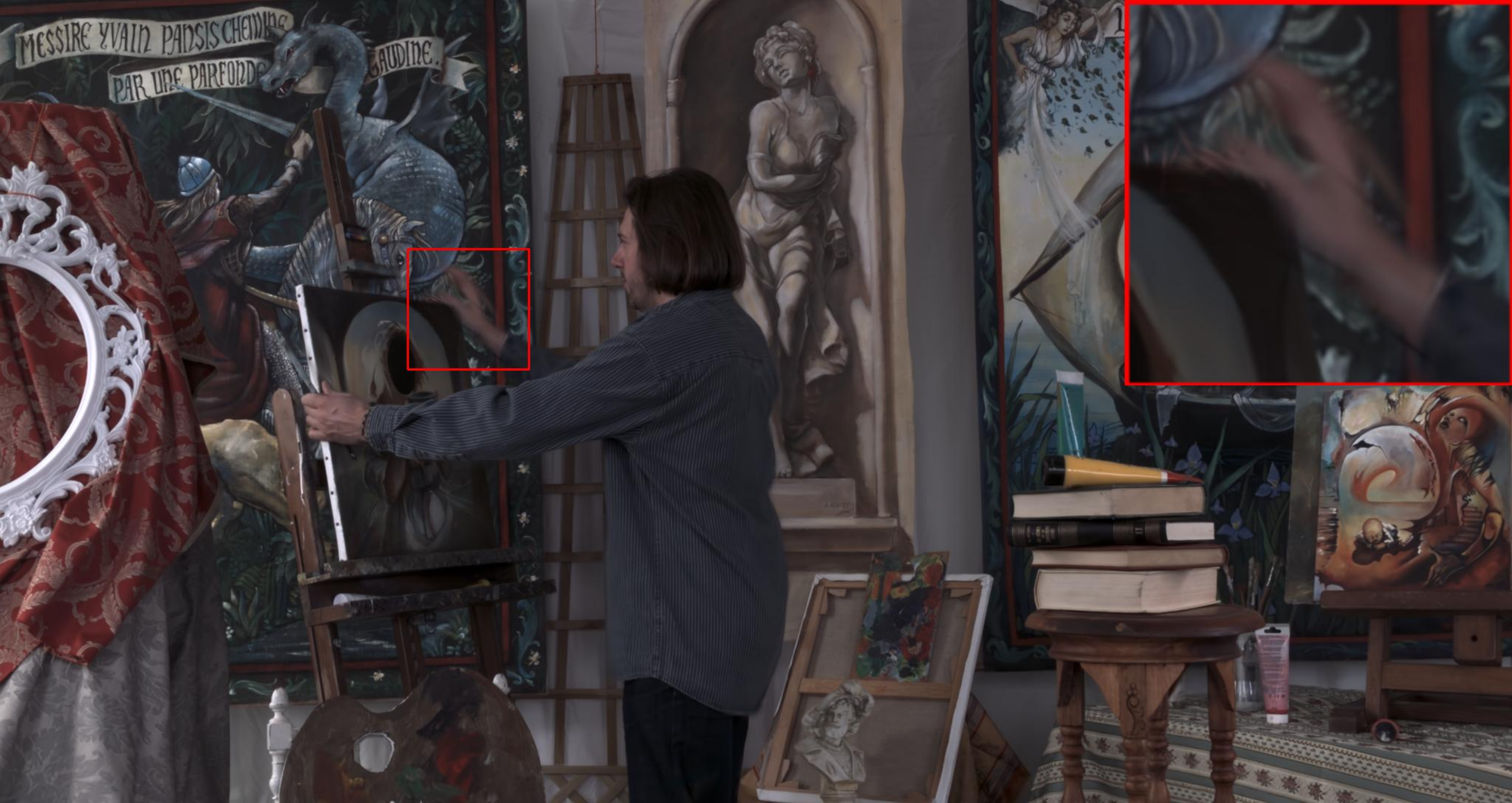}
    \end{subfigure}
    \begin{subfigure}{0.24\textwidth}
        \includegraphics[width=\linewidth]{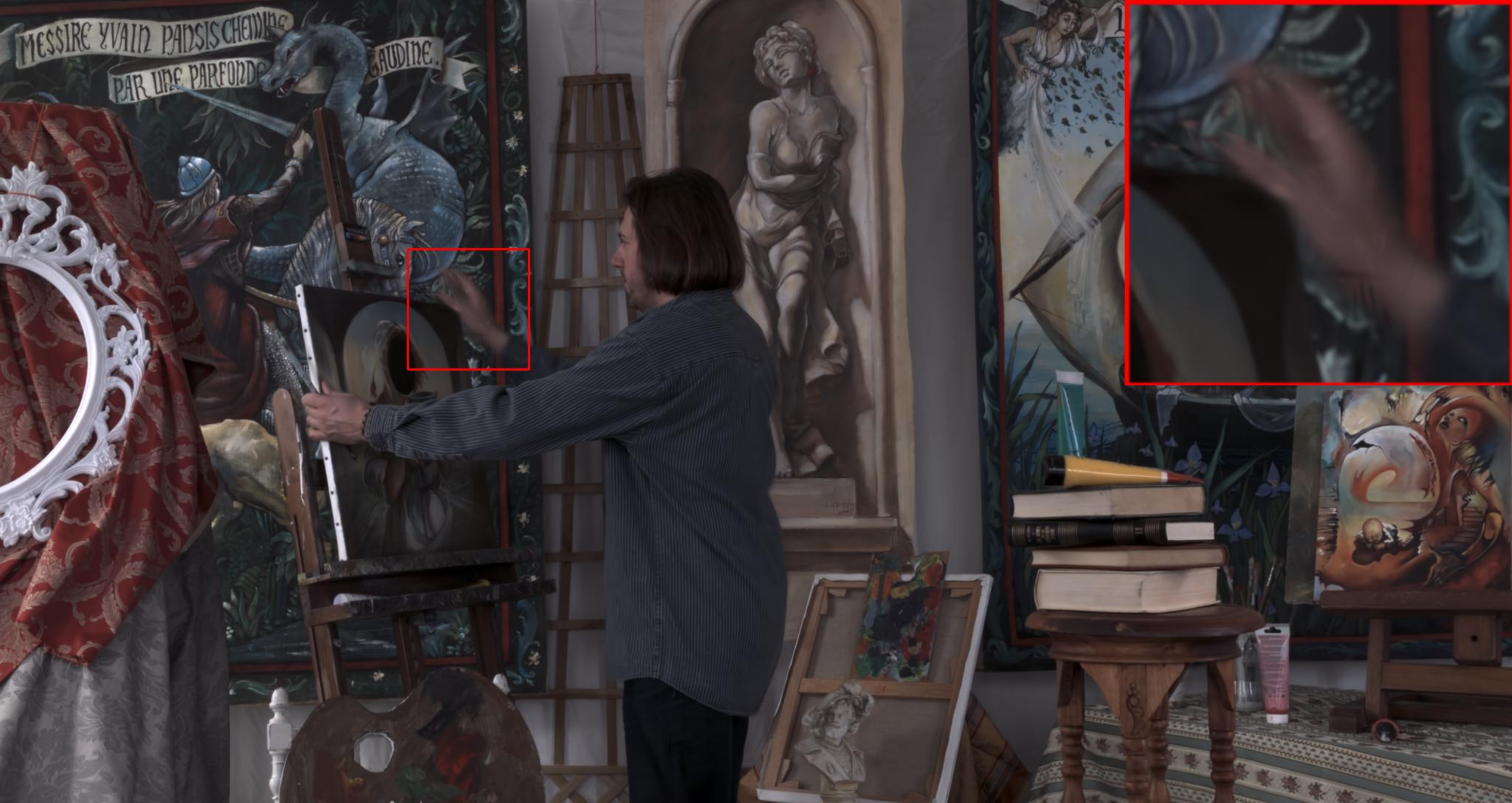}
    \end{subfigure}
    \begin{subfigure}{0.24\textwidth}
        \includegraphics[width=\linewidth]{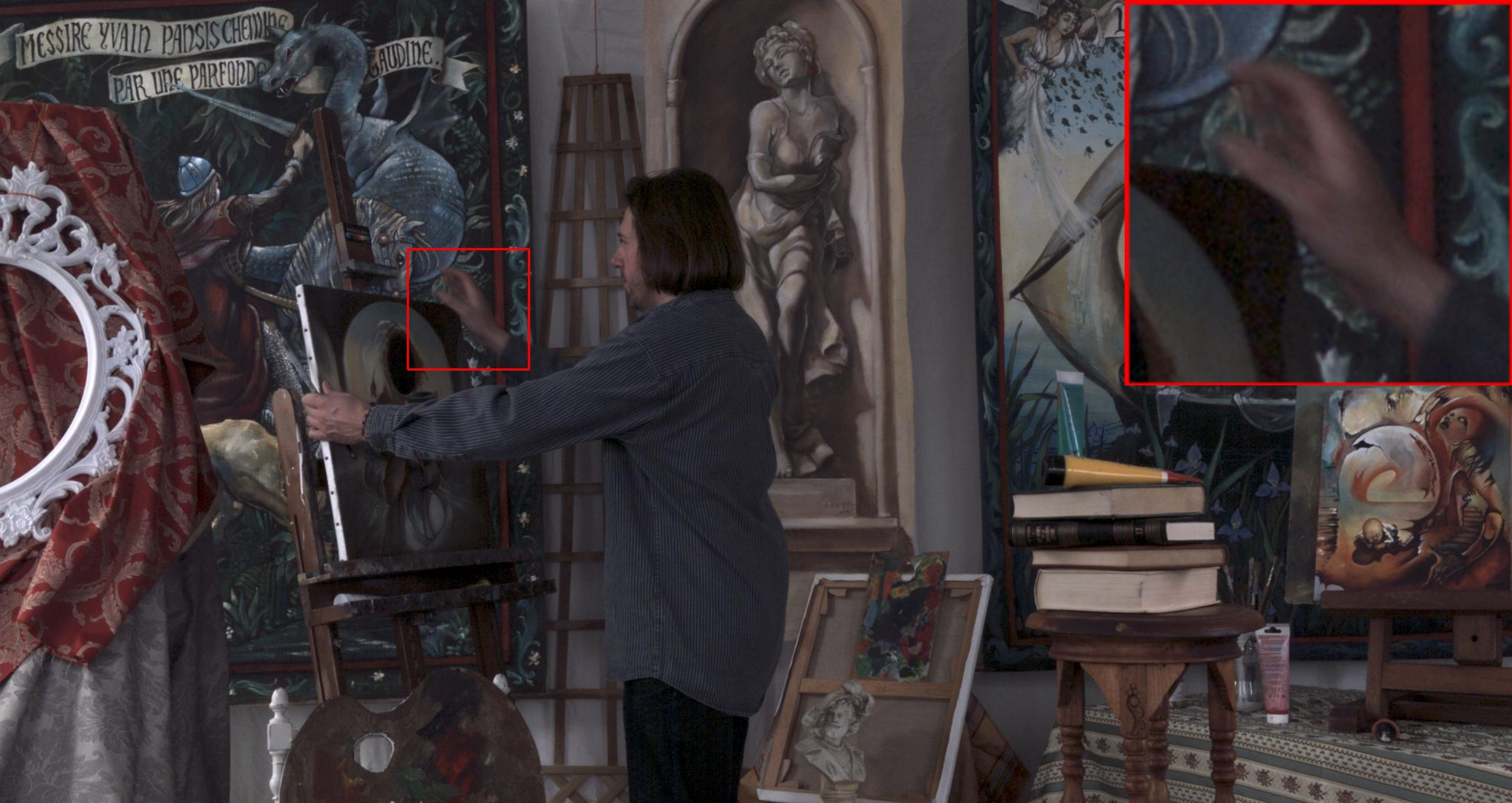}
    \end{subfigure}

    \vspace{1mm}

    % 第四行：Train
    \begin{subfigure}{0.24\textwidth}
      \includegraphics[width=\linewidth]{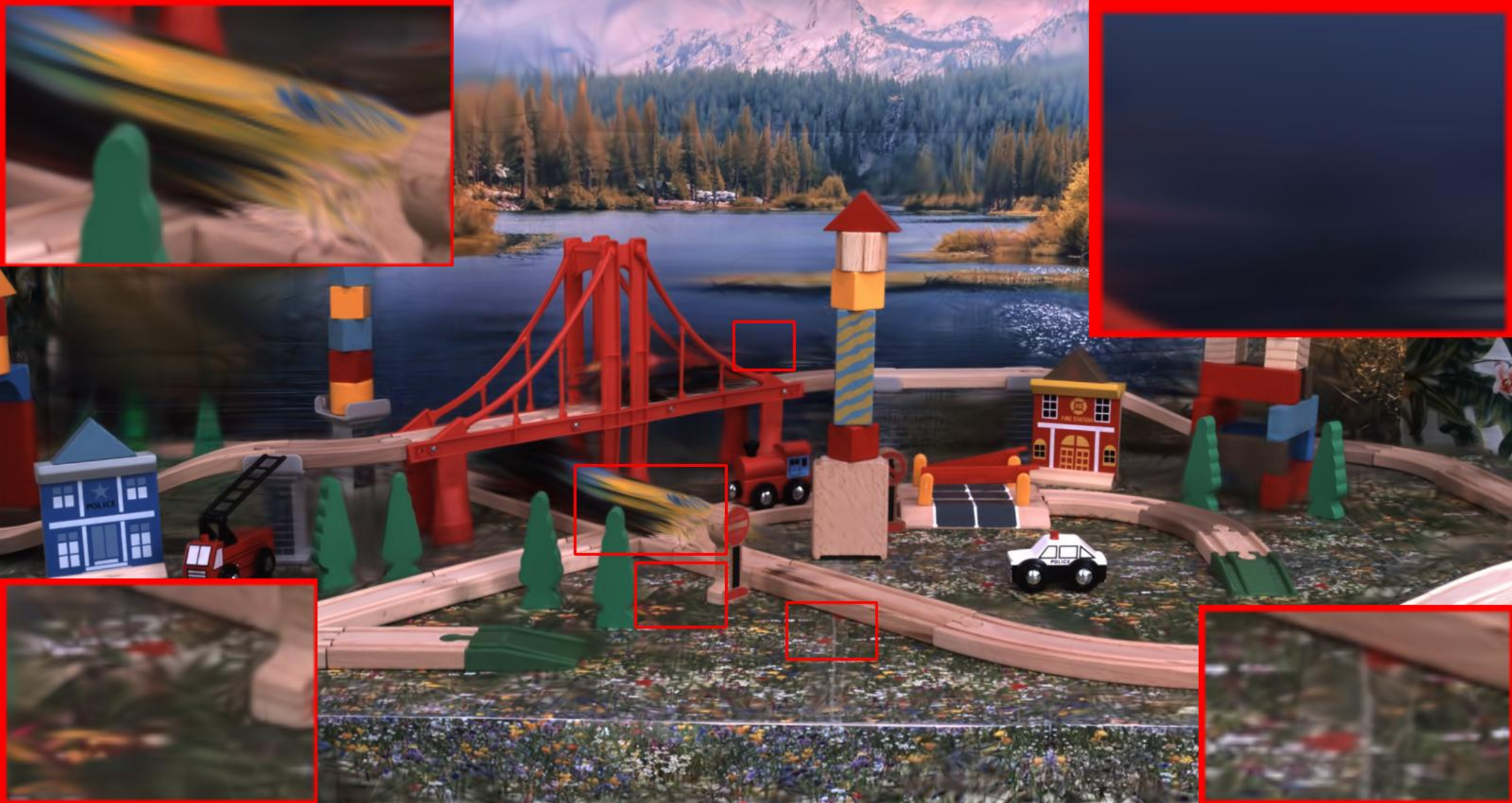}
      \caption*{(a) 4DGaussians~\cite{4dgs_kplanes}}
    \end{subfigure}
    \begin{subfigure}{0.24\textwidth}
        \includegraphics[width=\linewidth]{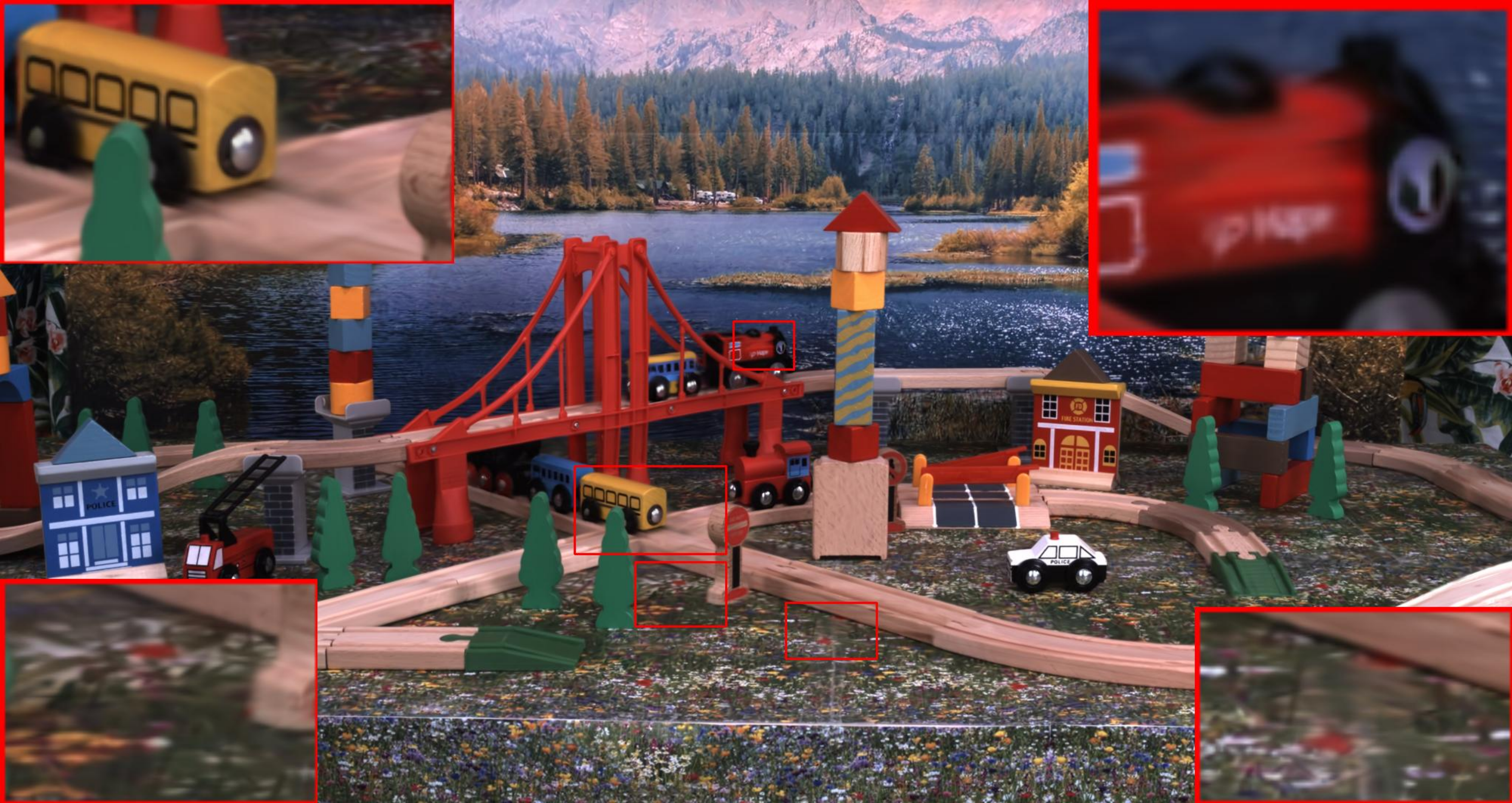}
        \caption*{(b) E-D3DGS~\cite{ed3dgs}}
    \end{subfigure}
    \begin{subfigure}{0.24\textwidth}
        \includegraphics[width=\linewidth]{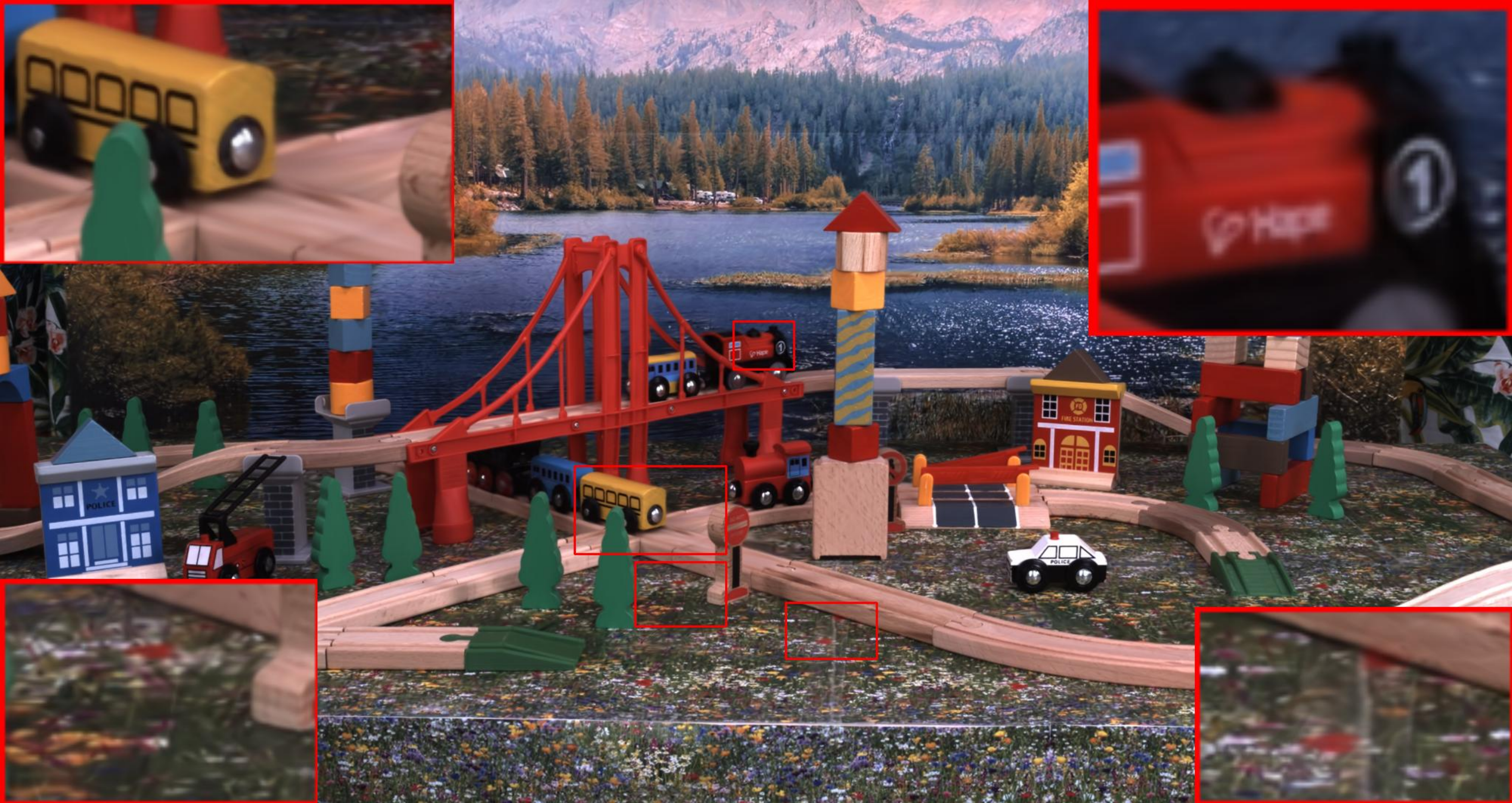}
        \caption*{(c) Ours}
    \end{subfigure}
    \begin{subfigure}{0.24\textwidth}
        \includegraphics[width=\linewidth]{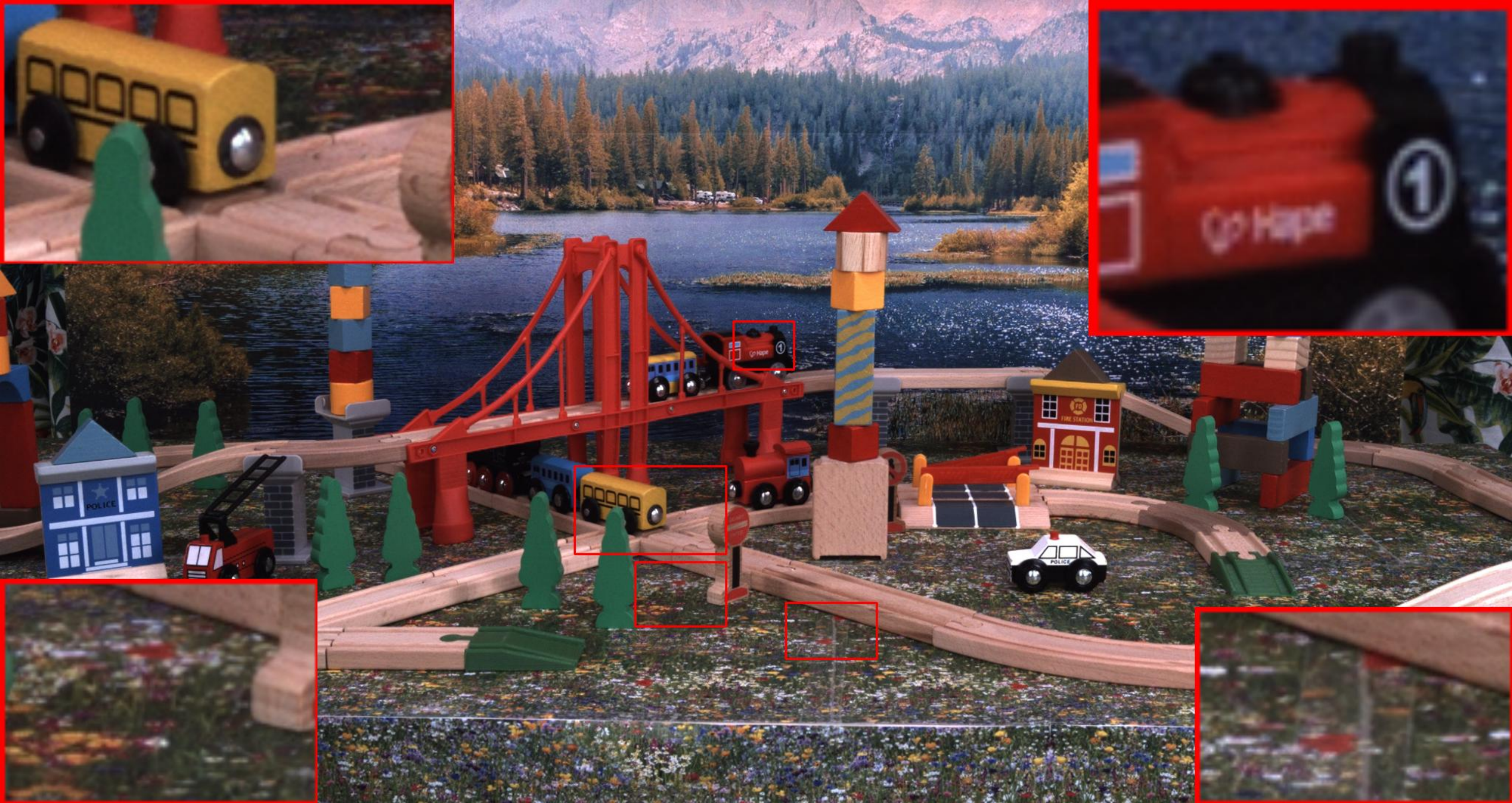}
        \caption*{(d) Ground Truth}
    \end{subfigure}
    \caption{\textbf{Qualitative comparisons} on the Technicolor dataset.}
    \label{fig:technicolar_visual_Comparisons}
    
  % \vspace{-2mm}
\end{figure*}
\par \textbf{Technicolor} is captured by 16 fixed cameras, with the camera in the second row and the second column selected as the test set and the others as the training set. This dataset features relatively simple motions but complex details. We compared various methods on 50 frames of five commonly used scenes (Birthday, Fabien, Painter, Theater, Train). 
The quantitative results are summarized in Tab.~\ref{tab:Technicolar_Comparisons_Avg}. Our method demonstrates exceptional performance, surpassing other deformation field-based methods while delivering fast rendering speed.
Qualitative comparisons are shown in Fig.~\ref{fig:technicolar_visual_Comparisons}. 4DGaussians exhibits severe blurring in the dynamic areas. In \textit{Painter}, E-D3DGS exhibits significant drift in 3DGs due to the intense motion of the painter’s hand. Additionally, E-D3DGS exhibits noticeable blurring around the moving red train and fails to reconstruct the details of the tracks and grass effectively in \textit{Train}. In contrast, our method consistently demonstrates superior motion handling and detail reconstruction capabilities in these challenging scenarios.

\subsection{Evaluation}\label{sec: evaluation}
In this section, we conduct a series of ablation studies to validate the effectiveness of the key components of our proposed method. We analyze the embedding fusion strategy, the canonical field sampling strategy, and the aggressive opacity reduction strategy.

\begin{table*}[t!]
  \centering
  \caption{\textbf{Comprehensive ablation study on the N3DV dataset.} The top section compares various temporal embedding fusion strategies, while the bottom section demonstrates the incremental performance gains provided by the Canonical Field Sampling and Aggressive Opacity Reduction strategies.}
  \label{tab:feature_fusion_comparison}
  \begin{tabular*}{\textwidth}{@{\extracolsep{\fill}}lcccccc}
    \toprule
    Method & PSNR$\uparrow$ & SSIM$\uparrow$ & LPIPS$\downarrow$ & Storage$\downarrow$ & Training Time$\downarrow$ & FPS$\uparrow$ \\
    \midrule
    Coarse & 30.44 & 0.930 & 0.059 & 57 MB & 1h 51m & 102.95 \\
    Fine & 30.38 & 0.927 & 0.062 & 59 MB & 1h 49m & 103.35 \\
    % \midrule
    Concat & 30.64 & 0.932 & 0.056 & 58 MB & 1h 56m & 92.03 \\
    Add & 30.55 & 0.930 & 0.057 & 59 MB & 1h 53m & 92.57 \\
    Attention & 30.74 & 0.935 & {0.053} & 61 MB & 2h 5m & 69.62 \\
    Dual & {30.79} & 0.937 & {0.051} & 60 MB & 2h 22m & 49.60 \\
    Product (Ours) & 30.76 & 0.937 & {0.053} & 58 MB & 1h 51m & 101.78 \\
    \midrule
    \textit{Incremental Components} & & & & & & \\
    \quad + Canonical Field Sampling Strategy & 30.86 & 0.939 & 0.052 & 60 MB & 1h 53m & 98.87 
    \\
    % \midrule
    \quad \quad + Aggressive Opacity Reduction Strategy & 31.01 & 0.941& 0.050 & 59 MB & 1h 51m & 100.62 
    \\
    
    \bottomrule
  \end{tabular*}
\end{table*}

\begin{figure*}[!htb]
    \centering
    
    % --- 第一行：abcd ---
    \begin{subfigure}{0.24\linewidth}
        \centering
        \includegraphics[width=\linewidth]{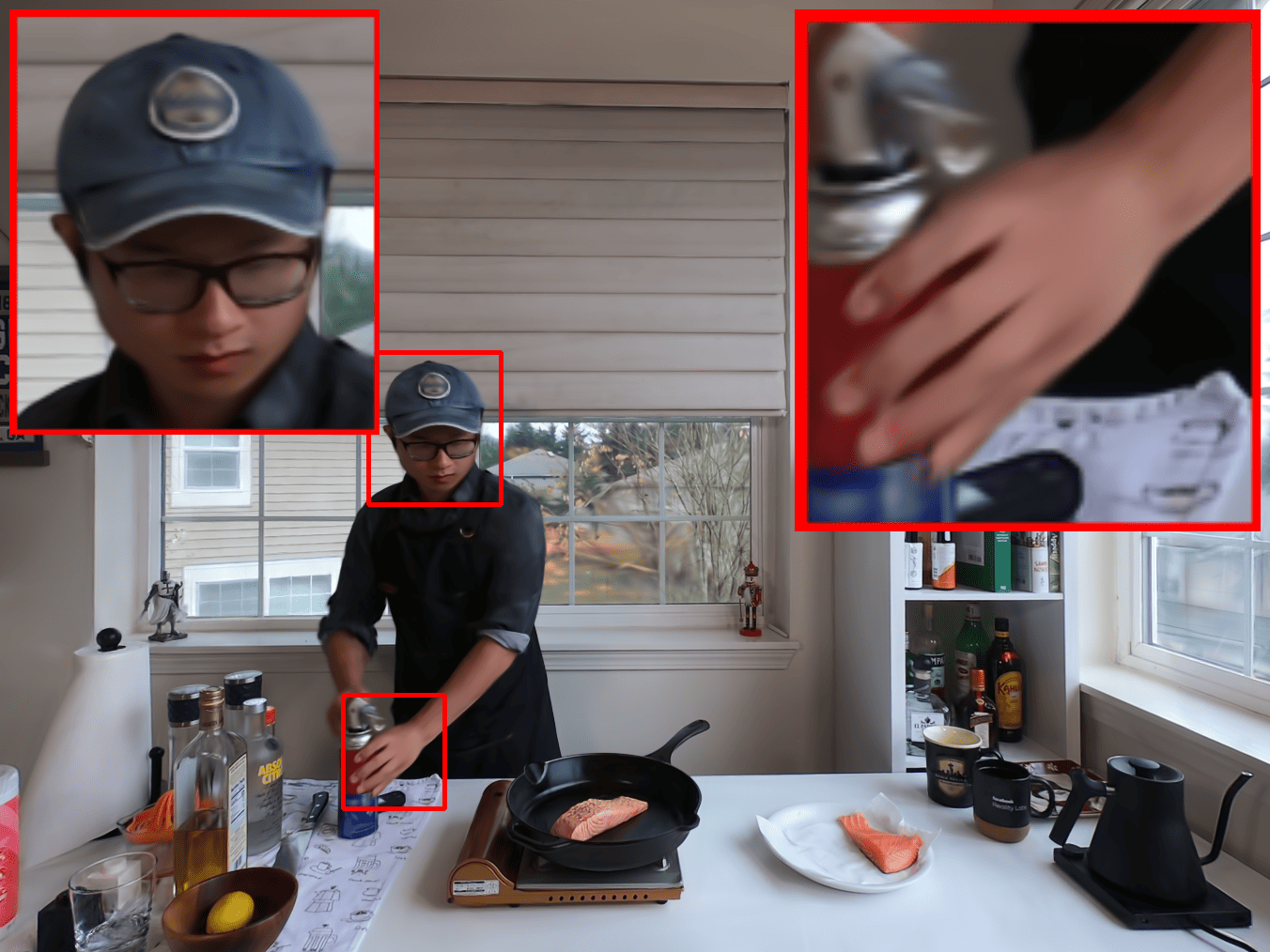}
        \caption*{(a) Coarse}
    \end{subfigure}
    \hfill
    \begin{subfigure}{0.24\linewidth}
        \centering
        \includegraphics[width=\linewidth]{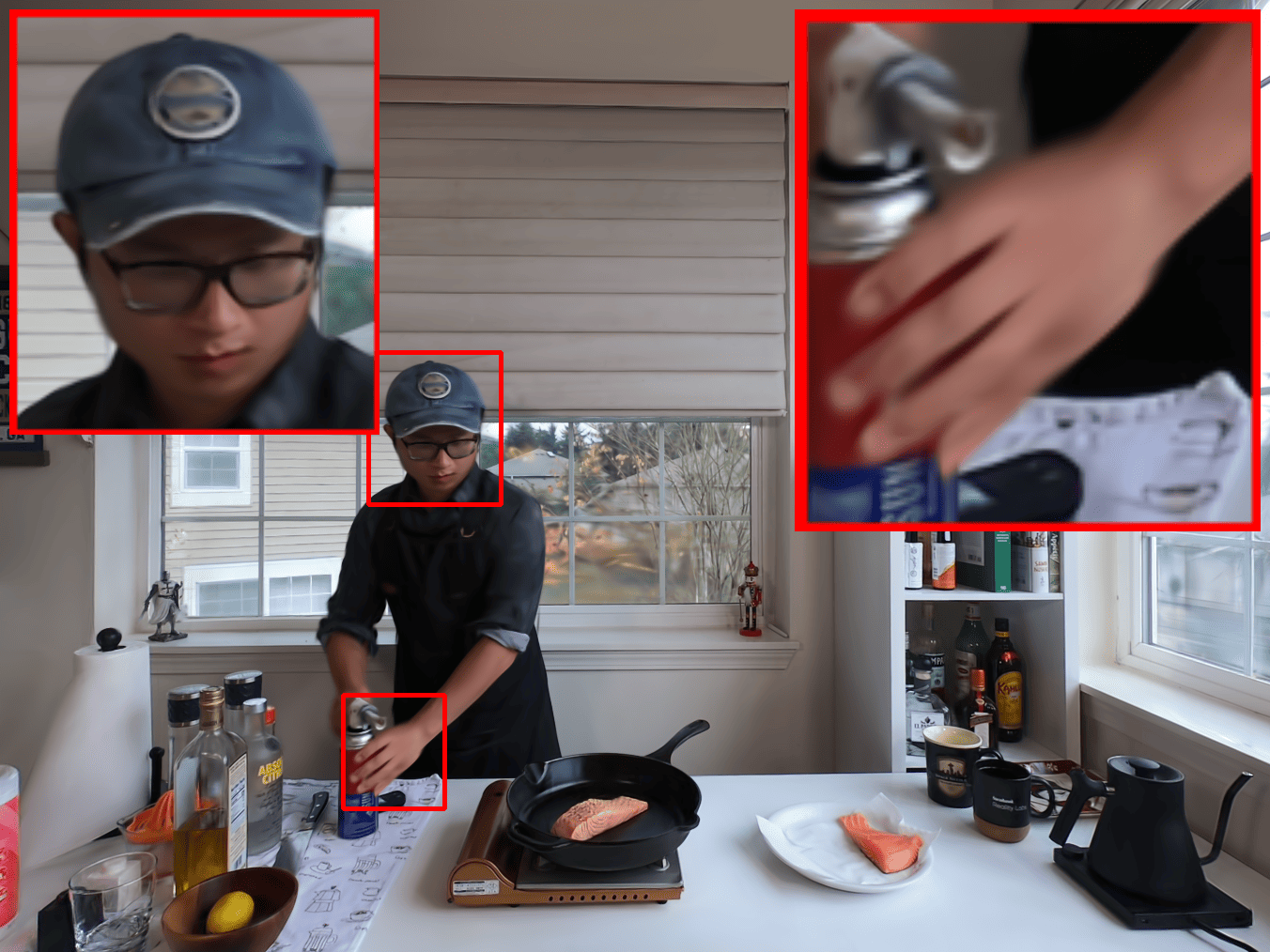}
        \caption*{(b) Fine}
    \end{subfigure}
    \hfill
    \begin{subfigure}{0.24\linewidth}
        \centering
        \includegraphics[width=\linewidth]{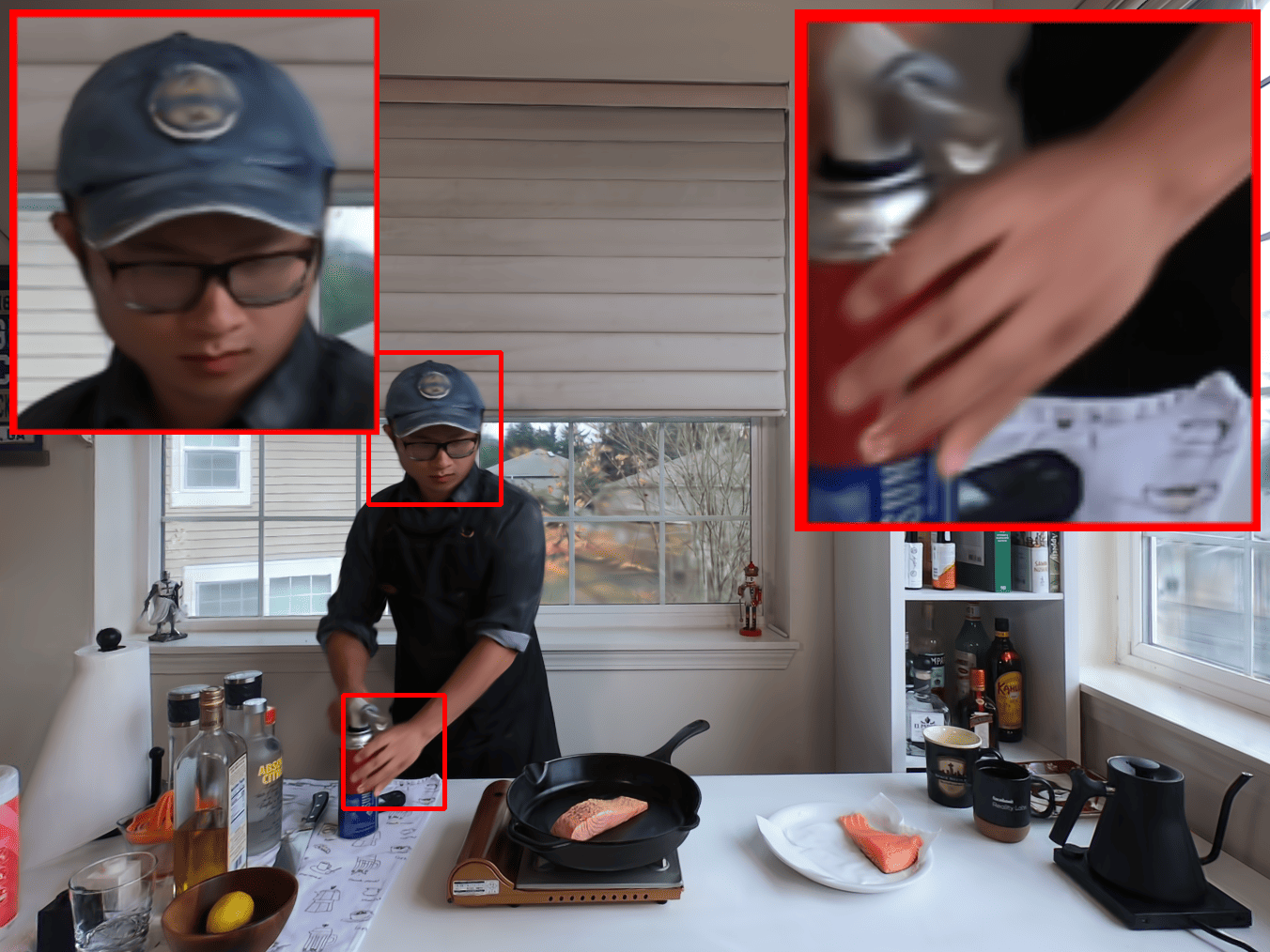}
        \caption*{(c) Add}
    \end{subfigure}
    \hfill
    \begin{subfigure}{0.24\linewidth}
        \centering
        \includegraphics[width=\linewidth]{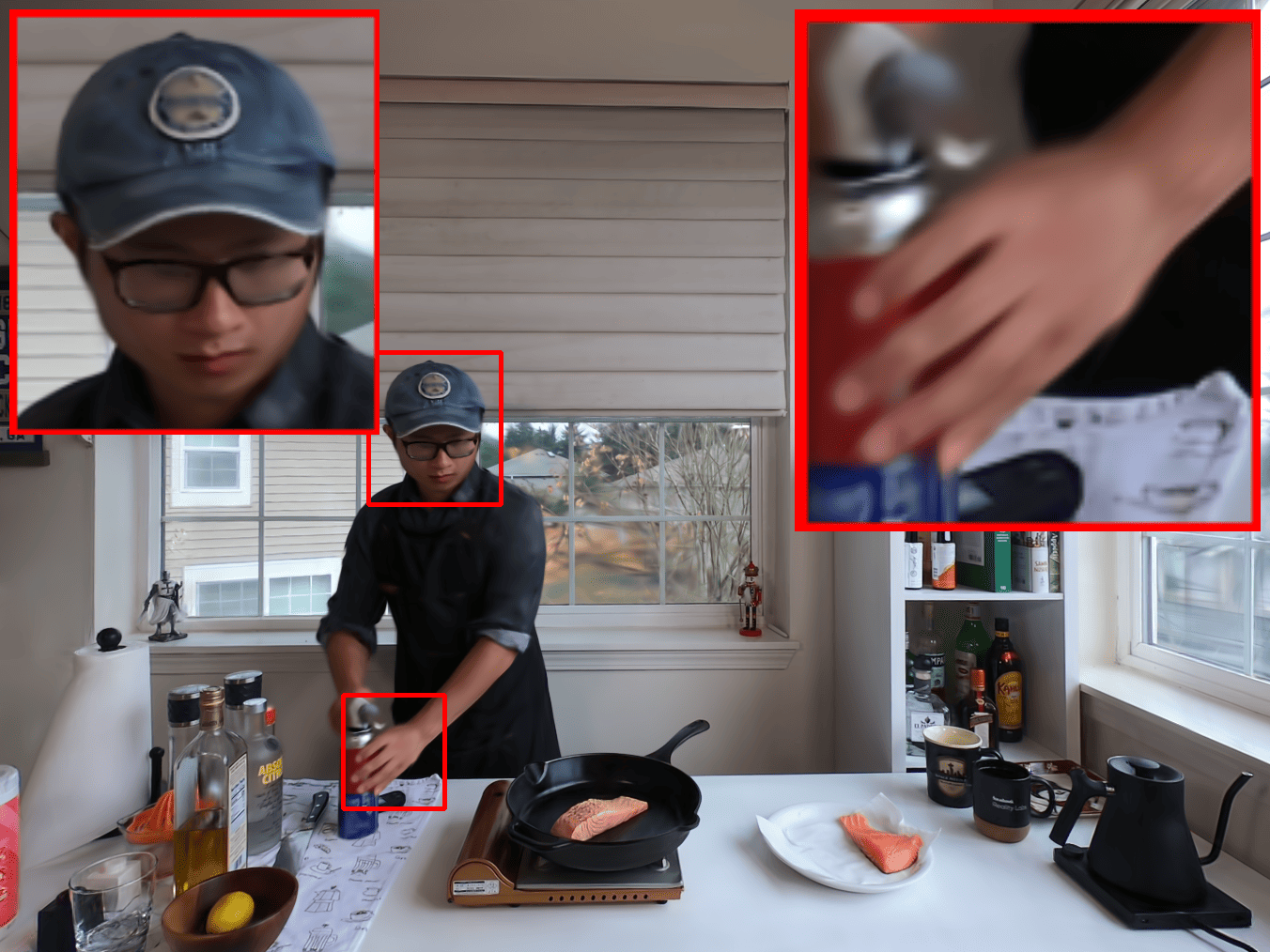}
        \caption*{(d) Concat}
    \end{subfigure}
    
    % --- 在两行之间增加垂直间距 ---
    \vspace{1mm} % 您可以调整这个值来改变间距大小，例如 3mm, 0.5cm 等

    % --- 第二行：efgh ---
    \begin{subfigure}{0.24\linewidth}
        \centering
        \includegraphics[width=\linewidth]{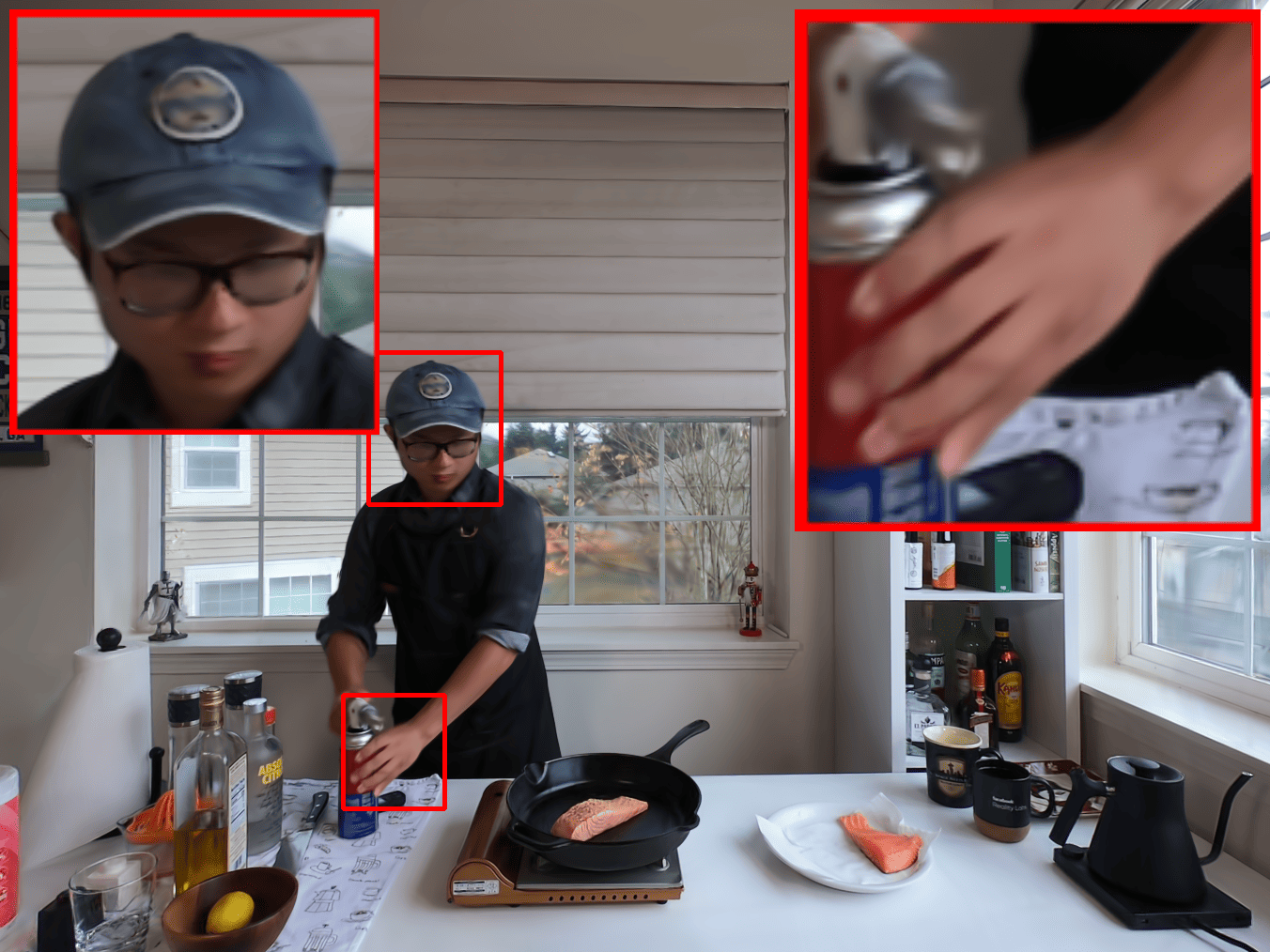}
        \caption*{(e) Attention}
    \end{subfigure}
    \hfill
    \begin{subfigure}{0.24\linewidth}
        \centering
        \includegraphics[width=\linewidth]{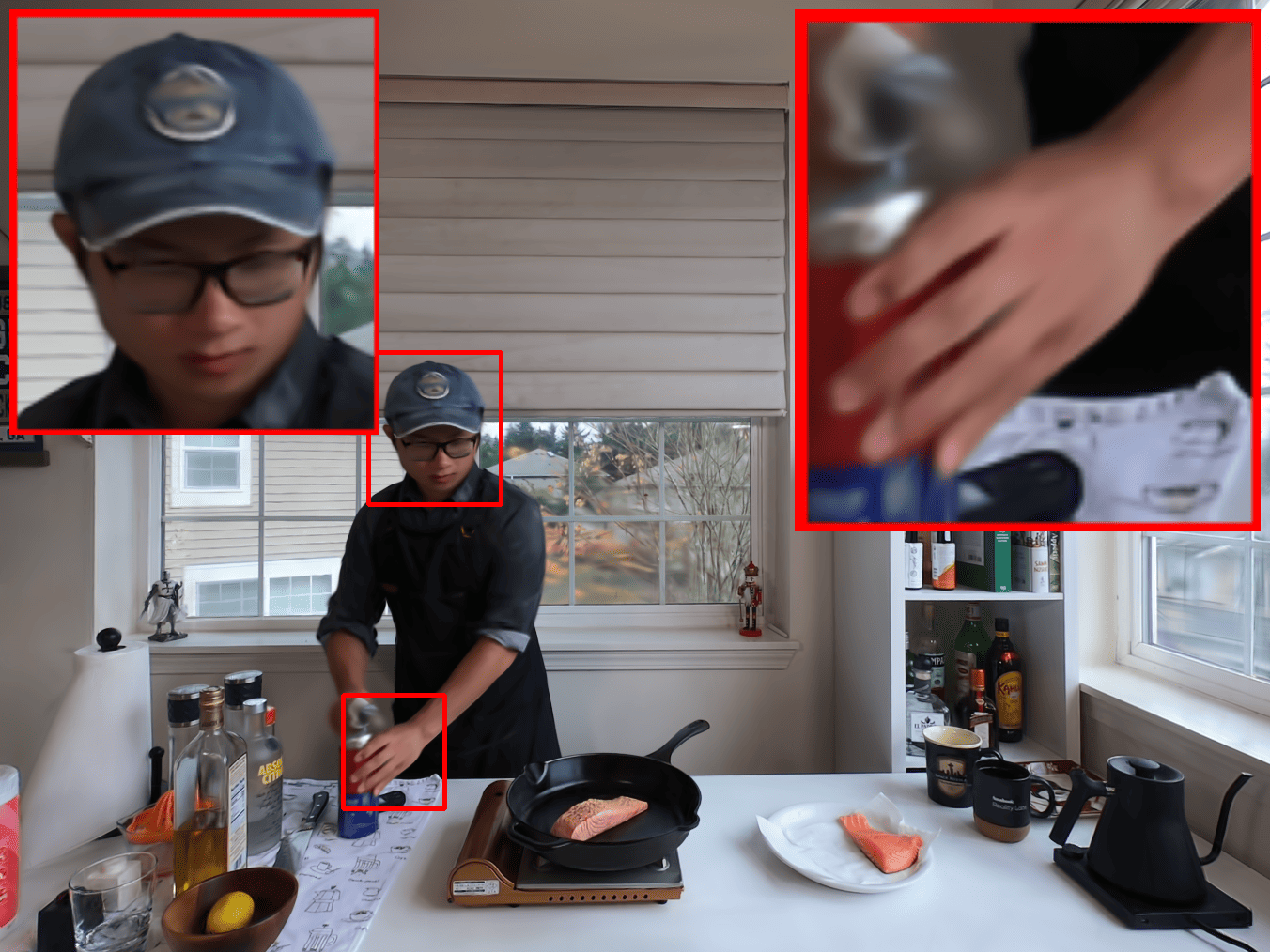}
        \caption*{(f) Dual}
    \end{subfigure}
    \hfill
    \begin{subfigure}{0.24\linewidth}
        \centering
        \includegraphics[width=\linewidth]{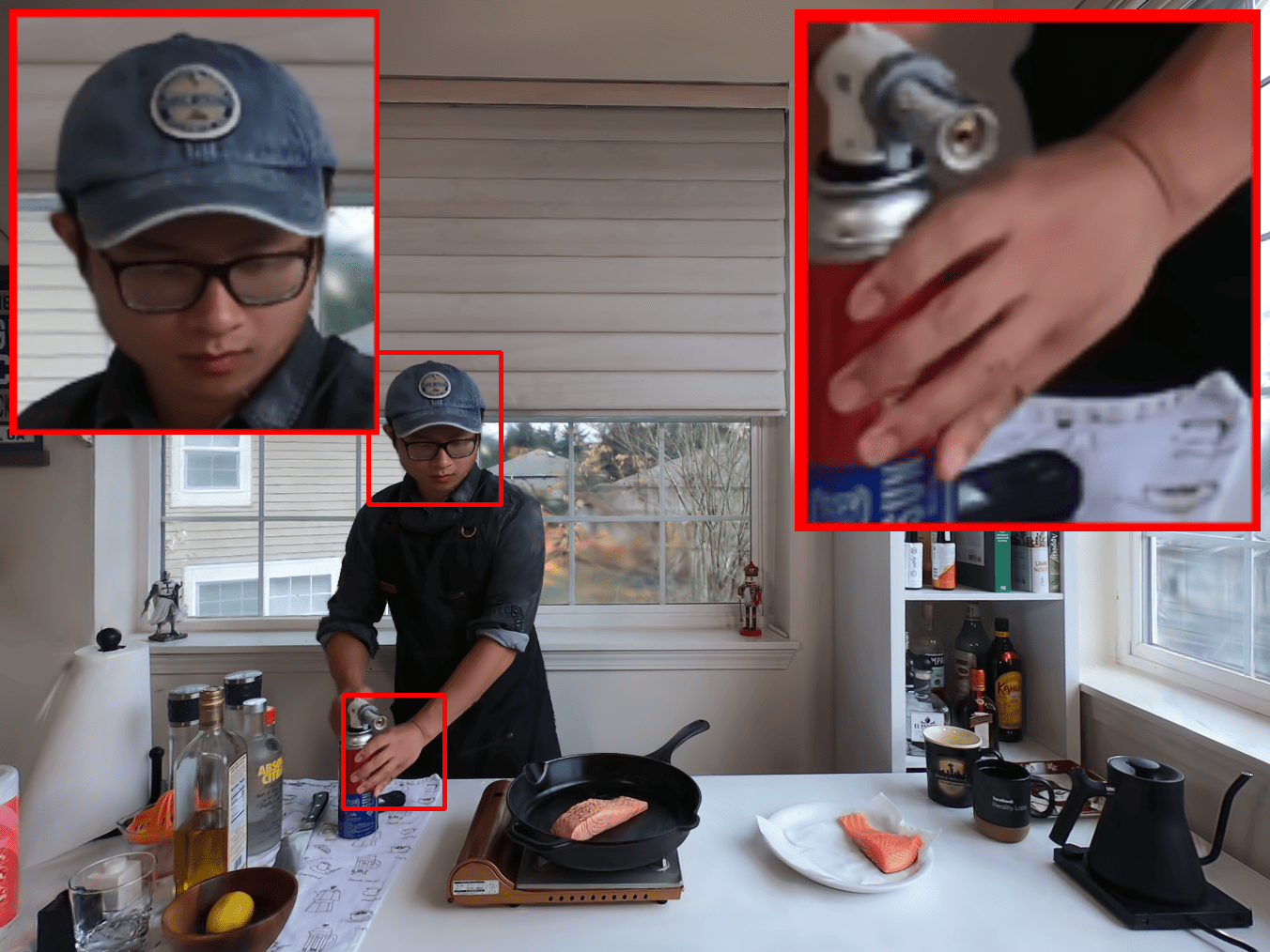}
        \caption*{(g) Product}
    \end{subfigure}
    \hfill
    \begin{subfigure}{0.24\linewidth}
        \centering
        \includegraphics[width=\linewidth]{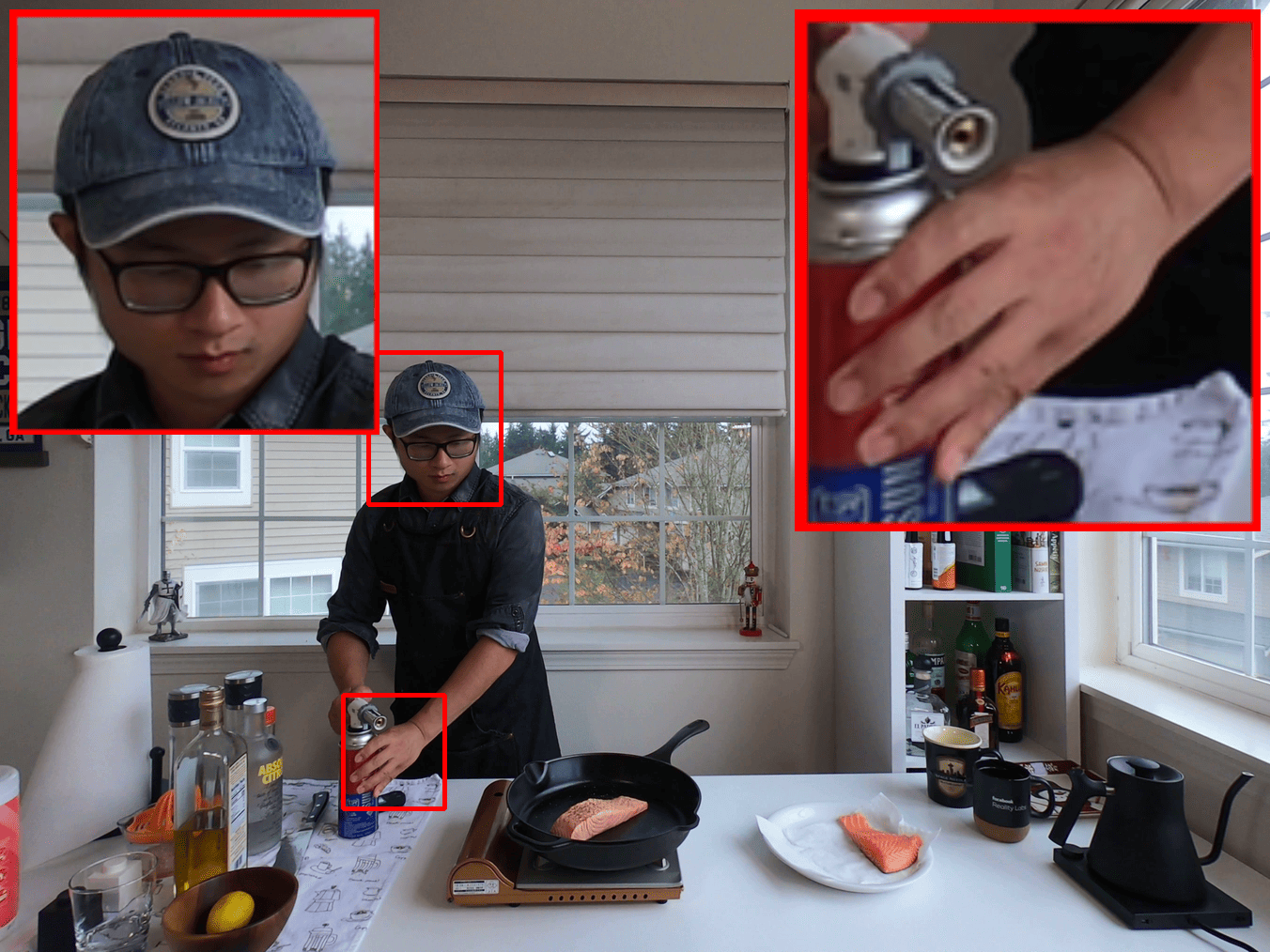}
        \caption*{(h) GT}
    \end{subfigure}
    
    % \vspace{-1mm} % 如果需要，可以调整标题前的间距
    \caption{\textbf{Ablation studies on embedding fusion}.}
    \label{fig:feature_fusion_abl}
    \vspace{-2mm}
\end{figure*}
\paragraph{Embedding Fusion}
We conduct a systematic ablation study to validate our proposed fusion strategy. To isolate its impact, we disabled the canonical field sampling and aggressive opacity reduction modules for this evaluation. We compare various early-fusion schemes against the late-fusion approach (``Dual''), which serves as a strong baseline.

The quantitative comparison is presented in Tab.~\ref{tab:feature_fusion_comparison}. Simple early-fusion variants like ``Coarse'', ``Fine'', ``Concat'', and ``Add'', while fast, exhibit noticeable quality deficiencies. The ``Attention'' mechanism, despite its decent quality, incurs an efficiency trade-off. The results decisively show that our Hadamard product-based (``Product'') strategy not only surpasses other early-fusion schemes but also achieves performance comparable to the computationally expensive late-fusion baseline (``Dual'') while maintaining a significantly higher rendering speed. This validates our core hypothesis: a well-designed early fusion for temporal embeddings is a highly effective paradigm that strikes a superior balance between quality and efficiency. The qualitative results in Fig.~\ref{fig:feature_fusion_abl} further corroborate this. When handling detailed regions with intense motion, such as the moving head, the hand, and the spray gun, our Hadamard product-based method demonstrates a markedly superior reconstruction, highlighting its capability to capture complex dynamic details effectively.
\begin{figure*}[t!]
    \centering

    % --- ROW 1: Dynamic Scene (Broom) with Captions ---
    \begin{subfigure}[b]{0.24\linewidth}
        \includegraphics[width=\linewidth, trim=10 40 10 10, clip]{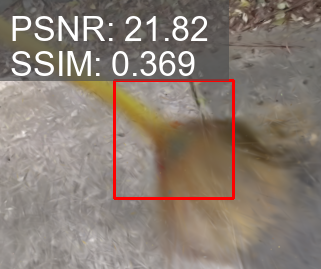}
    \end{subfigure}
    \hfill % Automatically creates space
    \begin{subfigure}[b]{0.24\linewidth}
        \includegraphics[width=\linewidth, trim=10 40 10 10, clip]{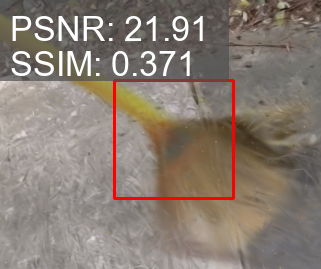}
    \end{subfigure}
    \hfill
    \begin{subfigure}[b]{0.24\linewidth}
        \includegraphics[width=\linewidth, trim=10 40 10 10, clip]{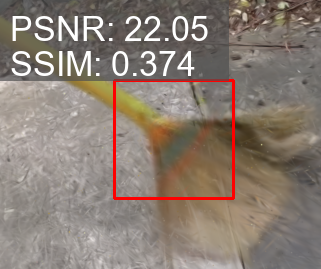}
    \end{subfigure}
    \hfill
    \begin{subfigure}[b]{0.24\linewidth}
        \includegraphics[width=\linewidth, trim=10 40 10 10, clip]{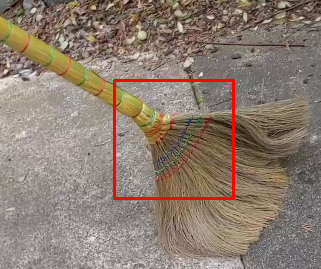}
    \end{subfigure}

    \vspace{2mm} % Vertical space between rows

    % --- ROW 2: Static Background Scene (flame_salmon) WITHOUT Captions ---
    \begin{subfigure}[b]{0.24\linewidth}
        \includegraphics[width=\linewidth]{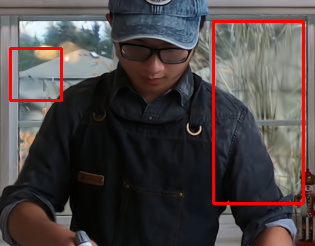}
        \caption*{(a) w/o Sampling}
    \end{subfigure}
    \hfill
    \begin{subfigure}[b]{0.24\linewidth}
        \includegraphics[width=\linewidth]{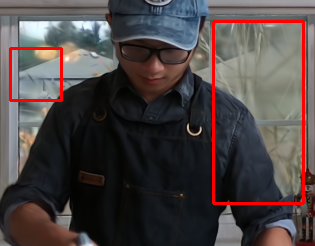}
        \caption*{(b) Inverse Deform}
    \end{subfigure}
    \hfill
    \begin{subfigure}[b]{0.24\linewidth}
        \includegraphics[width=\linewidth]{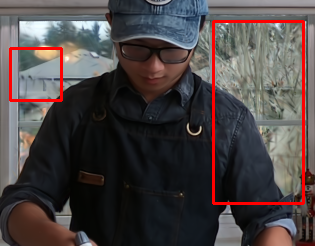}
        \caption*{(c) Ours}
    \end{subfigure}
    \hfill
    \begin{subfigure}[b]{0.24\linewidth}
        \includegraphics[width=\linewidth]{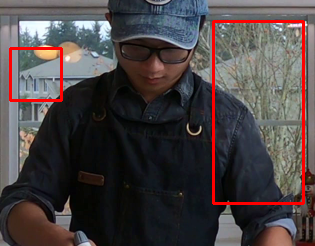}
        \caption*{(d) Ground Truth}
    \end{subfigure}

    \caption{
        \textbf{Ablation studies of Canonical Field Sampling.} The top row shows results on the dynamic \textit{broom} scene, and the bottom row shows results on the static background of the \textit{salmon} scene. Our method (c) shows clear improvements in both cases.
    }
    \label{fig:canonical_sampling_abl_broom_extended}
    \vspace{-2mm}
\end{figure*}
% ========== END OF FINAL VERSION ==========
\begin{table}
  \centering
  % \resizebox{\columnwidth}{!}{%
  \setlength{\tabcolsep}{22pt} 

  \caption{\textbf{Comparison of different sampling strategies on the HyperNeRF dataset.}}
    \begin{tabular}{@{}lccc@{}}
      \toprule
      Method & PSNR$\uparrow$ & SSIM$\uparrow$ & LPIPS$\downarrow$ \\
      \midrule
    w/o Sampling   & 25.96         &0.717          &0.194           \\
    Inverse Deform & 26.08    & 0.718    & 0.192   \\
    Ours  &\textbf{26.31}  &\textbf{0.721} &\textbf{0.187} \\
      \bottomrule
    \end{tabular}
  % }
  \label{tab:canonical_comparison}
\end{table}
\paragraph{Canonical Field Sampling Strategy}

To rigorously evaluate the effectiveness of our sampling strategy, we compare our full method (``Ours'') against two critical variants: (1) disabling the sampling mechanism entirely (``w/o Sampling'') and (2) employing an alternative strategy based on inverse deformation (``Inverse Deform''). As shown in Tab. \ref{tab:feature_fusion_comparison}, incorporating canonical field sampling into the Product baseline leads to consistent rendering quality improvements across the N3DV dataset. Furthermore, Tab. \ref{tab:canonical_comparison} demonstrates that our strategy achieves the highest average performance on the HyperNeRF dataset, outperforming the inverse deformation approach by a notable margin. We also provide qualitative comparisons on the challenging \textit{broom} and \textit{flame\_salmon} scenes in Fig.~\ref{fig:canonical_sampling_abl_broom_extended}. The visual results clearly demonstrate that our approach successfully reconstructs both the moving broom and complex background details, whereas the variants fail to produce comparable results. This confirms that our low-cost anchor injection directly simplifies the learning task, circumventing the instability of solving an ill-posed inverse mapping.

\begin{figure}[!t]
    \centering
    % 第一行：Ours, wo_OR, GT
    \begin{subfigure}{0.32\linewidth} % 使用 \linewidth 而不是 \textwidth
        \centering
        \includegraphics[width=\linewidth]{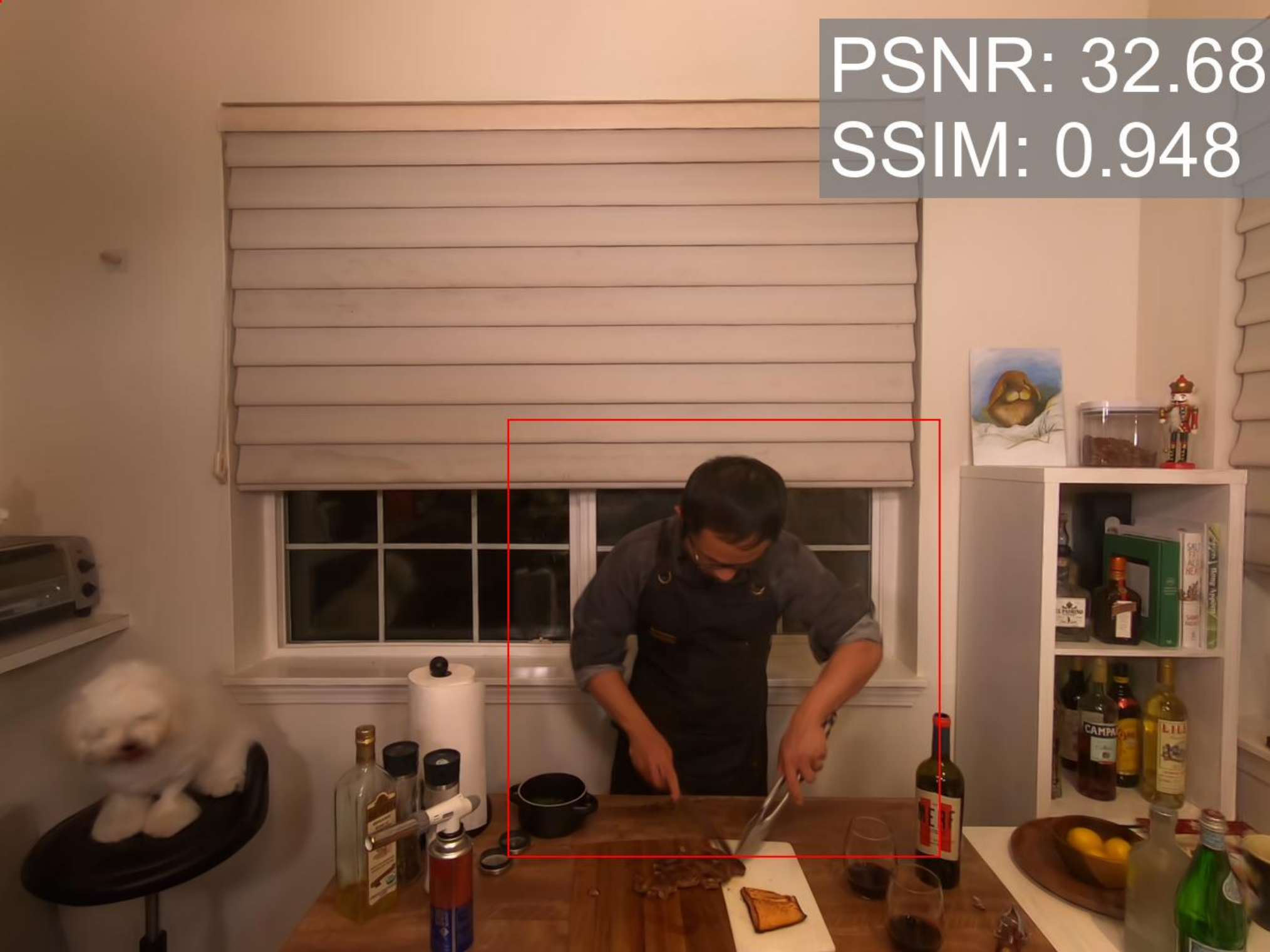}
        \caption*{(a) Ours}
    \end{subfigure}
    \hfill
    \begin{subfigure}{0.32\linewidth}
        \centering
        \includegraphics[width=\linewidth]{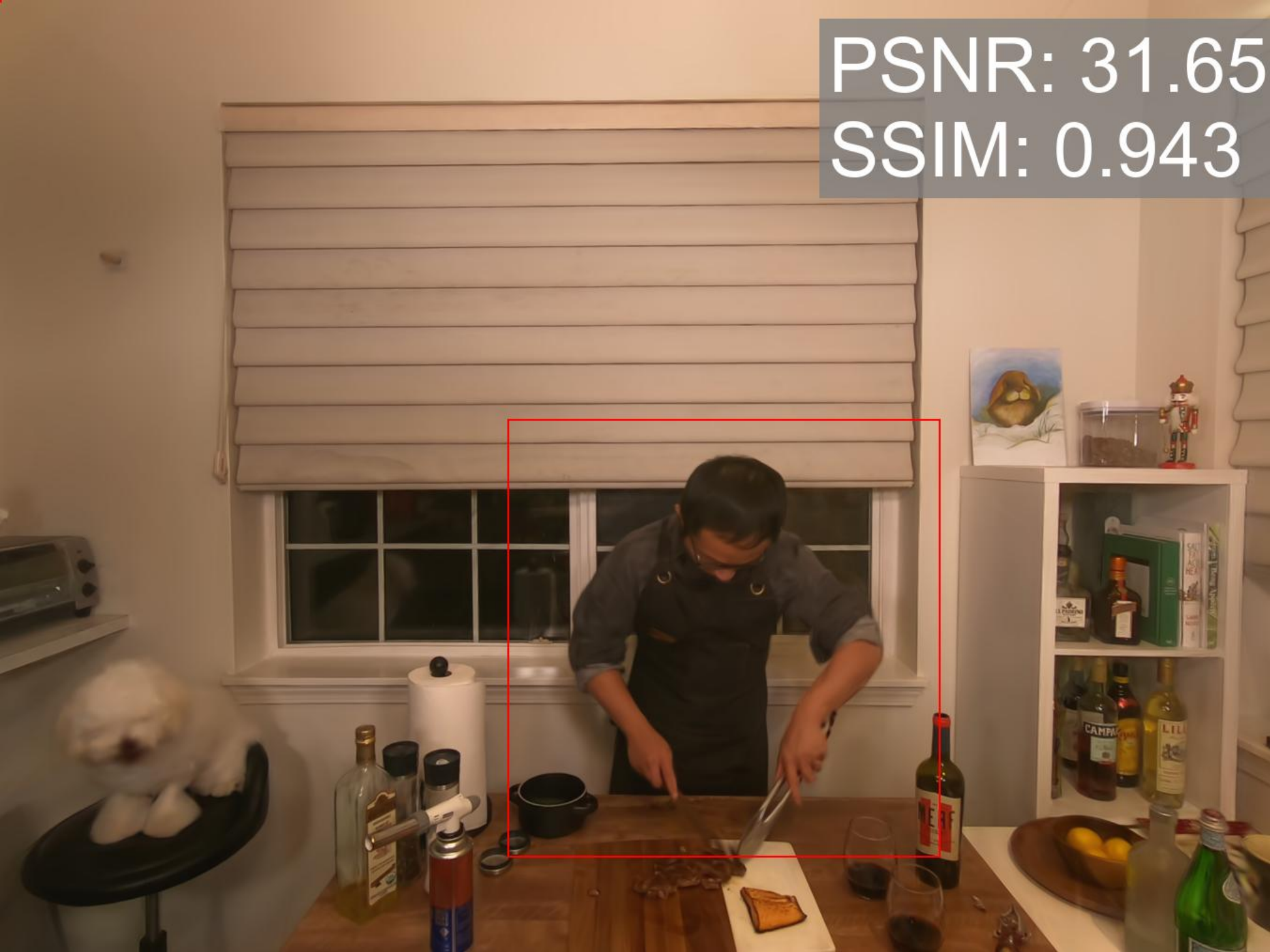}
        \caption*{(b) w/o\_OR}
    \end{subfigure}
    \hfill
    \begin{subfigure}{0.32\linewidth}
        \centering
        \includegraphics[width=\linewidth]{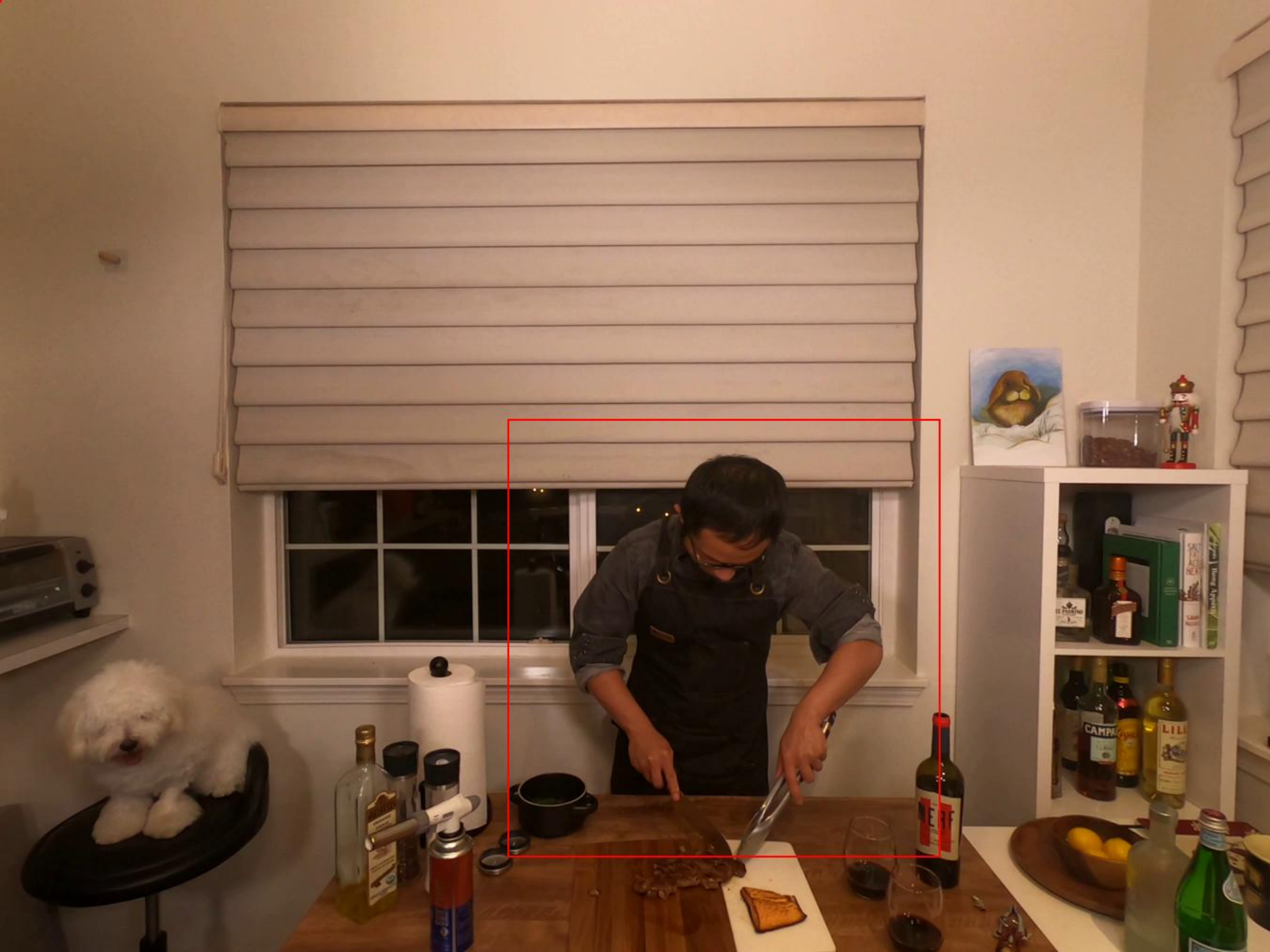}
        \caption*{(c) Ground Truth}
    \end{subfigure}
    % 第二行：热力图
    \begin{subfigure}{\linewidth}
        \centering
        \includegraphics[width=1.0\linewidth]{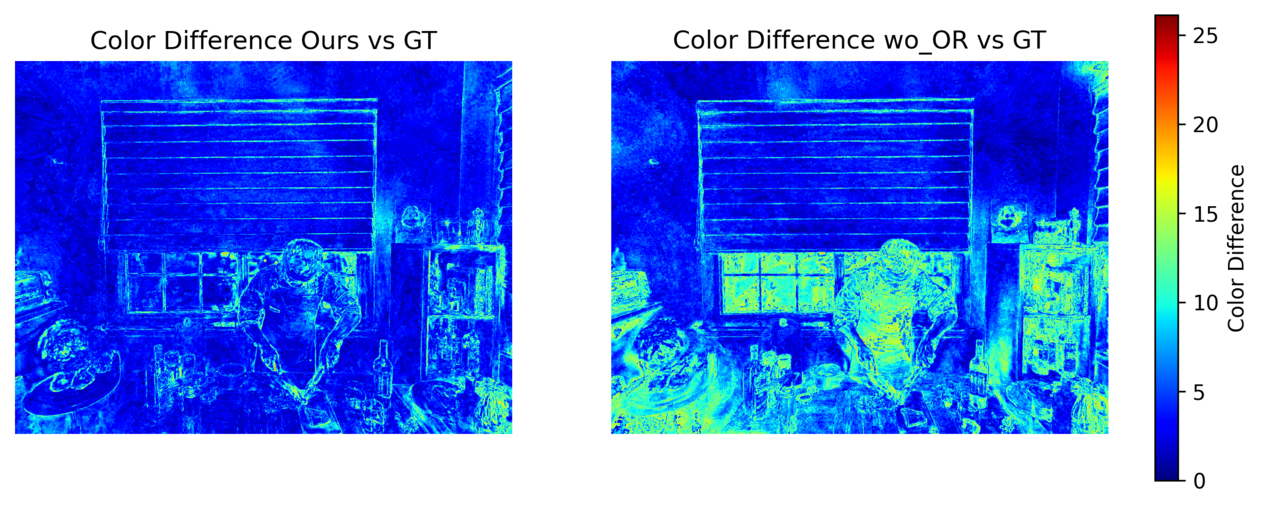} 
        \vspace{-6mm}
        \caption*{(d) Color Difference Heatmap}
    \end{subfigure}
    \caption{\textbf{Ablation studies on adaptive opacity reduction}.}
    \label{fig:opacity_reduction_abl}
    \vspace{-3mm}
\end{figure}
% % ========== UPDATED TABLE WITH LPIPS COLUMN ==========
% \begin{table}[h]
%   \centering
  
%   % Adjusted tabcolsep to accommodate the new column. 
%   % You can fine-tune this value.
%   \setlength{\tabcolsep}{18pt} 
  
%   \caption{
%     \textbf{Sensitivity analysis of the parameter $k$} in our aggressive opacity reduction strategy. 
%     The experiment was conducted on the \textit{cut\_roasted\_beef} scene. 
%     $k=0$ corresponds to disabling the strategy. Our chosen value is highlighted in bold.
%   }
  
%   % --- KEY CHANGES: ---
%   % 1. Changed "ccc" to "cccc" to add a fourth column.
%   % 2. Added the LPIPS column header.
%   % 3. Added LPIPS data for each row.
%   \begin{tabular}{cccc}
%     \toprule
%     k value & PSNR$\uparrow$ & SSIM$\uparrow$ & LPIPS$\downarrow$ \\
%     \midrule
%     0         & 31.65 & 0.943 & 0.047 \\ % <-- TODO: Add your data here
%     1         & 31.99 & 0.942 & 0.046 \\ % <-- TODO: Add your data here
%     5         & 32.55 & 0.945 & 0.044 \\ % <-- TODO: Add your data here
%     \textbf{10} & \textbf{32.68} & \textbf{0.948} & \textbf{0.040} \\ % <-- TODO: Add your data here
%     20        & 32.13 & 0.940 & 0.045 \\ % <-- TODO: Add your data here
%     \bottomrule
%   \end{tabular}
%   \label{tab:k_sensitivity_ablation}
% \end{table}
% % ========== END OF UPDATED TABLE ==========

% ========== UPDATED TABLE WITH LPIPS COLUMN ==========
\begin{table}[h]
  \centering
  
  % Adjusted tabcolsep to accommodate the new column. 
  % You can fine-tune this value.
  \setlength{\tabcolsep}{18pt} 
  
  \caption{
    \textbf{Sensitivity analysis of the parameter $k$} in our aggressive opacity reduction strategy. 
    The experiment was conducted on the N3DV dataset. 
    $k=0$ corresponds to disabling the strategy. Our chosen value is highlighted in bold.
  }
  
  % --- KEY CHANGES: ---
  % 1. Changed "ccc" to "cccc" to add a fourth column.
  % 2. Added the LPIPS column header.
  % 3. Added LPIPS data for each row.
  \begin{tabular}{cccc}
    \toprule
    k value & PSNR$\uparrow$ & SSIM$\uparrow$ & LPIPS$\downarrow$ \\
    \midrule
    0         & 30.86 & 0.939 & 0.052 \\ % <-- TODO: Add your data here
    1         & 30.89 & 0.940 & 0.052 \\ % <-- TODO: Add your data here
    5         & 30.93 & 0.940 & 0.051 \\ % <-- TODO: Add your data here
    \textbf{10} & \textbf{31.01} & \textbf{0.941} & \textbf{0.050} \\ % <-- TODO: Add your data here
    20        & 30.80 & 0.935 & 0.055 \\ % <-- TODO: Add your data here
    \bottomrule
  \end{tabular}
  \label{tab:k_sensitivity_ablation}
\end{table}
% ========== END OF UPDATED TABLE ==========
\paragraph{Aggressive Opacity Reduction Strategy}
We first evaluate the overall impact of our aggressive opacity reduction strategy across the N3DV dataset. As shown in Tab.~\ref{tab:feature_fusion_comparison}, integrating this strategy leads to a consistent improvement in average rendering quality.
We further validate this quantitatively and qualitatively on the challenging \textit{cut\_roasted\_beef} scene. We conducted six trials each for our method with (``Ours'') and without (``w/o\_OR'') this strategy on the scene. As shown in Fig.~\ref{fig:opacity_reduction_abl}, the inclusion of our strategy yields a clear quantitative improvement. The qualitative comparison further validates its impact: upon closer inspection, our method exhibits significantly closer color reproduction to the Ground Truth, an effect highlighted by the color difference heatmaps. This experiment confirms the necessity of the proposed module in mitigating color shifts in challenging dim scenes.

Having established the effectiveness of the strategy, we now analyze its controlling hyperparameter, $k$, to justify our choice of its value. The parameter $k$ determines the intensity of the opacity attenuation. We experimented with several values for $k$, and the results are summarized in Tab.~\ref{tab:k_sensitivity_ablation}. As the table indicates, performance gradually improves as $k$ increases from 0, peaking around $k=10$. When $k$ becomes too large (e.g., $k=20$), the reduction becomes overly aggressive, leading to a slight degradation in quality. This analysis confirms that our chosen value of $k=10$ provides an optimal balance between mitigating artifacts and preserving useful details.

\section{Limitation}
\label{sec:limitation} % Added a label for easy referencing

\begin{figure}[t!]
    \centering % Center the figure within the column
    
    % --- First Image: Broom (cropped to top half) ---
    \begin{subfigure}{0.48\columnwidth}
        \adjustbox{width=\linewidth, trim={0 {0.3\height} 0 {0.1\height}}, clip}{%
            \includegraphics{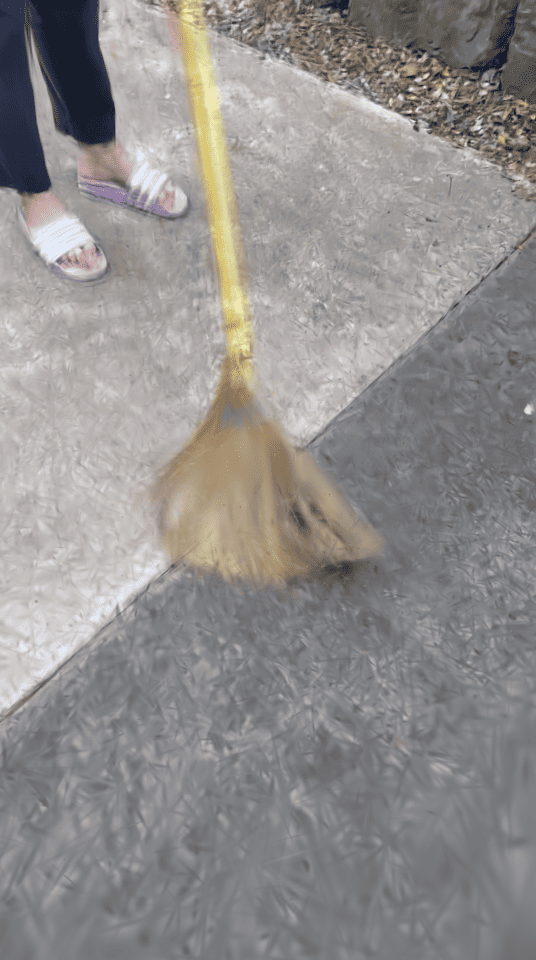}%
        }
        \caption{Broom}
        \label{fig:limitation_broom}
    \end{subfigure}
    \hfill % Adds a flexible space between the two images
    % --- Second Image: 3D Printer (cropped to top half) ---
    \begin{subfigure}{0.48\columnwidth}
        \adjustbox{width=\linewidth, trim={0 {0.3\height} 0 {0.1\height}}, clip}{%
            \includegraphics{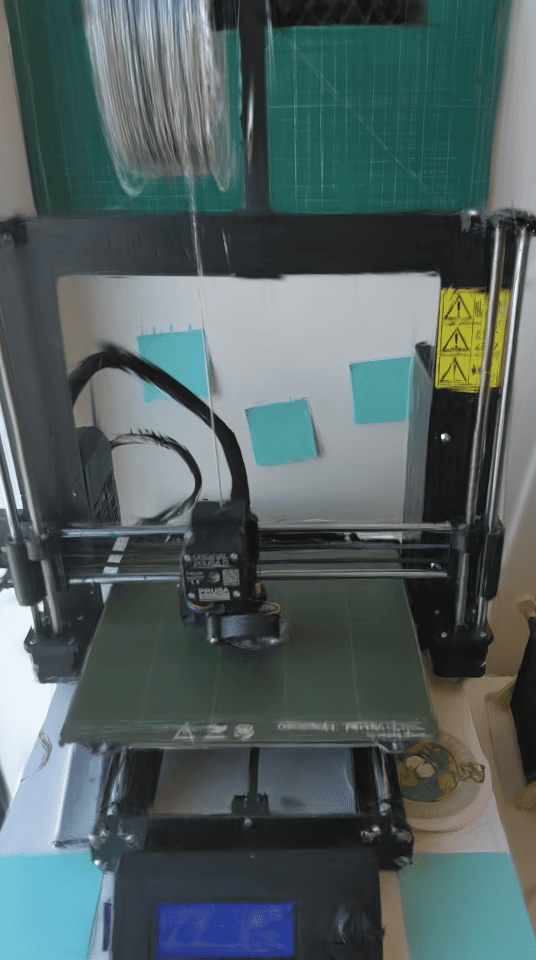}%
        }
        \caption{3D Printer}
        \label{fig:limitation_printer}
    \end{subfigure}

    \caption{
        \textbf{Illustration of limitations on challenging monocular scenes.}
        Failure cases show blurry reconstructions and artifacts in regions with significant view information loss. (a) The Broom scene. (b) The 3D Printer scene.
    }
    \label{fig:limitation_examples}
    \vspace{-1mm}
\end{figure}

Our method faces challenges when dealing with scenes that have significant missing view information, a common issue particularly prevalent in monocular videos. This manifests as a tendency for the model to produce blurry reconstructions when certain regions of the scene are never effectively observed due to persistent occlusion, as illustrated in Fig.~\ref{fig:limitation_examples}.

We attribute this limitation to the highly ill-posed nature of reconstructing these unobserved or sparsely observed regions. Lacking any direct 3D geometric constraints, the optimization process faces inherent ambiguities and struggles to converge to a precise and unique solution.

A potential future direction to address this is to introduce strong external priors to regularize this ill-posed problem. For instance, one could explore using pre-trained 2D diffusion models to infer and generate plausible, view-consistent completions for the occluded areas based on observed regions. These generated images could then serve as additional supervision, compensating for the lack of real observations.

% ========== START OF THE MODIFIED LIMITATION FIGURE ==========

% Make sure you have \usepackage{subcaption} and \usepackage{adjustbox} in your document's preamble

% ========== END OF THE MODIFIED LIMITATION FIGURE ==========

\section{Conclusion}
\label{sec:conclusion} % Added a label for the conclusion

We present FRoG, a novel deformable 3D Gaussian splatting method for fast and robust dynamic scene reconstruction. FRoG effectively captures temporal information of dynamic scenes through a Hadamard product-based temporal embedding fusion strategy, achieving fast rendering while maintaining high reconstruction quality. Additionally, FRoG introduces a canonical field sampling strategy and an aggressive opacity reduction strategy to further enhance the model's robustness, significantly improving the reconstruction quality in regions with sparse initial point clouds and enhancing the model's effects in dim scenes. Experimental results demonstrate that FRoG not only achieves fast rendering speed but also exhibits superior rendering quality.

% \section{Acknowledgement}  This work was supported in part by Zhejiang Province Program (22024C03263, 025C01068, LZ25F020006), Zhejiang Provincial Cultural Relics Protection Science and Technology Project (2024009), the National Program of China (62172365), Macau project: Key technology research and display system development for new personalized controllable dressing dynamic display, and Ningbo Science and Technology Plan Project (2025Z052, 2025Z062, 2022Z167, 2023Z137).

% ========== END OF THE RESTRUCTURED SECTIONS ==========
\bibliographystyle{IEEEtran} 
\bibliography{IEEEtran}

% 个人信息
\vspace{-3em}
\begin{IEEEbiography}[{\includegraphics[width=1in,height=1.25in,clip,keepaspectratio]{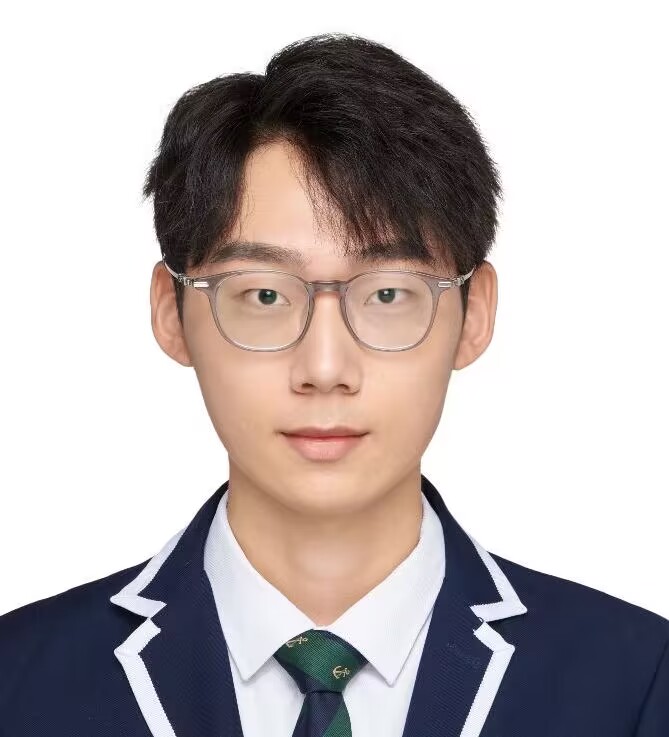}}] {Han Jiao} received his bachelor’s degree from Zhejiang University, Hangzhou, China. He is currently pursuing a master's degree at the College of Computer Science and Technology, Zhejiang University. His research interests lie in computer graphics and computer vision, with a focus on 3D vision and physics simulation.
\end{IEEEbiography}

\vspace{-3em}
\begin{IEEEbiography}[{\includegraphics[width=1in,height=1.25in,clip,keepaspectratio]{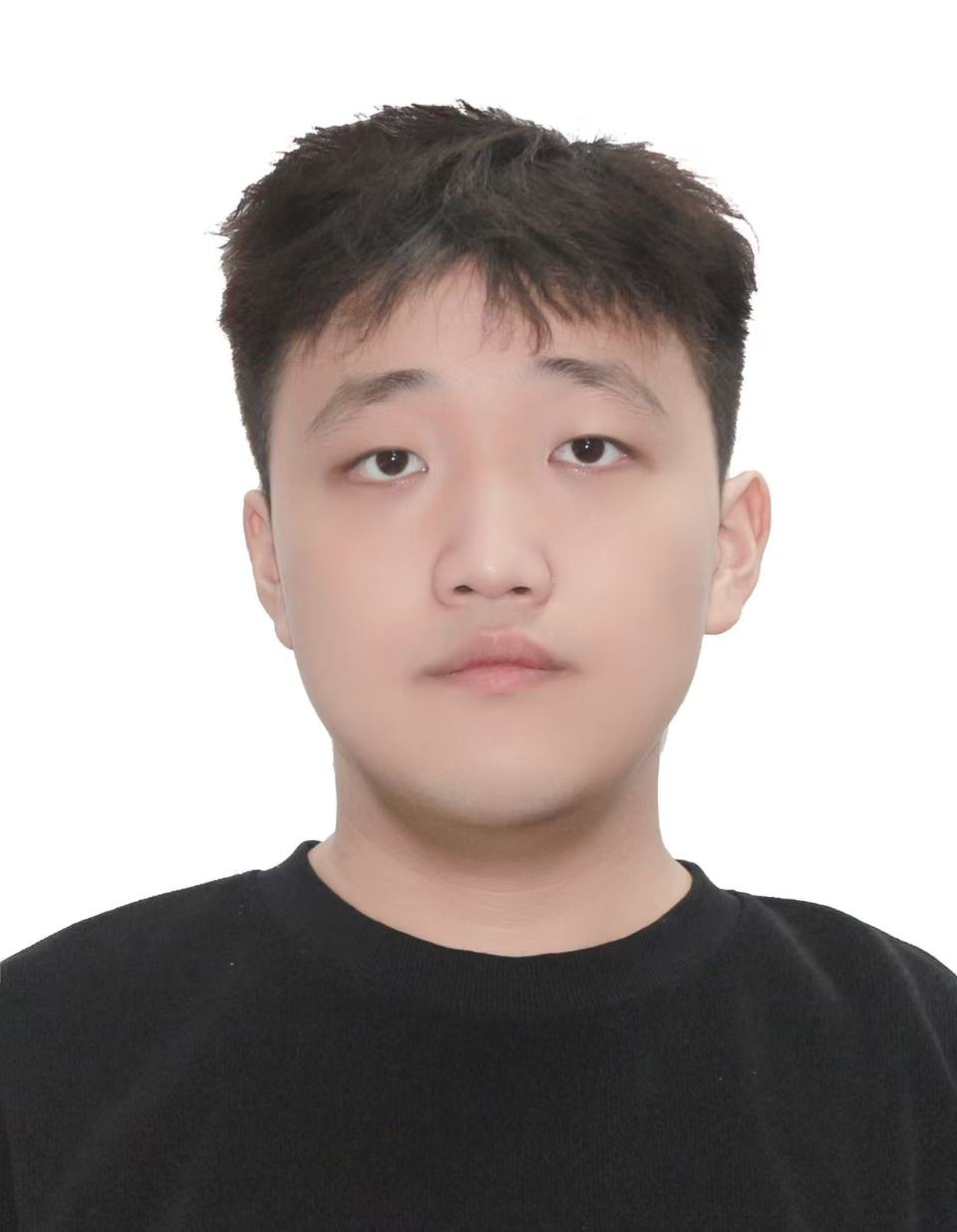}}] 
{Jiakai Sun} Jiakai Sun received his B.E. degree in 2022 and his M.S. degree in 2025, both from the College of Computer Science and Technology, Zhejiang University. His research interests lie in computer graphics and computer vision, with a focus on neural rendering and 3D vision.
\end{IEEEbiography}

\vspace{-3em}
\begin{IEEEbiography}[{\includegraphics[width=1in,height=1.25in,trim=1pt 0 0 0,clip,keepaspectratio]{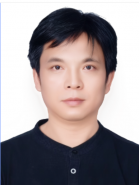}}] 
{Lei Zhao} received his Ph.D. degree from Zhejiang
University in 2009. He is now an associate professor
at Zhejiang University. His research interests include
spatial intelligence, embodied intelligence, and multimodal large models.
\end{IEEEbiography}

\vspace{-3em}
\begin{IEEEbiography}[{\includegraphics[width=1in,height=1.25in,clip,keepaspectratio]{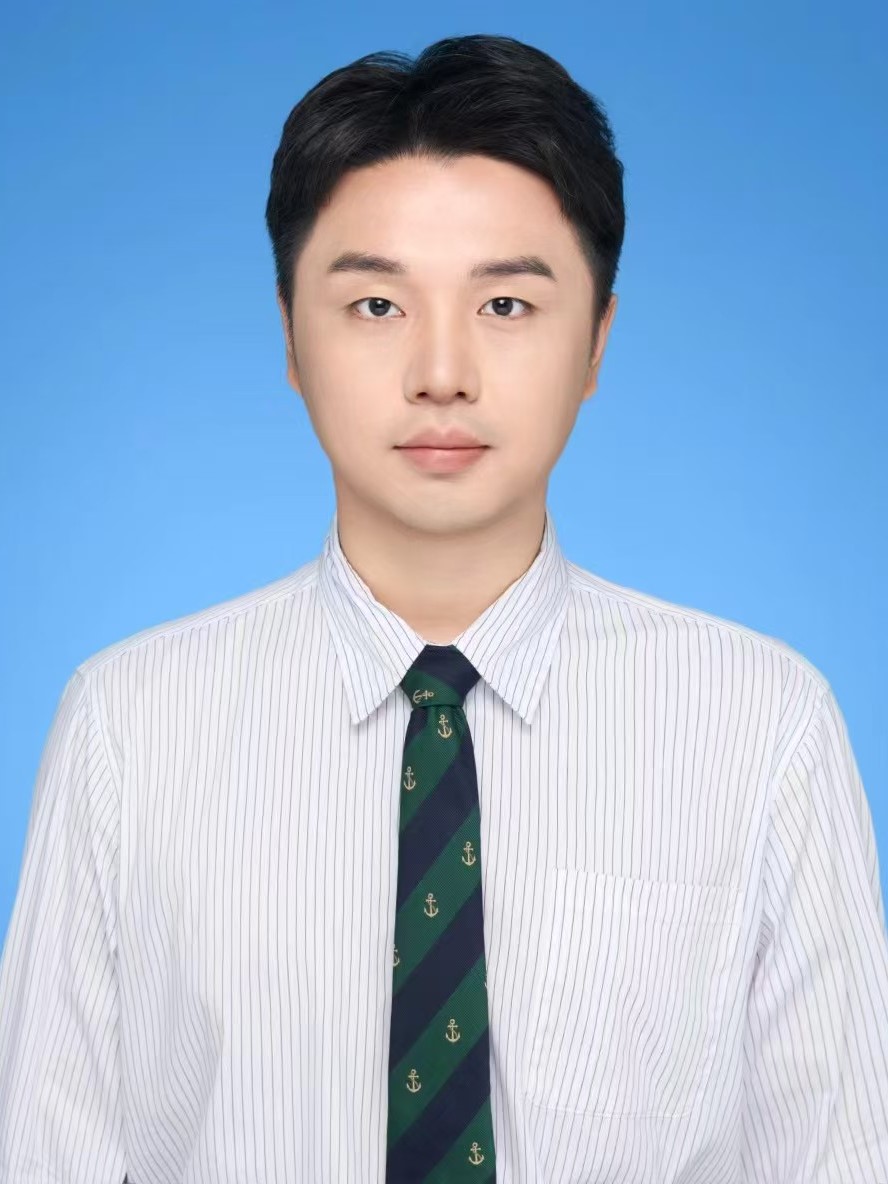}}] 
{Zhanjie Zhang} received the bachelor’s degree from Jiangsu University, Zhenjiang, China, in 2019, and the master’s degree from Jiangnan University, Wuxi, China, in 2022. He is a Ph.D. candidate at the College of Computer Science and Technology, Zhejiang University, Hangzhou. His research interests are generative adversarial networks, style transfer, computer vision, and deep learning.
\end{IEEEbiography}
 
\vspace{-3em}
\begin{IEEEbiography}[{\includegraphics[width=1in,height=1.25in,clip,keepaspectratio]{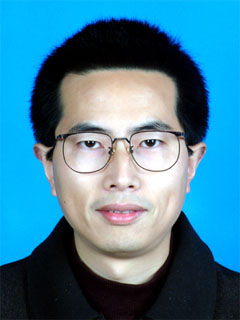}}] 
{Wei Xing} is an associate professor at the College of Computer Science and Technology, Zhejiang University, Hangzhou, 310013, China. His research interests include AI, knowledge learning, and intelligent image processing and multimedia technology.
\end{IEEEbiography}

\vspace{-3em}
\begin{IEEEbiography}[{\includegraphics[width=1in,height=1.25in,clip,keepaspectratio]{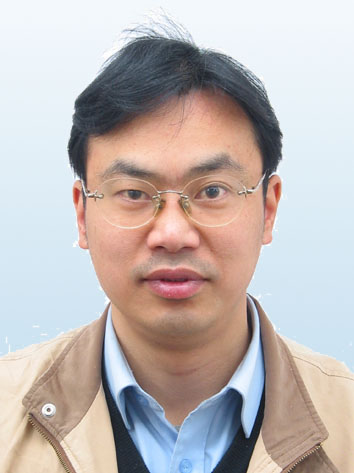}}] 
{Huaizhong Lin} received his Ph.D. degree from Zhejiang
University in 2002 He is now an associate professor
in the College of Computer Science and Technology
of Zhejiang University. His research interests include
computer graphics and multimodal large models.
\end{IEEEbiography}

\end{document}